\documentclass{article}
\usepackage{iclr2027_conference,times}

\usepackage{amsmath,amsfonts,bm}
\usepackage{algorithm}
\usepackage{algorithmic}
\usepackage{amssymb}
\usepackage{amsmath}
\usepackage{booktabs}
\usepackage{multirow}
\usepackage{pifont}
\usepackage{colortbl}
\usepackage{graphicx}

\usepackage{xspace}
\usepackage{subcaption}
\usepackage{multirow}
\usepackage{amsmath,amssymb}
\usepackage{mathtools}
\usepackage{amsfonts}
\usepackage{bm}
\usepackage{xfrac}
\usepackage[english]{babel}
\usepackage{hyphenat}
\usepackage{tabularx}
\usepackage{xcolor}
\usepackage{colortbl}
\usepackage{soul}
\usepackage{csquotes}
\usepackage{listings}
\usepackage{seqsplit}
\usepackage{arydshln}
\usepackage{wrapfig}
\usepackage{enumitem}
\usepackage{algorithm}
\usepackage{algorithmic}
\usepackage{graphicx}
\usepackage{afterpage}
\usepackage{pifont}
\usepackage{lipsum}
\usepackage{nicematrix}

\newcommand{\name}{\textsc{Vis}\xspace}

\definecolor{Gray}{gray}{0.85}
\definecolor{Redo}{rgb}{0.95,0.69,0.51}
\definecolor{LightCyan}{rgb}{0.88,1,1}

\newcommand{\bbR}{\mathbb{R}}

\newcommand{\D}{\mathcal{D}}
\newcommand{\Y}{\mathcal{Y}}

\newcommand{\vx}{\mathbf{x}}

\newcommand{\vt}{\mathbf{t}}

\newcommand{\vone}{\mathbf{1}}

\newcommand{\x}{\vx}

\newcommand{\1}{\vone}

\newcommand{\ie}{\textit{i.e.}\@\xspace}

\def\eqref#1{equation~\ref{#1}}

\def\1{\bm{1}}

\def\vone{{\bm{1}}}

\def\vt{{\bm{t}}}

\def\vx{{\bm{x}}}

\DeclareMathAlphabet{\mathsfit}{\encodingdefault}{\sfdefault}{m}{sl}
\SetMathAlphabet{\mathsfit}{bold}{\encodingdefault}{\sfdefault}{bx}{n}

\DeclareMathOperator*{\argmax}{arg\,max}

\usepackage{graphicx}
\usepackage{booktabs}
\usepackage{multirow}
\usepackage{hyperref}
\usepackage{url}
\usepackage{xspace}
\hypersetup{hidelinks}
\usepackage{marvosym}
\title{Visual Branch is What You Need for CLIP-based Class-Incremental Learning}

\author{
{Tao Hu$^{1,2}$  \quad
Zhen-Hao Xie$^{1,2}$    \quad
Jingcai Guo$^{3}$    \quad
De-Chuan Zhan$^{1,2}$    \quad
Da-Wei Zhou$^{1,2}\textsuperscript{(\Letter)}$} \\
$^{1} $ School of Artificial Intelligence, Nanjing University \\
$^{2} $ State Key Laboratory for Novel Software Technology, Nanjing University \\
$^{3} $ Hong Kong Polytechnic University \\
\texttt{\small \{hut, wenzh, zhandc zhoudw\}@lamda.nju.edu.cn}\\
}

\iclrfinalcopy

\begin{document}

\maketitle
\fancyhead[L]{Preprint.}

\begin{abstract}
Class-Incremental Learning (CIL) requires models to recognize new classes over time without forgetting previously learned ones. 
With the rise of vision-language pre-training, CLIP has become a strong foundation for CIL. 
A common design in CLIP-based CIL is to construct textual classifier weights by encoding class-name templates with the CLIP text encoder, and then classify visual features by image-text cosine similarity. This design is appealing: since CLIP aligns images and text in a shared embedding space, textual weights appear to provide an off-the-shelf classifier for incremental classes. 
However, we show that this seemingly natural design is not always beneficial, as a modality gap can still separate the two modalities and make textual classifier weights deviate from visual class distributions.
Empirically, under identical task-wise CIL training, initializing the cosine classifier with visual class centers yields lower loss and better incremental accuracy than using CLIP textual features.
Motivated by these observations, we propose \name (\underline{\textbf{V}}isual \underline{\textbf{I}}ncremental \underline{\textbf{S}}VM), a visual-only method for CLIP-based CIL that removes the deployed textual branch and constructs the incremental classifier entirely in the visual space. To obtain stronger task-adaptive visual representations, \name uses only base-session data to enhance CLIP's final visual representation with informative visual-layer features. Built on the enhanced visual representation, \name employs a simple kernelized incremental least-squares SVM, whose classifier weights are solved in closed form from additive sufficient statistics. When new classes arrive, \name accumulates their sufficient statistics and recomputes the classifier weights for all seen classes, enabling efficient incremental updates while preserving historical class knowledge.
Extensive experiments show that \name achieves state-of-the-art performance without a textual branch.
\end{abstract}
   
\section{Introduction}
\label{sec:intro}

In recent years, deep neural networks have achieved remarkable progress in visual recognition~\citep{he2016deep,dosovitskiy2020image}. However, real-world visual systems often face dynamic environments, where new categories emerge continuously and data distributions evolve over time~\citep{rebuffi2017icarl,zhao2020maintaining,Xie2026SAME}. When conventional models are updated on such non-stationary streams, they may overwrite previously acquired knowledge, leading to catastrophic forgetting~\citep{french1999catastrophic,serra2018overcoming,shi2021overcoming}. Class-Incremental Learning (CIL)~\citep{de2021continual,gao2022r,zhou2024class,zhou2025revisiting} addresses this challenge by requiring a model to recognize newly arriving classes while maintaining discrimination over previously learned classes. Recently, large-scale pre-trained models have become increasingly attractive for CIL by providing generalizable representations and enabling lightweight adaptation instead of training incremental models from scratch~\citep{wang2022learning,zhou2025learning}. In particular, vision-language models such as CLIP~\citep{radford2021learning} offer a promising foundation for CIL by combining transferable visual representations with open-vocabulary recognition capabilities~\citep{huang2025mind,zhou2025external}.

Among vision-language models, CLIP is particularly appealing for CIL because it provides a powerful visual encoder together with a zero-shot classification mechanism~\citep{radford2021learning}. Specifically, CLIP constructs textual classifier weights by encoding class-name templates with the text encoder, and classifies visual features according to their image-text cosine similarities. This zero-shot capability naturally motivates CLIP-based CIL methods to construct incremental classifiers using the textual branch: some directly classify images by visual-textual similarity~\citep{wang2023attriclip}, while others use textual representations to guide prompt tuning~\citep{lu2025continual}, adapter learning and textual-prior guidance~\citep{liu2023class,zhang2025visual}, or classifier construction and representation adjustment~\citep{huang2024class,zhou2025external}. For a new class, such a strategy can derive its classifier weight directly from the class name, making the textual branch a convenient source for incremental classifier construction. However, this strategy relies on a strong assumption: textual classifier weights should provide reliable decision directions for visual samples.

\begin{figure*}[t]
   \vspace{-9mm}
    \centering
    \captionsetup[subfigure]{font=small,labelfont=bf}

    \begin{subfigure}[t]{0.49\textwidth}
        \centering
        \includegraphics[width=\linewidth]{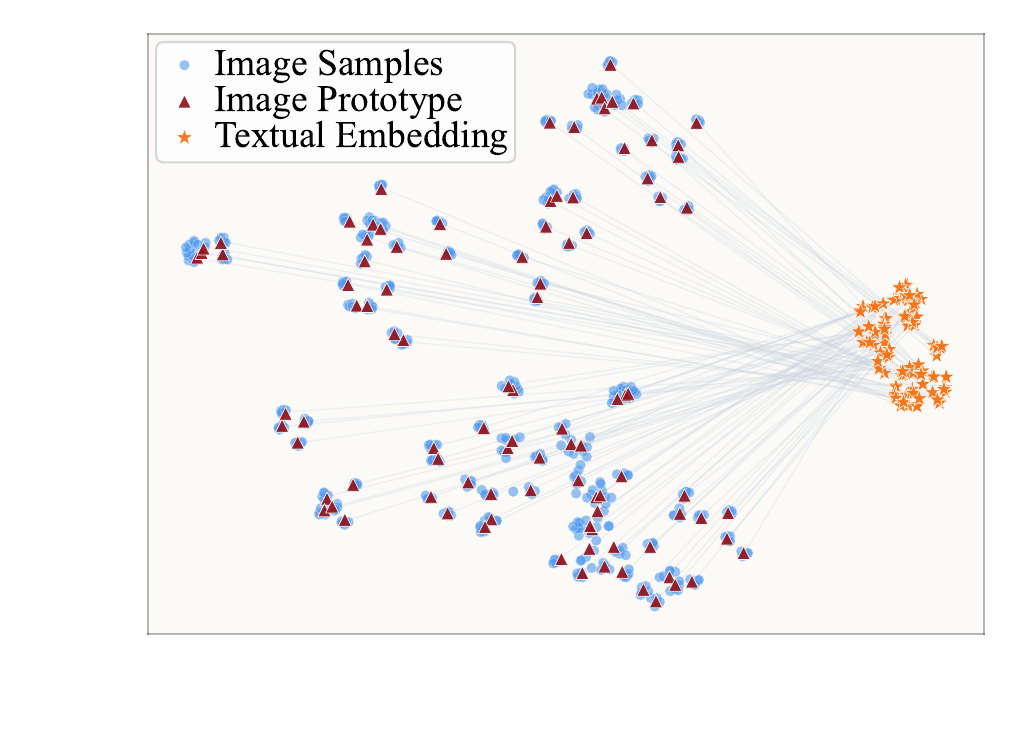}
        \vspace{-8mm}
        \caption{Image-text modality gap in CLIP space.}
        \label{fig:pre_tsne}
    \end{subfigure}
    \hfill
    \begin{subfigure}[t]{0.49\textwidth}
        \centering
        \includegraphics[width=\linewidth]{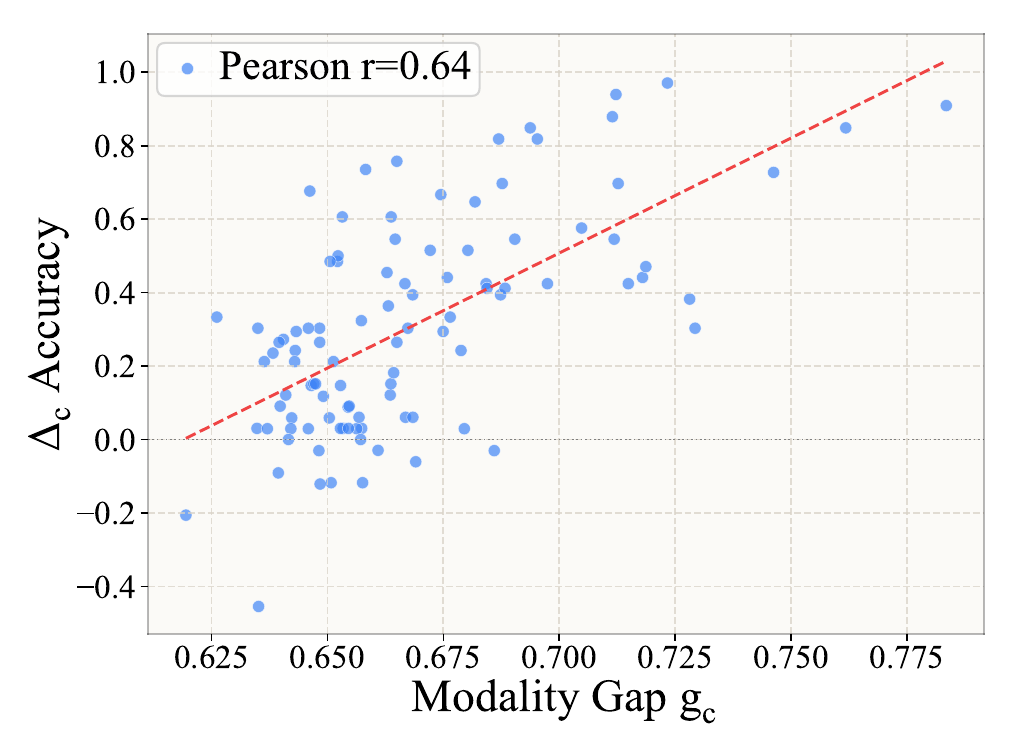}
        \vspace{-8mm}
        \caption{Class-level gap versus textual-head degradation.}
        \label{fig:pre_gapacc}
    \end{subfigure}
    \\
    \begin{subfigure}[t]{1\textwidth}
        \centering
        \includegraphics[width=\linewidth]{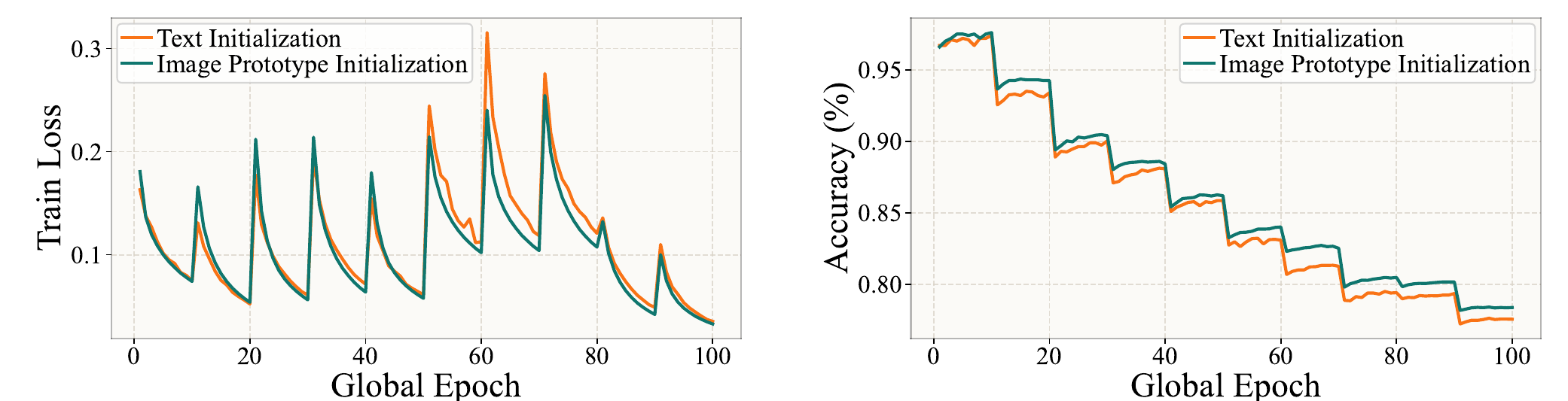}
        \caption{Optimization with text- and visual-based initialization.}
        \label{fig:pre_optim}
    \end{subfigure}
    \vspace{-3mm}
    \caption{\small
Preliminary diagnosis of textual embeddings in CLIP-based CIL.
(a) t-SNE visualization shows a clear separation between visual prototypes and textual embeddings in CLIP's shared image-text embedding space.
(b) The class-level modality gap $g_c=1-\boldsymbol{\mu}_c^\top\boldsymbol{t}_c$ is positively correlated with textual-head degradation $\Delta_c=\mathrm{acc}^{\mathrm{NCM}}_c-\mathrm{acc}^{\mathrm{text}}_c$.
(c) Visual-prototype initialization yields lower training loss and higher seen-class accuracy than textual initialization under the same cosine classifier.
    }
    \label{fig:pre_diagnosis}
    \vspace{-11mm}
\end{figure*}

To verify this assumption, we conduct a preliminary diagnosis in Figure~\ref{fig:pre_diagnosis} from both geometric and optimization perspectives. For each class $c$, we compute a normalized visual prototype $\boldsymbol{\mu}_c$ by averaging normalized CLIP visual features of its samples, and obtain the textual feature $\boldsymbol{t}_c$ by encoding its class-name template. Following the image-text modality gap in CLIP~\citep{liang2022mind}, we define the class-level gap as $g_c = 1-\boldsymbol{\mu}_c^\top \boldsymbol{t}_c$, where a larger value indicates weaker visual-textual alignment. On CIFAR100~\citep{krizhevsky2009learning}, Figure~\ref{fig:pre_tsne} visualizes this gap: image samples and visual prototypes cluster on the visual side, while textual features lie in a separate region of the shared CLIP space. To quantify its classification impact, we use FGVCAircraft~\citep{maji2013fine} to compare two training-free classifiers: the CLIP zero-shot textual head and a nearest-class-mean classifier~\citep{mensink2013distance} based on visual prototypes. We define the class-level textual-head degradation as $\Delta_c=\mathrm{acc}^{\mathrm{NCM}}_c-\mathrm{acc}^{\mathrm{text}}_c$, where larger values mean that visual prototypes outperform textual features. As shown in  Figure~\ref{fig:pre_gapacc}, $g_c$ and $\Delta_c$ exhibit a positive Pearson correlation, indicating that larger modality gaps are associated with larger textual-head degradation. 
We further examine optimization on CIFAR100~\citep{krizhevsky2009learning} under a ten-task CIL protocol, freezing the CLIP visual encoder and training only the same cosine classifier for ten epochs per task; the only difference is whether the classifier is initialized by textual features or visual prototypes.
Figure~\ref{fig:pre_optim} reports the training loss and seen-class accuracy on the test sets of all observed classes over global epochs. Compared with visual-prototype initialization, textual initialization consistently results in higher loss and lower accuracy, indicating a less favorable optimization trajectory for the same cosine classifier. These results suggest that textual classifier weights can be geometrically misaligned with visual class distributions and harder to optimize when used to initialize the classifier. This raises a natural question: can we simply dispense with the textual branch at deployment?

Motivated by this question, we propose \name (\underline{\textbf{V}}isual \underline{\textbf{I}}ncremental \underline{\textbf{S}}VM), a visual-only method for CLIP-based CIL that removes the textual branch at deployment and constructs the incremental classifier entirely in the visual space. Rather than using textual embeddings as classifier weights, \name first strengthens CLIP's final visual representation by incorporating informative visual-layer features with only base-session data. Built on the enhanced visual representation, \name employs a simple kernelized incremental least-squares SVM, whose classifier weights are solved in closed form from additive sufficient statistics. When new classes arrive, \name accumulates their sufficient statistics and recomputes the classifier weights for all seen classes, enabling efficient incremental updates while preserving historical class knowledge.

\section{Related Work}
\label{sec:related_work}

\noindent\textbf{Class-Incremental Learning.}
Class-incremental learning (CIL) studies how to update a model with continuously arriving new classes while maintaining its recognition ability on previously learned ones~\citep{de2021continual,masana2022class}. 
Early CIL methods mainly address catastrophic forgetting from several perspectives. 
Regularization-based approaches reduce forgetting by restricting the change of important parameters or penalizing updates in sensitive parameter directions~\citep{kirkpatrick2017overcoming,aljundi2018memory,zenke2017continual}. 
Replay-based approaches preserve information from old tasks by storing exemplar samples or synthesizing previous data distributions with generative models~\citep{rebuffi2017icarl,chaudhry2018efficient,ostapenko2019learning,xiang2019incremental}. 
Dynamic-architecture methods allocate additional capacity for new tasks through neuron, backbone, or prompt expansion~\citep{yoon2017lifelong,xu2018reinforced,wang2022beef,liu2021adaptive}. 
Another widely used strategy is knowledge distillation, which transfers responses or representations from previous models to the current one to retain old-task knowledge~\citep{hinton2015distilling,li2017learning,rebuffi2017icarl}.

\noindent\textbf{Pre-Trained Model-Based CIL.}
The emergence of large-scale pre-trained models has shifted CIL from learning representations from scratch toward adapting strong frozen backbones such as ViT~\citep{dosovitskiy2020image} and CLIP~\citep{radford2021learning}. 
This paradigm aims to exploit the generalization ability of pre-trained models while introducing only a small number of task-specific parameters~\citep{wang2022learning,smith2023coda,qi2025adaptive,li2025addressing}. 
Prompt-based methods, including L2P~\citep{wang2022learning}, DualPrompt~\citep{wang2022dualprompt}, and CODA-Prompt~\citep{smith2023coda}, adapt frozen transformers by learning or selecting prompt tokens for different tasks. 
Adapter-based methods insert lightweight modules into the backbone to improve task adaptation with limited parameter updates~\citep{fukuda2025adapter,yu2024boosting}. 
For vision-language pre-trained models, CLIP-based CIL methods further leverage cross-modal alignment by using textual prototypes, prompt tuning, adapters, or language-guided representations for incremental recognition~\citep{wang2023attriclip,huang2024class,yu2025language,lu2025continual}.
Another line of work studies analytic or random-feature classifiers~\citep{zhuang2022acil,mcdonnell2023ranpac,zhuang2023gkeal,zhuang2024ds}, which replace gradient-based classifier optimization with closed-form or recursive updates~\citep{lewandowski2025learning,Peng-ICLR2025,DBLP:conf/aaai/MomeniM025} over pre-trained features, enabling efficient exemplar-free incremental learning.\looseness -1

\section{Preliminaries}
\label{sec:prelim}

\noindent\textbf{Class-Incremental Learning.}
We consider the standard CIL setting with a sequence of tasks 
$\{\D_1,\dots,\D_B\}$~\citep{rebuffi2017icarl}. 
Each task $\D_t=\{(\x_i,y_i)\}_{i=1}^{N_t}$ contains samples from a class set $\Y_t$, where different tasks have disjoint label spaces, \emph{i.e.}, 
$\Y_t\cap\Y_s=\varnothing$ for $s<t$. 
After learning task $t$, the model is evaluated over all seen classes 
$\Y_{1:t}=\bigcup_{s=1}^{t}\Y_s$, with cumulative class count $C_t=|\Y_{1:t}|$. 
The goal is to learn a unified classifier 
$f_t:\mathcal{X}\rightarrow \Y_{1:t}$ that recognizes both old and new classes. 
Following the \emph{exemplar-free} protocol~\citep{zhu2021prototype, wang2022dualprompt, wang2022learning}, the learner cannot store or replay samples from previous tasks. 
Therefore, when learning $\D_t$, only the current training data are accessible, while the model is required to preserve discriminative ability on $\Y_{1:t-1}$. 
We refer to the first task $\D_1$ as the base session and the remaining tasks as incremental sessions.
\\\noindent\textbf{Zero-shot CLIP classification.}
A key capability of CLIP is its zero-shot classification mechanism: for a given set of classes, a classifier can be constructed on the fly from their names~\citep{radford2021learning}. 
Specifically, CLIP uses a visual encoder $\Phi_v$ and a textual encoder $\Phi_t$ to encode images and class-name prompts into a shared image-text embedding space.
Given an image $\x_i$, the visual encoder produces a visual embedding $\vt_i=\Phi_v(\x_i)\in\bbR^d$. 
For each class $c\in\Y_{1:t}$, a class-name template, such as ``a photo of a [CLASS]'', is encoded by the textual encoder to produce a textual embedding $\vt_c$, which serves as the classifier weight for class $c$. 
Zero-shot CLIP classifies the image by comparing the cosine similarity between the visual embedding $\vt_i$ and each textual embedding $\vt_c$:
\begin{equation}
\begin{aligned}
    \hat y(\x_i)
    =
    \argmax_{c\in\Y_{1:t}}
    p(y=c\mid\x_i), 
    \quad
    p(y=c\mid\x_i)
    =
    \frac{
    \exp\left(\mathrm{sim}(\vt_i,\vt_c)/\tau\right)
    }{
    \sum_{c'\in\Y_{1:t}}
    \exp\left(\mathrm{sim}(\vt_i,\vt_{c'})/\tau\right)
    },
\end{aligned}
\label{eq:zeroshot-clip}
\end{equation}
where $\tau$ is a temperature parameter. 

\noindent\textbf{Discussions.}
The zero-shot classification makes the textual branch a convenient source for constructing classifiers in CIL, but it also ties the classifier to textual embeddings. 
As discussed in \autoref{sec:intro}, our preliminary diagnosis shows that textual embeddings can be geometrically misaligned with visual class distributions and can hinder optimization when used to initialize the classifier. 
These observations suggest that textual embeddings may be unreliable anchors for the classifier. 
We therefore revisit the problem from a visual classification perspective: once CLIP provides a strong visual encoder, does the classifier still need to rely on the textual branch?
Motivated by this question, we remove the textual branch from the classifier and construct it entirely in the visual space.

\section{\name: Visual Incremental SVM}
\label{sec:method}
To realize this classifier without the textual branch, \name is built on two key components. First, it constructs task-adaptive visual representations by enhancing CLIP's final visual representation with multi-level visual-layer features, while keeping the CLIP visual encoder frozen. Second, it employs a kernelized incremental least-squares SVM, which computes a closed-form solution in the kernel-induced feature space from additive sufficient statistics. When new classes arrive, \name updates these statistics and recomputes the classifier over all seen classes, enabling efficient incremental learning while preserving previously learned class information. All updates are performed entirely in the visual feature space.

\subsection{Task-Adaptive Visual Representation}
\label{sec:method-representation}

To obtain visual representations better suited for downstream CIL, \name does not rely solely on CLIP's final visual feature. Prior studies have shown that useful visual information is distributed across ViT/CLIP layers, and that intermediate features can complement final-layer representations for downstream recognition~\citep{raghu2021vision,ghiasi2022vision,liu2025unveiling}. Motivated by this observation, \name enhances the final visual feature with multi-layer visual cues.
\\{\bf Visual-layer features.}
The frozen CLIP visual encoder consists of $L$ transformer blocks. Given an input image $\x$, we denote the token sequence produced by the $\ell$-th visual block as $\mathbf{H}^{\ell}(\x)=\big[\mathbf{h}^{\ell}_{\mathrm{cls}}(\x), \mathbf{h}^{\ell}_{1}(\x), \dots, \mathbf{h}^{\ell}_{N}(\x)\big]$. We use the CLS token of each block as its visual-layer feature:
\begin{equation}
    \mathbf{h}_{\ell}(\x)
    =
    \mathbf{h}^{\ell}_{\mathrm{cls}}(\x)
    \in \mathbb{R}^{d_v},
    \qquad \ell=1,\dots,L .
    \label{eq:method-layer-feature}
\end{equation}
Here, $\mathbf{h}_{L}(\x)$ is the final-layer visual feature before CLIP's cross-modal projection. For ViT-B/16, this feature is $768$-dimensional. We use these raw visual features without additional normalization.
\\{\bf Residual fusion centered at the final feature.}
To adapt the visual representation to the downstream task distribution, \name learns a residual correction to the final-layer feature from multi-layer visual features. In the full formulation, we concatenate features from all visual blocks:
\begin{equation}
    \mathbf{m}(\x)
    =
    \mathrm{concat}\big(
    \mathbf{h}_{1}(\x),\mathbf{h}_{2}(\x),\dots,\mathbf{h}_{L}(\x)
    \big).
    \label{eq:method-all-layer-concat}
\end{equation}
To transform the concatenated multi-layer features into a residual correction that enhances the final feature, we use a lightweight two-layer MLP as the residual mixer $\mathcal{M}$:
\begin{equation}
    \mathcal{M}\big(\mathbf{m}(\x)\big)
    =
    U\,\mathrm{GELU}\!\left(V\mathbf{m}(\x)+\mathbf{b}_V\right)
    +\mathbf{b}_U,
    \label{eq:method-residual-mixer}
\end{equation}
where $V\in\mathbb{R}^{d_h\times Ld_v}$ and $U\in\mathbb{R}^{d_v\times d_h}$ are the two linear weights, and $\mathbf{b}_V,\mathbf{b}_U$ are the bias terms. 
The enhanced visual representation is then defined as:
\begin{equation}
    \mathbf{u}(\x)
    =
    \mathbf{h}_{L}(\x)
    +
    \mathcal{M}\big(\mathbf{m}(\x)\big).
    \label{eq:method-residual-fusion}
\end{equation}
We zero-initialize the final linear layer of $\mathcal{M}$, i.e., both $U$ and $\mathbf{b}_U$, so the residual branch initially outputs zero and the fused representation starts from $\mathbf{u}(\x)=\mathbf{h}_{L}(\x)$. During base-session training, the mixer learns only a corrective residual from other visual layers.

To make this residual fusion task-adaptive, we train $\mathcal{M}$ only in the base session with an auxiliary linear classifier $g_{\omega}$. 
Given the base-session training set $\mathcal{D}_{1}$, we define the adaptation loss as:
\begin{equation}
    \mathcal{L}_{\mathrm{adapt}}
    =
    \frac{1}{|\mathcal{D}_{1}|}
    \sum_{(\x,y)\in\mathcal{D}_{1}}
    \left[
    \ell_{\mathrm{ce}}
    \big(g_{\omega}(\mathbf{u}(\x)),y\big)
    +
    \lambda_{\mathrm{adapt}}
    \|\mathbf{u}(\x)-\mathbf{h}_{L}(\x)\|_2^2
    \right].
    \label{eq:method-adaptation-loss}
\end{equation}
Here, $\ell_{\mathrm{ce}}$ denotes the cross-entropy loss, and $\lambda_{\mathrm{adapt}}$ controls the identity regularization that keeps the fused representation close to the final-layer visual feature. 
The CLIP visual encoder remains frozen throughout this stage. 
After the base session, $\mathcal{M}$ is fixed and the classifier $g_{\omega}$ is discarded.
\begin{figure*}[t]
\vspace{-4mm}
  \centering
    \includegraphics[width=1\linewidth]{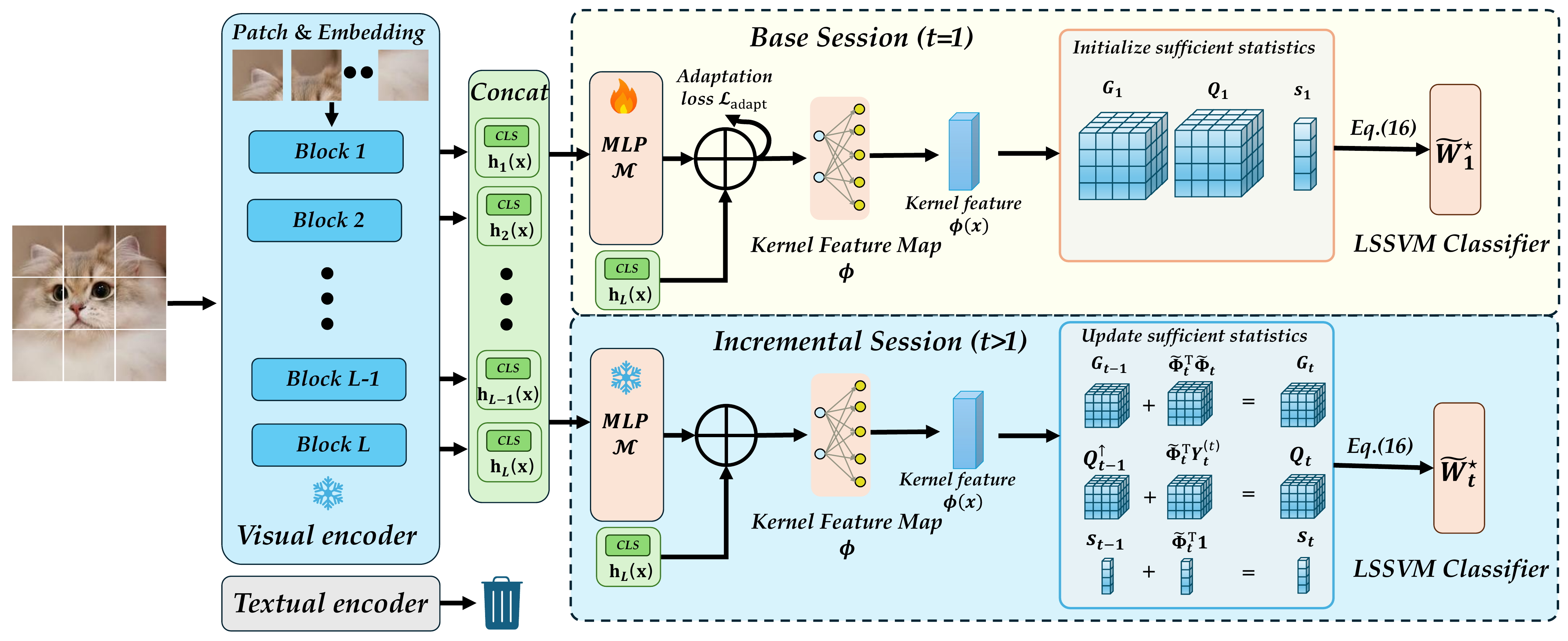}
\caption{
Illustration of \name. 
The classifier is constructed entirely in the visual space: the frozen CLIP visual encoder extracts multi-layer CLS features, and the residual fusion module is trained with the adaptation loss $\mathcal{L}_{\mathrm{adapt}}$. 
Residual fusion and the fixed kernel feature map then produce the kernel feature $\boldsymbol{\phi}(\x)$ for classification. 
In the base session, \name initializes the LS-SVM sufficient statistics $(G_1,Q_1,\mathbf{s}_1)$; in incremental sessions $(t>1)$, it updates $(G_t,Q_t,\mathbf{s}_t)$ with new data and recomputes the classifier weights $\tilde W_t^{\star}$ in closed form.
}
  \label{fig:arch}
  \vspace{-5mm}
\end{figure*}

\subsection{Incremental Kernelized LS-SVM Classifier}
\label{sec:method-lssvm}
With the enhanced visual representation, \name learns an incremental SVM-style classifier that can incorporate new classes without being optimized only on the latest task. 
Several classifier choices are possible in the visual space, such as nearest-class-mean classifiers, gradient-trained linear heads, or ordinary least-squares classifiers. 
We choose LS-SVM because it provides discriminative one-vs-all supervision, naturally supports kernelized nonlinear decision boundaries, and admits a closed-form solution based on additive sufficient statistics. 
These properties make it well suited for exemplar-free CIL, where the classifier should be updated over all seen classes without storing previous samples or repeatedly fine-tuning on only the latest task.
\\{\bf From SVM to kernelized LS-SVM.}
For class $c$, let $y_{i,c}\in\{+1,-1\}$ be the one-vs-all target of sample $\x_i$. 
Given a generic feature mapping $\psi(\x)$, a standard soft-margin SVM learns a separating hyperplane in the corresponding feature space:
\begin{align}
    \min_{\mathbf{w}_{c},b_c,\{\xi_{i,c}\}}
    &\quad
    \frac{1}{2}\|\mathbf{w}_{c}\|_2^2
    +
    C_{\mathrm{svm}}\sum_{i=1}^{N}\xi_{i,c}
    \nonumber\\
    \mathrm{s.t.}
    &\quad
    y_{i,c}
    \left(
    \mathbf{w}_{c}^{\top}\psi(\x_i)+b_c
    \right)
    \geq
    1-\xi_{i,c},
    \qquad
    \xi_{i,c}\geq 0 .
    \label{eq:method-standard-svm}
\end{align}
Directly taking $\psi(\x)=\mathbf{u}(\x)$ gives a linear classifier on the enhanced visual representation. Although $\mathbf{u}(\x)$ is more task-adaptive, visually similar classes may still require nonlinear decision boundaries. We therefore introduce an explicit finite-dimensional approximation to a nonlinear kernel space: a linear classifier in the transformed space corresponds to a nonlinear decision function with respect to $\mathbf{u}(\x)$. This increases classifier capacity while keeping the representation fixed for incremental updates. Instead of forming an implicit kernel matrix, we instantiate $\psi$ with a fixed nonlinear feature map:\looseness -1
\begin{equation}
    \boldsymbol{\phi}(\x)
    =
    \sigma\!\left(R^{\top}\mathbf{u}(\x)\right)
    \in \mathbb{R}^{D},
    \label{eq:method-random-feature}
\end{equation}
where $R\in\mathbb{R}^{d_v\times D}$ is sampled once and then fixed, and $\sigma(\cdot)$ is a nonlinear activation. 
This map induces a nonlinear kernel $k(\x,\x')=\boldsymbol{\phi}(\x)^{\top}\boldsymbol{\phi}(\x')$, so a linear SVM on $\boldsymbol{\phi}(\x)$ corresponds to an SVM in the induced kernel space. 
Because $\boldsymbol{\phi}$ is explicit and fixed, the resulting classifier remains compatible with the closed-form sufficient-statistics update introduced below.
We absorb the bias term by defining:
\begin{equation}
    \tilde{\boldsymbol{\phi}}(\x)
    =
    [\boldsymbol{\phi}(\x);1]
    \in \mathbb{R}^{D+1}.
    \label{eq:method-aug-feature}
\end{equation}
To enable efficient incremental updates, we adopt the LS-SVM relaxation~\citep{suykens1999least}, which replaces the hinge constraints with equality residuals:
\begin{equation}
    \min_{\mathbf{w}_{c},b_c,\{e_{i,c}\}}
    \frac{1}{2}\|\mathbf{w}_{c}\|_2^2
    +
    \frac{C_{\mathrm{svm}}}{2}
    \sum_{i=1}^{N} e_{i,c}^{2}
    \qquad
    \mathrm{s.t.}\quad
    y_{i,c}
    \left(
    \mathbf{w}_{c}^{\top}\boldsymbol{\phi}(\x_i)+b_c
    \right)
    =
    1-e_{i,c}.
    \label{eq:method-lssvm-binary}
\end{equation}
Since $y_{i,c}^{2}=1$, this is equivalent to fitting classifier scores to $\pm1$ targets with squared residuals, while retaining the SVM-style one-vs-all coding.
Stacking all seen classes together, let $\tilde{\Phi}\in\mathbb{R}^{N\times(D+1)}$ be the augmented feature matrix and $Y\in\{-1,+1\}^{N\times C}$ be the one-vs-all target matrix for $C$ seen classes. The multi-class LS-SVM objective becomes:\looseness -1
\begin{equation}
    \min_{\tilde W}
    \frac{1}{2}
    \mathrm{tr}\!\left(\tilde W^{\top}\Gamma \tilde W\right)
    +
    \frac{C_{\mathrm{svm}}}{2}
    \left\|
    \tilde{\Phi}\tilde W - Y
    \right\|_{F}^{2},
    \label{eq:method-lssvm-matrix}
\end{equation}
where $\tilde W\in\mathbb{R}^{(D+1)\times C}$ contains classifier weights and biases, and $\Gamma=\lambda I$ is the regularizer used for all augmented feature dimensions, including the bias term. Setting the derivative to zero yields:
\begin{equation}
    \left(
    \Gamma
    +
    C_{\mathrm{svm}}\tilde{\Phi}^{\top}\tilde{\Phi}
    \right)\tilde W
    =
    C_{\mathrm{svm}}\tilde{\Phi}^{\top}Y ,
    \label{eq:method-normal-equation}
\end{equation}
and the closed-form solution is:
\begin{equation}
    \tilde W^{\star}
    =
    \left(
    \Gamma
    +
    C_{\mathrm{svm}}\tilde{\Phi}^{\top}\tilde{\Phi}
    \right)^{-1}
    \left(
    C_{\mathrm{svm}}\tilde{\Phi}^{\top}Y
    \right).
    \label{eq:method-lssvm-closed}
\end{equation}
This form is suitable for CIL because the solution depends only on feature correlations and feature-label correlations, which can be accumulated task by task.
\\{\bf Additive sufficient statistics.}
Let $\mathcal{C}_{1:t}$ denote all classes observed up to session $t$, and let $C_t=|\mathcal{C}_{1:t}|$. For session $s$, let $\tilde{\Phi}_{s}$ be the augmented feature matrix of its samples. At session $t$, the target matrix of samples from session $s$ is expanded to all current seen classes, denoted by $Y_{s}^{(t)}\in\{-1,+1\}^{N_s\times C_t}$. For each sample, the entry of its ground-truth class is $+1$, and all other entries are $-1$.
\\After session $t$, \name maintains:
\begin{align}
    G_t
    =
    \sum_{s=1}^{t}
    \tilde{\Phi}_{s}^{\top}\tilde{\Phi}_{s}
    \in\mathbb{R}^{(D+1)\times(D+1)},\
    Q_t
    =
    \sum_{s=1}^{t}
    \tilde{\Phi}_{s}^{\top}Y_{s}^{(t)}
    \in\mathbb{R}^{(D+1)\times C_t},\
    \mathbf{s}_t
    =
    \sum_{s=1}^{t}
    \tilde{\Phi}_{s}^{\top}\mathbf{1}
    \in\mathbb{R}^{D+1}.
    \label{eq:method-statistics}
\end{align}
Here, $G_t$ stores feature correlations, $Q_t$ stores feature-label correlations, and $\mathbf{s}_t$ stores the cumulative feature sum. The statistic $\mathbf{s}_t$ is used when new classes are introduced: previous samples should serve as negative examples for these new classes, and their constant $-1$ contribution can be recovered from the stored feature sum without revisiting old data.

Suppose session $t$ introduces $m_t$ new classes. Before adding the new data contribution, we expand $Q_{t-1}$ by appending $m_t$ columns:
\begin{equation}
    Q_{t-1}^{\uparrow}
    =
    \left[
    Q_{t-1},
    -
    \mathbf{s}_{t-1}\mathbf{1}_{m_t}^{\top}
    \right],
    \label{eq:method-q-expand}
\end{equation}
where each appended column represents the negative contribution of all previous samples to a newly introduced class. We then update:
\begin{equation}
    G_t=G_{t-1}+\tilde{\Phi}_{t}^{\top}\tilde{\Phi}_{t},\
    Q_t=Q_{t-1}^{\uparrow}+\tilde{\Phi}_{t}^{\top}Y_{t}^{(t)},\
    \mathbf{s}_t=\mathbf{s}_{t-1}+\tilde{\Phi}_{t}^{\top}\mathbf{1}.
    \label{eq:method-stat-update}
\end{equation}
The classifier for all seen classes is recomputed as:
\begin{equation}
    \tilde W_t^{\star}
    =
    \left(
    \Gamma
    +
    C_{\mathrm{svm}}G_t
    \right)^{-1}
    \left(
    C_{\mathrm{svm}}Q_t
    \right).
    \label{eq:method-incremental-solve}
\end{equation}
{\bf Discussions.} This update is not a expansion that appends weights for new classes. Instead, it recomputes classifier weights for all seen classes from the accumulated normal equations. Thus, the classifier corresponds to fitting an LS-SVM on all seen data represented by sufficient statistics, rather than fine-tuning on only the latest task. By retaining old-class information in the accumulated statistics instead of overwriting it with new-task gradients, this mechanism mitigates catastrophic forgetting.\looseness -1

\subsection{\bf Summary of \textsc{Vis}}
\label{sec:method-summary}
In \name, we construct a visual-only incremental classifier by combining task-adaptive visual representations with closed-form LS-SVM updates. 
The residual fusion module is trained only in the base session with the adaptation loss $\mathcal{L}_{\mathrm{adapt}}$ in Eq.~\ref{eq:method-adaptation-loss}, enhancing CLIP's final visual representation through residual fusion of multi-level visual-layer features. 
The classifier is built in the kernel-induced visual feature space and recomputed from additive sufficient statistics using Eq.~\ref{eq:method-incremental-solve}. 

We also determine all validation-based choices in the base session. 
Specifically, we split the base-session training set $\mathcal{D}_1$ into training and validation subsets. 
The LS-SVM regularization coefficient in $\Gamma$ is selected from a predefined candidate set by minimizing the validation MSE of the one-vs-all LS-SVM targets. 
Although \name is formulated with visual features from all CLIP layers, using every layer is not always necessary: it increases computation and may introduce redundant or less relevant cues. 
Therefore, we use the same validation split to choose a compact visual-layer subset $\mathcal{B}\subseteq\{1,\dots,L\}$ that achieves the highest validation accuracy among candidate subsets. 
When a subset is used, Eq.~\ref{eq:method-all-layer-concat} is replaced by
$\mathbf{m}_{\mathcal{B}}(\x)=\mathrm{concat}_{\ell\in\mathcal{B}}\mathbf{h}_{\ell}(\x)$.
These choices are fixed after the base session and do not use incremental-session data or any testing data.

After the base session, the residual fusion module is fixed, and each incremental session only updates the sufficient statistics $(G_t,Q_t,\mathbf{s}_t)$ and recomputes the LS-SVM weights for all seen classes. 
During inference, \name relies solely on the visual branch. 
Given an image $\x$, we first compute the augmented feature $\tilde{\boldsymbol{\phi}}(\x)$ as in Eq.~\ref{eq:method-aug-feature}, and use the classifier weights $\tilde W_t^{\star}$ obtained by Eq.~\ref{eq:method-incremental-solve}. 
The prediction is:
\begin{equation}
    \hat{y}
    =
    \operatorname*{arg\,max}_{c\in\mathcal{C}_{1:t}}
    \tilde W_t^{\star}[:,c]^{\top}
    \tilde{\boldsymbol{\phi}}(\x).
    \label{eq:method-inference}
\end{equation}

\begin{table*}[t]
	\caption{\small Main benchmark results in terms of average accuracy $\bar{\mathcal{A}}$ and final-session accuracy $\mathcal{A}_B$. The best results are shown in bold. \textbf{All methods start from the same pre-trained CLIP backbone for fair comparison.}}\label{tab:benchmark}
	\vspace{-2mm}
    \setlength{\tabcolsep}{6.5pt}
	\centering
	\resizebox{\textwidth}{!}{%
		\begin{NiceTabular}{@{} l *{12}{c}}
			\toprule
			\multicolumn{1}{c}{\multirow{3}{*}{Method}}
			&
			\multicolumn{4}{c}{Aircraft }   & 
			\multicolumn{4}{c}{CIFAR100 }	&	
			\multicolumn{4}{c}{Cars }   
			\\ 
			& 
			\multicolumn{2}{c}{B0 Inc10}   & 
			\multicolumn{2}{c}{B50 Inc10}	&		
			\multicolumn{2}{c}{B0 Inc10}   & 
			\multicolumn{2}{c}{B50 Inc10}	& 
			\multicolumn{2}{c}{B0 Inc10}   & 
			\multicolumn{2}{c}{B50 Inc10}	 
			\\  
			& 
			{$\bar{\mathcal{A}}$} & ${\mathcal{A}_B}$  
			& {$\bar{\mathcal{A}}$} & ${\mathcal{A}_B}$
			& {$\bar{\mathcal{A}}$} & ${\mathcal{A}_B}$ 
			&  {$\bar{\mathcal{A}}$} & ${\mathcal{A}_B}$  
			& {$\bar{\mathcal{A}}$} & ${\mathcal{A}_B}$
			& {$\bar{\mathcal{A}}$} & ${\mathcal{A}_B}$ 
			\\
			\midrule
            SimpleCIL~\cite{zhou2025revisiting} &59.24 & 48.09 & 53.05 & 48.09 & 84.15 & 76.63& 80.20 & 76.63& 92.04 & 86.85 & 88.96 & 86.85\\
			\rowcolor{gray!10} 
            ACIL~\citep{zhuang2022acil}  &64.99&  56.68&    58.48&  56.71&    89.41&  83.73&    86.56&  83.72&    92.91&  87.48&    89.79&  87.48\\
			DualPrompt~\cite{wang2022dualprompt}  & 44.30& 25.83 &46.07&33.57 & 81.63 & 72.44& 80.12 & 72.57& 76.26 & 62.94& 76.88 & 67.55 \\
            \rowcolor{gray!10}
			CODA-Prompt~\cite{smith2023coda}  & 45.98 & 27.69 & 45.14 & 32.28& 82.43 & 73.43& 78.69 & 71.58& 80.21 & 66.47& 75.06 & 64.19 \\
			RanPAC~\cite{mcdonnell2023ranpac}  & 69.77&60.28&63.83&60.64&89.3&83.18&86.46&83.4&93.67&89.98&91.47&89.96 \\
            \rowcolor{gray!10}
			RAPF~\cite{huang2024class}   &  50.38  & 23.61 &  40.47 &  25.44 & 86.14 & 78.04 & 82.17 &  77.93  & 82.89 & 62.85 &  75.87 & 63.19\\
            CLG-CBM~\citep{yu2025language}  & 66.05   &55.93 &59.25 & 55.39 &  86.58 & 80.15  &83.59  &79.28 &93.25  & 88.76    & 90.11  &88.19 \\
            \rowcolor{gray!10}
            PROOF~\citep{zhou2025learning}  &  63.81  & 56.14 &59.47 & 57.10& 86.77  &79.11   &83.32  & 79.73 &90.74  & 86.51    & 88.00  & 85.58\\
            BOFA~\citep{li2026bofa}  &70.96    &60.43  &66.09 &61.36 &  86.07 &  79.19 &  83.02&  79.44& 94.21 &  90.20 & 92.13  &90.50 \\
            \rowcolor{LightCyan} \name (Ours) & \bf 75.21 & \bf 66.46 & \bf 71.77 & \bf 68.35 & \bf 90.64 & \bf 85.26 & \bf 87.17 & \bf 84.47 & \bf 94.47 & \bf 91.43 & \bf 92.55 & \bf 91.47 \\
		\end{NiceTabular}
	}	
	\resizebox{\textwidth}{!}{%
		\begin{NiceTabular}{@{} l *{12}{c}}
			\toprule
			\multicolumn{1}{c}{\multirow{3}{*}{Method}}
			& 
			\multicolumn{4}{c}{ImageNet-R }   & 
			\multicolumn{4}{c}{CUB }	&	\multicolumn{4}{c}{UCF }   
			\\ 
			& 
			\multicolumn{2}{c}{B0 Inc20}   & 
			\multicolumn{2}{c}{B100 Inc20}	&	\multicolumn{2}{c}{B0 Inc20}   & 
			\multicolumn{2}{c}{B100 Inc20}	& 
			\multicolumn{2}{c}{B0 Inc10}   & 
			\multicolumn{2}{c}{B50 Inc10}	 
			\\  
			& 
			{$\bar{\mathcal{A}}$} & ${\mathcal{A}_B}$  
			& {$\bar{\mathcal{A}}$} & ${\mathcal{A}_B}$
			& {$\bar{\mathcal{A}}$} & ${\mathcal{A}_B}$ 
			&  {$\bar{\mathcal{A}}$} & ${\mathcal{A}_B}$  
			& {$\bar{\mathcal{A}}$} & ${\mathcal{A}_B}$
			& {$\bar{\mathcal{A}}$} & ${\mathcal{A}_B}$ 
			\\
			\midrule
			SimpleCIL~\cite{zhou2025revisiting} & 81.06 & 74.48& 76.84 & 74.48& 83.81 & 77.52& 79.75 & 77.52& 90.44 & 85.68& 88.12 & 85.68\\
            \rowcolor{gray!10}
			 ACIL~\citep{zhuang2022acil}  &82.60&   77.45& 79.86&   77.47& 84.18&   78.75& 80.73&   78.71& 98.71&   97.54& 98.34&   97.54\\
			DualPrompt~\cite{wang2022dualprompt}  &76.21 & 66.65 & 73.22 & 67.58&69.89 &57.46 & 74.40 &64.84 & 85.21 & 75.82& 84.31 & 76.35 \\
            \rowcolor{gray!10}
			CODA-Prompt~\cite{smith2023coda}  & 77.69 & 68.95 & 73.71 & 68.05& 73.12&62.98 &73.95&62.21 & 87.76 & 80.14&83.04 & 75.03 \\
            RanPAC~\cite{mcdonnell2023ranpac}  & 85.19&  79.78&    82.5&   80.23& 86.81&   80.49& 82.87&   79.81& 98.28&   96.82& 98.05&   96.93 \\
            \rowcolor{gray!10}
			RAPF~\cite{huang2024class}  & 81.26  & 70.48 & 76.10 & 70.23 &  79.09 & 62.77& 72.82 & 62.93 & 92.28 & 80.33&90.31 & 81.55\\
            CLG-CBM~\citep{yu2025language}  &  84.64  & 78.50& 81.46& 77.88 &  85.37 & 78.24  &77.74  &76.97 & 95.04&  91.36   & 94.17  &91.85 \\
            \rowcolor{gray!10}
            PROOF~\citep{zhou2025learning}  &83.84    &78.40  &81.20 & 78.92&82.31   & 76.64  &79.20  &76.37  & 94.58 & 91.10    &  93.58 &90.91 \\
            BOFA~\citep{li2026bofa}   & 84.53   &78.77  & 81.60 &79.12 & 86.66  &80.58   & 83.18 & 80.79 & 93.19 &  88.71  &92.60   &89.43 \\
            \rowcolor{LightCyan} \name (Ours) & \bf 86.05 & \bf 80.58 & \bf 82.61 & \bf 80.38 & \bf 87.59 & \bf 82.02 & \bf 84.27 & \bf 81.98 & \bf 99.23 & \bf 98.37 & \bf 99.09 & \bf 98.75 \\
		\end{NiceTabular}
	}
	
	\resizebox{\textwidth}{!}{%
		\begin{NiceTabular}{@{} l *{12}{c}}
			\toprule
			\multicolumn{1}{c}{\multirow{3}{*}{Method}}
			&
			\multicolumn{4}{c}{SUN }   & 
			\multicolumn{4}{c}{Food }	&	\multicolumn{4}{c}{ObjectNet }   
			\\ 
			& 
			\multicolumn{2}{c}{B0 Inc30}   & 
			\multicolumn{2}{c}{B150 Inc30}	&		\multicolumn{2}{c}{B0 Inc10}   & 
			\multicolumn{2}{c}{B50 Inc10}	& 
			\multicolumn{2}{c}{B0 Inc20}   & 
			\multicolumn{2}{c}{B100 Inc20}	 
			\\  
			& 
			{$\bar{\mathcal{A}}$} & ${\mathcal{A}_B}$  
			& {$\bar{\mathcal{A}}$} & ${\mathcal{A}_B}$
			& {$\bar{\mathcal{A}}$} & ${\mathcal{A}_B}$ 
			&  {$\bar{\mathcal{A}}$} & ${\mathcal{A}_B}$  
			& {$\bar{\mathcal{A}}$} & ${\mathcal{A}_B}$
			& {$\bar{\mathcal{A}}$} & ${\mathcal{A}_B}$ 
			\\
			\midrule
			SimpleCIL~\cite{zhou2025revisiting} & 82.13 & 75.58& 78.62 & 75.58& 87.89 & 81.65& 84.73 & 81.65& 52.06 & 40.13& 45.11 & 40.13\\
            \rowcolor{gray!10}
            ACIL~\citep{zhuang2022acil}  &86.99&  80.25&    83.64&  80.24&    90.97&  86.03&    88.38&  86.03&    53.46&  44.78&    49.7&   44.75\\
			DualPrompt~\cite{wang2022dualprompt} & 82.46 & 74.40 & 79.37 & 73.02& 84.92 &77.29& 80.00 & 72.75& 52.62 & 40.72& 49.08 & 42.92 \\
            \rowcolor{gray!10}
			CODA-Prompt~\cite{smith2023coda} &   83.34 & 75.71 & 80.38 & 74.17& 86.18 & 78.78& 80.98 & 74.13& 46.49 & 34.13& 40.57 & 34.13 \\
            RanPAC~\cite{mcdonnell2023ranpac}  & 87.58&  81.75&    84.59&  81.51&    90.56&  85.37&    87.26&  84.52&    57.94&  45.41&    53.52&  48.40 \\
            \rowcolor{gray!10}
			RAPF~\cite{huang2024class}  & 82.13 & 72.47 & 78.04 & 73.10 & 88.57 & 81.15& 85.53 & 81.17&  48.67 & 27.43 & 39.28 &  28.73 \\
            CLG-CBM~\citep{yu2025language}  &  84.85  & 78.09&81.58 & 77.79 &  89.12 &83.05   & 86.76 &83.85 &58.53 &45.11   & 49.80  &43.56 \\
            \rowcolor{gray!10}
            PROOF~\citep{zhou2025learning}  & 83.89   &77.25  & 80.15&76.54 &  90.04 & 84.73  &87.52  & 84.74 & 56.07 &  43.69   & 48.90  & 43.62\\
            BOFA~\citep{li2026bofa}  & 84.38   &77.60  & 81.34 &77.93 &88.14   & 82.08  & 85.97 &  82.84& 59.21 & 46.95    & 51.89  &46.76 \\ 
            \rowcolor{LightCyan} \name (Ours) & \bf 88.73 & \bf 83.00 & \bf 84.91 & \bf 82.02 & \bf 92.06 & \bf 87.52 & \bf 89.23 & \bf 87.09 & \bf 60.80 & \bf 48.13 & \bf 54.49 & \bf 49.03 \\
			\bottomrule
		\end{NiceTabular}
	}
    \vspace{-7mm}
\end{table*}

\begin{figure*}[t] 
\vspace{-4mm}
	\centering
	\begin{subfigure}{0.32\textwidth}		\includegraphics[width=1\columnwidth]{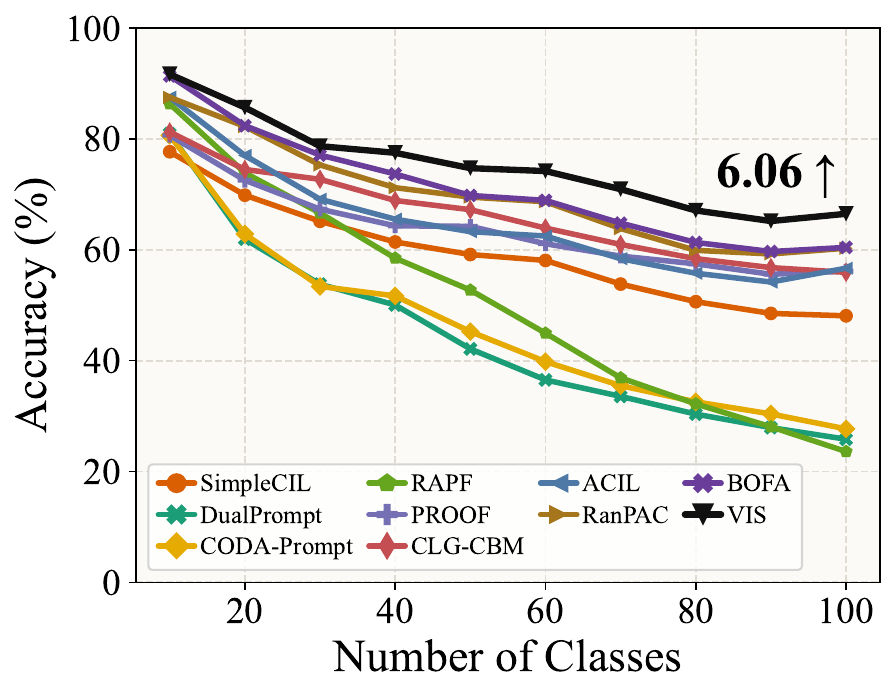}
		\caption{Aircraft Base0 Inc10}
	\end{subfigure}
	\hfill
	\begin{subfigure}{0.32\linewidth}
		\includegraphics[width=1\linewidth]{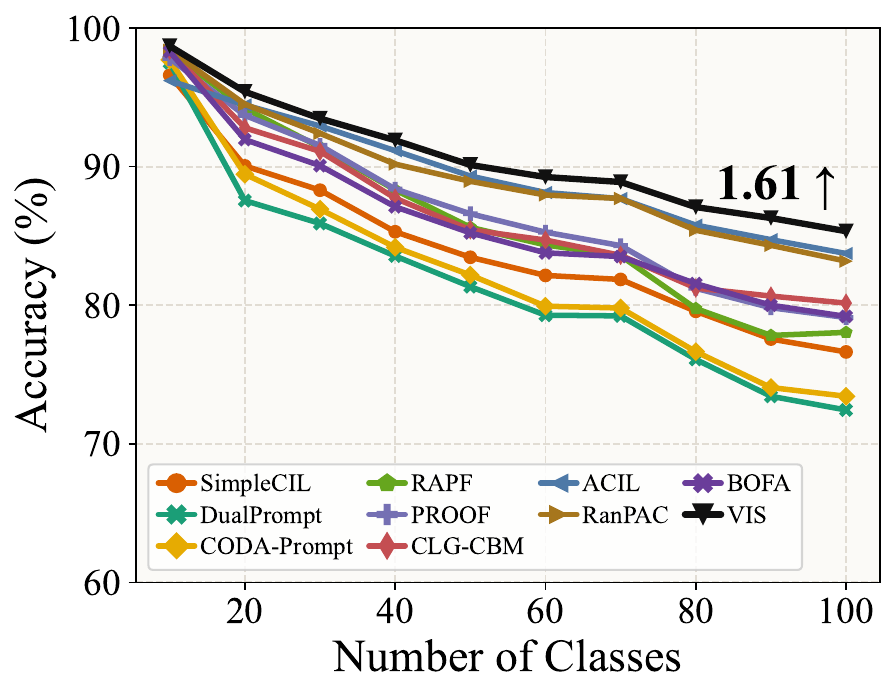}
		\caption{CIFAR100 Base0 Inc10}
	\end{subfigure}
	\hfill
	\begin{subfigure}{0.32\linewidth}
		\includegraphics[width=1\linewidth]{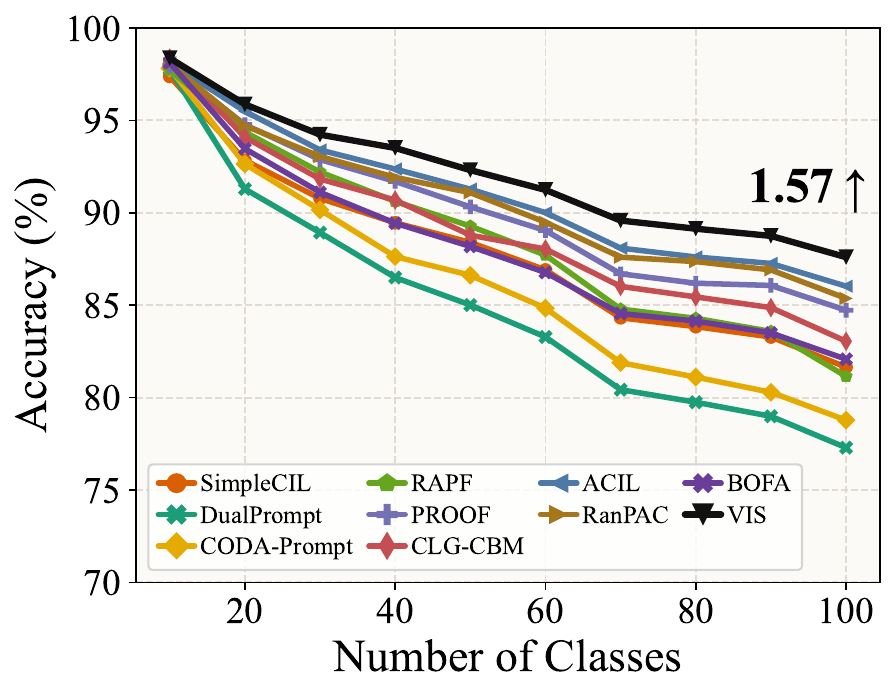}
		\caption{Food Base0 Inc10}
	\end{subfigure}
	\vspace{-2mm}
	\caption{\small Incremental performance of different methods. We report the performance gap after the last incremental stage of \name and the runner-up method at the end of the line. More results are in the supplementary.}
		\vspace{-7mm}
	\label{fig:benchmark}
\end{figure*}

\section{Experiments}
\label{sec:exp}
We evaluate \name on nine CLIP-based CIL benchmarks. We first compare it with state-of-the-art CIL methods, and then conduct ablations and additional analyses to verify the contribution of each component and the reliability of the framework. More results are provided in the appendix.

\subsection{Implementation Details}
\noindent {\bf Dataset.} We follow the evaluation protocol used in recent CLIP-based CIL work~\citep{zhou2025learning,zhou2022learning,wang2022learning} and report results on CIFAR100~\citep{krizhevsky2009learning}, CUB200~\citep{WahCUB2002011}, ObjectNet~\citep{barbu2019objectnet}, ImageNet-R~\citep{hendrycks2021many}, FGVCAircraft~\citep{maji2013fine}, StanfordCars~\citep{krause20133d}, Food101~\citep{bossard2014food}, SUN397~\citep{xiao2010sun}, and UCF101~\citep{soomro2012ucf101}. Following prior work~\citep{zhou2025learning}, we use sampled subsets for practical partitioning: 100 classes each from CIFAR100, FGVCAircraft, StanfordCars, Food101, and UCF101; 200 classes each from CUB200, ObjectNet, and ImageNet-R; and 300 classes from SUN397.
\\ {\bf Task splits.} We use the standard `B-$m$ Inc-$n$' protocol, where $m$ is the number of base-task classes and $n$ is the number of classes introduced in each later task. The class order is shuffled once with seed 1993 and then shared across all methods.
\\ {\bf Comparison methods.} We compare against representative pre-trained model-based CIL baselines, including DualPrompt~\citep{wang2022dualprompt}, CODA-Prompt~\citep{smith2023coda}, and SimpleCIL~\citep{zhou2025revisiting}, as well as CLIP-based methods including RAPF~\citep{huang2024class}, CLG-CBM~\citep{yu2025language}, PROOF~\citep{zhou2025learning}, and BOFA~\citep{li2026bofa}. We also include ACIL~\citep{zhuang2022acil} and RanPAC~\citep{mcdonnell2023ranpac} as representative closed-form classifier baselines. All methods start from the same CLIP ViT-B/16 backbone.
\\ {\bf Training details.}
All experiments are implemented in PyTorch~\citep{paszke2019pytorch} and run on an NVIDIA RTX 4090 GPU. 
Following~\citep{zhou2025external,zhou2025learning}, we use the LAION-400M pre-trained CLIP ViT-B/16 model~\citep{ilharco2021openclip} as the visual backbone for all methods. 
The residual fusion module is trained only in the base session for $5$ epochs using SGD with learning rate $0.01$ and batch size $64$. 
The identity regularization coefficient is set to $\lambda_{\mathrm{adapt}}=0.01$, and the dimension of the kernel-induced feature space is set to $D=15000$.
\\ {\bf Evaluation metric.} Following~\citep{rebuffi2017icarl,zhou2025learning}, we denote the model’s top-1 accuracy after the $b$-th stage by $\mathcal{A}_b$. We report $\mathcal{A}_B$ as the final-stage accuracy and $\bar{\mathcal{A}}=\frac{1}{B}\sum_{b=1}^{B}\mathcal{A}_b$ as the mean accuracy across incremental stages.

\subsection{Benchmark Comparison}
We compare \name with state-of-the-art CIL methods on benchmark datasets, and report the results in Table~\ref{tab:benchmark} and Figure~\ref{fig:benchmark}. As shown, \name consistently achieves state-of-the-art performance, demonstrating strong generalization and robustness across incremental settings. Compared with visual-only closed-form baselines such as ACIL and RanPAC, \name shows clear gains, indicating the effectiveness of its task-adaptive visual representation and incremental LS-SVM classifier. More importantly, \name also surpasses recent CLIP-based methods such as CLG-CBM and BOFA, which retain or exploit the textual branch for classifier construction or cross-modal guidance. These results suggest that the textual branch is not necessary for building a strong incremental classifier, and that a visual-space formulation can achieve superior performance for CLIP-based CIL.

\begin{figure}
\vspace{-4mm}
    \begin{subfigure}[t]{0.32\textwidth}
        \centering
        \includegraphics[width=\linewidth]{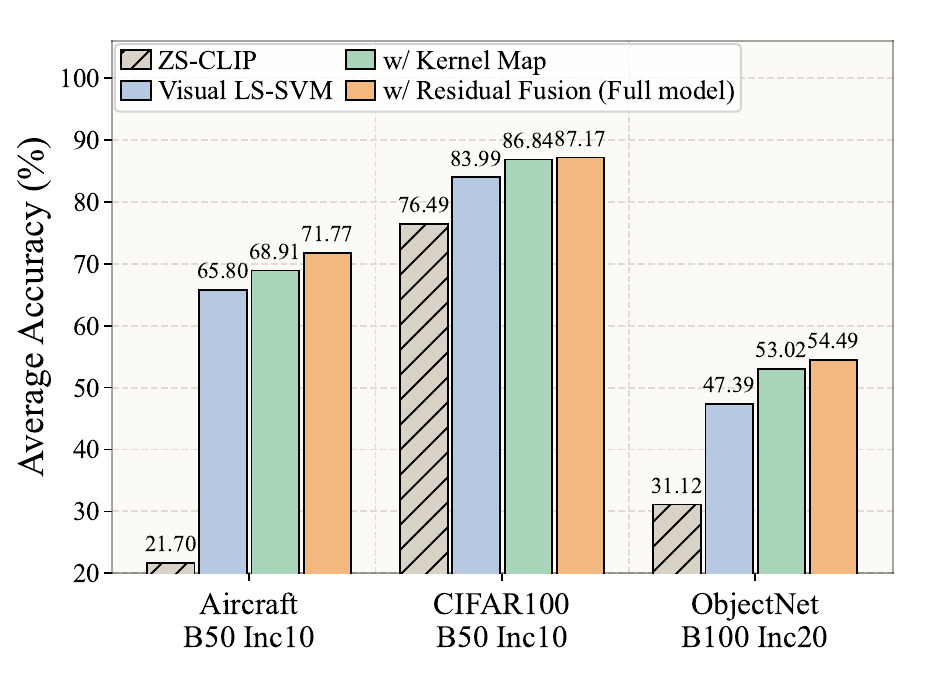}
        \caption{\small Ablation study}
        \label{fig:ablation}
    \end{subfigure}
    \hfill
    \begin{subfigure}[t]{0.33\textwidth}
        \centering
        \includegraphics[width=\linewidth]{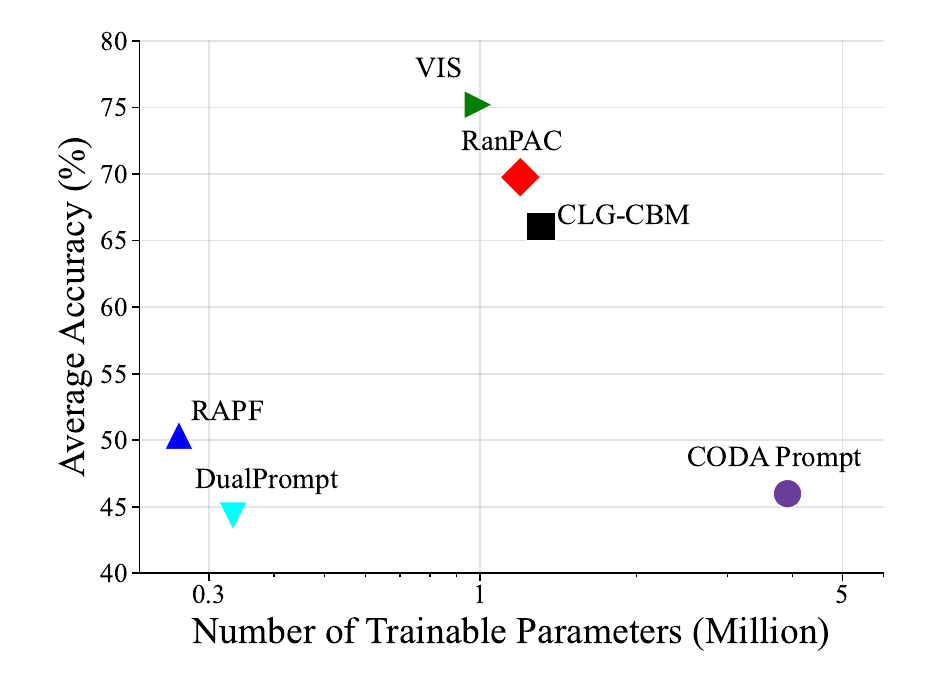}
        \caption{\small Trainable-parameter comparison}
        \label{fig:trainable_params}
    \end{subfigure}
    \hfill
    \begin{subfigure}[t]{0.32\textwidth}
        \centering
        \includegraphics[width=\linewidth]{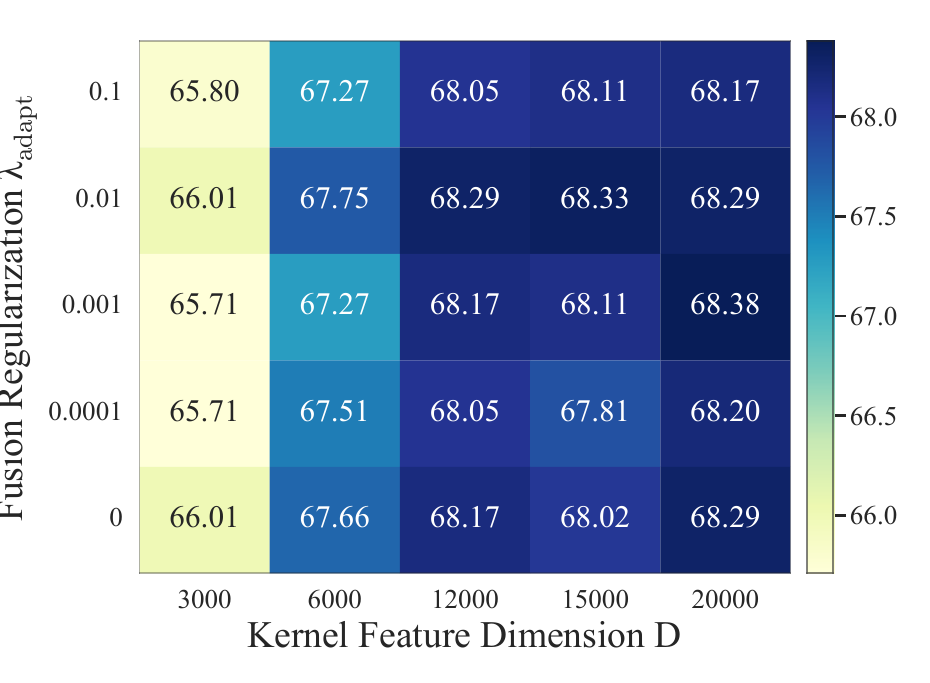}
        \caption{\small Parameter sensitivity}
        \label{fig:sens}
    \end{subfigure}
    \vspace{-3mm}
\caption{
Ablation study, trainable-parameter comparison, and parameter sensitivity.
}
    \vspace{-8mm}
\end{figure}

\subsection{Further Analysis}
\noindent\textbf{Ablation study.}
We conduct component-wise ablations to analyze the contribution of each design in \name, and report the results in Figure~\ref{fig:ablation}. 
We start from \textbf{``ZS-CLIP''}, which uses the original CLIP zero-shot textual head. 
Its performance varies significantly across datasets, especially on Aircraft and ObjectNet, indicating that directly relying on textual embeddings is not always reliable for CIL. 
We then replace the textual head with our visual-space LS-SVM classifier on CLIP's final visual representation, denoted as \textbf{``Visual LS-SVM''}. 
This variant uses neither the kernel map nor residual fusion, and its large improvement shows that constructing the classifier in the visual space is the key factor. 
Based on this visual LS-SVM, we introduce the explicit nonlinear kernel feature map, denoted as \textbf{``w/ Kernel Map''}, which further improves accuracy by increasing classifier capacity. 
Finally, we add the base-session trained residual fusion module, denoted as \textbf{``w/ Residual Fusion (Full)''}. 
The full model achieves the best performance on all datasets, showing the benefit of enhancing CLIP's final visual representation with multi-level visual-layer features.
\\\textbf{Parameter robustness.}
We analyze the sensitivity of \name on Aircraft B0 Inc10 to two main hyperparameters: the adaptation regularization coefficient $\lambda_{\mathrm{adapt}}$ in Eq.~\ref{eq:method-adaptation-loss} and the kernel feature dimension $D$ in Eq.~\ref{eq:method-random-feature}. 
We vary $\lambda_{\mathrm{adapt}}\in\{0,\allowbreak 10^{-4},\allowbreak 10^{-3},\allowbreak 10^{-2},\allowbreak 10^{-1}\}$ 
and $D\in\{3000,\allowbreak 6000,\allowbreak 12000,\allowbreak 15000,\allowbreak 20000\}$, 
and report the final-stage validation accuracy in Figure~\ref{fig:sens}. 
As shown, \name remains stable across a wide range of $\lambda_{\mathrm{adapt}}$ values, indicating that the residual fusion module does not rely on a carefully tuned regularization strength. 
Increasing $D$ generally improves performance until saturation, showing that a sufficiently expressive kernel-induced feature space benefits the LS-SVM classifier. 
These results demonstrate the robustness of \name to different hyperparameter choices.\looseness -1
\\\textbf{Parameter efficiency.}
We compare average accuracy and trainable parameters, \ie, parameters optimized by gradient descent, on Aircraft B0 Inc10 in Figure~\ref{fig:trainable_params}. \name achieves the best accuracy with a compact trainable parameter budget, outperforming prompt-based methods such as DualPrompt and CODA-Prompt, as well as pre-trained feature-based and CLIP-based CIL methods such as RanPAC, RAPF, and CLG-CBM. These results suggest that the gains come from the proposed visual-space formulation rather than a larger trainable parameter budget. In addition, \name maintains non-trainable solver states $(G_t,Q_t,\mathbf{s}_t)$ for closed-form LS-SVM updates. Although these states introduce extra storage for accumulating feature statistics, they are not learnable parameters and do not require storing past images. This exemplar-free property also reduces the risk of exposing raw training data, which helps preserve data privacy in privacy-sensitive scenarios.

\section{Conclusion}
\label{sec:conclusion}
We revisit the role of the textual branch in CLIP-based class-incremental learning. Although CLIP enables zero-shot classification through textual embeddings, our analysis shows that these embeddings can be misaligned with visual class distributions and hinder optimization when used for classifier initialization. We therefore propose \name, a visual-only method that removes the textual branch at deployment and constructs the incremental classifier entirely in the visual space. \name enriches CLIP's final visual representation with multi-level visual features and employs a kernelized incremental LS-SVM whose weights are recomputed from additive sufficient statistics over all seen classes. Extensive experiments demonstrate state-of-the-art performance, showing that strong CLIP-based CIL can be achieved without relying on a deployed textual branch.
\\{\bf Limitations.}
\name fixes the visual representation module and lifted feature space after the base session, which keeps incremental updates stable and efficient but limits further representation adaptation for later tasks. Future work may explore adaptive visual-space updates while preserving the closed-form statistic-based classifier.

\subsection*{AI Use Statement}

In this work, we used generative AI tools for editorial assistance, including checking language, notation consistency, and presentation clarity. We have not used generative AI tools for experimental execution or result generation, and the remaining required disclosure tasks are not applicable to this work. We reviewed and verified all AI-assisted work. We take responsibility for the final content of this work, including text, claims, or artifacts produced with the aid of generative AI.

\bibliography{iclr2027_conference}

\begin{thebibliography}{74}
\providecommand{\natexlab}[1]{#1}
\providecommand{\url}[1]{\texttt{#1}}
\expandafter\ifx\csname urlstyle\endcsname\relax
  \providecommand{\doi}[1]{doi: #1}\else
  \providecommand{\doi}{doi: \begingroup \urlstyle{rm}\Url}\fi

\bibitem[Aljundi et~al.(2018)Aljundi, Babiloni, Elhoseiny, Rohrbach, and
  Tuytelaars]{aljundi2018memory}
Rahaf Aljundi, Francesca Babiloni, Mohamed Elhoseiny, Marcus Rohrbach, and
  Tinne Tuytelaars.
\newblock Memory aware synapses: Learning what (not) to forget.
\newblock In \emph{Proceedings of the European conference on computer vision
  (ECCV)}, pp.\  139--154, 2018.

\bibitem[Barbu et~al.(2019)Barbu, Mayo, Alverio, Luo, Wang, Gutfreund,
  Tenenbaum, and Katz]{barbu2019objectnet}
Andrei Barbu, David Mayo, Julian Alverio, William Luo, Christopher Wang, Dan
  Gutfreund, Josh Tenenbaum, and Boris Katz.
\newblock Objectnet: A large-scale bias-controlled dataset for pushing the
  limits of object recognition models.
\newblock \emph{Advances in neural information processing systems}, 32, 2019.

\bibitem[Bossard et~al.(2014)Bossard, Guillaumin, and
  Van~Gool]{bossard2014food}
Lukas Bossard, Matthieu Guillaumin, and Luc Van~Gool.
\newblock Food-101--mining discriminative components with random forests.
\newblock In \emph{European conference on computer vision}, pp.\  446--461.
  Springer, 2014.

\bibitem[Chaudhry et~al.(2018)Chaudhry, Ranzato, Rohrbach, and
  Elhoseiny]{chaudhry2018efficient}
Arslan Chaudhry, Marc'Aurelio Ranzato, Marcus Rohrbach, and Mohamed Elhoseiny.
\newblock Efficient lifelong learning with a-gem.
\newblock \emph{arXiv preprint arXiv:1812.00420}, 2018.

\bibitem[Cho \& Saul(2009)Cho and Saul]{cho2009kernel}
Youngmin Cho and Lawrence Saul.
\newblock Kernel methods for deep learning.
\newblock \emph{Advances in neural information processing systems}, 22, 2009.

\bibitem[De~Lange et~al.(2021)De~Lange, Aljundi, Masana, Parisot, Jia,
  Leonardis, Slabaugh, and Tuytelaars]{de2021continual}
Matthias De~Lange, Rahaf Aljundi, Marc Masana, Sarah Parisot, Xu~Jia,
  Ale{\v{s}} Leonardis, Gregory Slabaugh, and Tinne Tuytelaars.
\newblock A continual learning survey: Defying forgetting in classification
  tasks.
\newblock \emph{IEEE transactions on pattern analysis and machine
  intelligence}, 44\penalty0 (7):\penalty0 3366--3385, 2021.

\bibitem[Dosovitskiy et~al.(2020)Dosovitskiy, Beyer, Kolesnikov, Weissenborn,
  Zhai, Unterthiner, Dehghani, Minderer, Heigold, Gelly,
  et~al.]{dosovitskiy2020image}
Alexey Dosovitskiy, Lucas Beyer, Alexander Kolesnikov, Dirk Weissenborn,
  Xiaohua Zhai, Thomas Unterthiner, Mostafa Dehghani, Matthias Minderer, Georg
  Heigold, Sylvain Gelly, et~al.
\newblock An image is worth 16x16 words: Transformers for image recognition at
  scale.
\newblock \emph{arXiv preprint arXiv:2010.11929}, 2020.

\bibitem[French(1999)]{french1999catastrophic}
Robert~M French.
\newblock Catastrophic forgetting in connectionist networks.
\newblock \emph{Trends in cognitive sciences}, 3\penalty0 (4):\penalty0
  128--135, 1999.

\bibitem[Fukuda et~al.(2025)Fukuda, Kera, and Kawamoto]{fukuda2025adapter}
Takuma Fukuda, Hiroshi Kera, and Kazuhiko Kawamoto.
\newblock Adapter merging with centroid prototype mapping for scalable
  class-incremental learning.
\newblock In \emph{Proceedings of the computer vision and pattern recognition
  conference}, pp.\  4884--4893, 2025.

\bibitem[Gao et~al.(2022)Gao, Zhao, Ghanem, and Zhang]{gao2022r}
Qiankun Gao, Chen Zhao, Bernard Ghanem, and Jian Zhang.
\newblock R-dfcil: Relation-guided representation learning for data-free class
  incremental learning.
\newblock In \emph{European Conference on Computer Vision}, pp.\  423--439.
  Springer, 2022.

\bibitem[Ghiasi et~al.(2022)Ghiasi, Kazemi, Borgnia, Reich, Shu, Goldblum,
  Wilson, and Goldstein]{ghiasi2022vision}
Amin Ghiasi, Hamid Kazemi, Eitan Borgnia, Steven Reich, Manli Shu, Micah
  Goldblum, Andrew~Gordon Wilson, and Tom Goldstein.
\newblock What do vision transformers learn? a visual exploration.
\newblock \emph{arXiv preprint arXiv:2212.06727}, 2022.

\bibitem[He et~al.(2016)He, Zhang, Ren, and Sun]{he2016deep}
Kaiming He, Xiangyu Zhang, Shaoqing Ren, and Jian Sun.
\newblock Deep residual learning for image recognition.
\newblock In \emph{Proceedings of the IEEE conference on computer vision and
  pattern recognition}, pp.\  770--778, 2016.

\bibitem[Hendrycks et~al.(2021)Hendrycks, Basart, Mu, Kadavath, Wang, Dorundo,
  Desai, Zhu, Parajuli, Guo, et~al.]{hendrycks2021many}
Dan Hendrycks, Steven Basart, Norman Mu, Saurav Kadavath, Frank Wang, Evan
  Dorundo, Rahul Desai, Tyler Zhu, Samyak Parajuli, Mike Guo, et~al.
\newblock The many faces of robustness: A critical analysis of
  out-of-distribution generalization.
\newblock In \emph{Proceedings of the IEEE/CVF international conference on
  computer vision}, pp.\  8340--8349, 2021.

\bibitem[Hinton et~al.(2015)Hinton, Vinyals, and Dean]{hinton2015distilling}
Geoffrey Hinton, Oriol Vinyals, and Jeff Dean.
\newblock Distilling the knowledge in a neural network.
\newblock \emph{arXiv preprint arXiv:1503.02531}, 2015.

\bibitem[Huang et~al.(2024)Huang, Cao, Lu, and Liu]{huang2024class}
Linlan Huang, Xusheng Cao, Haori Lu, and Xialei Liu.
\newblock Class-incremental learning with clip: Adaptive representation
  adjustment and parameter fusion.
\newblock In \emph{European Conference on Computer Vision}, pp.\  214--231.
  Springer, 2024.

\bibitem[Huang et~al.(2025)Huang, Cao, Lu, Meng, Yang, and Liu]{huang2025mind}
Linlan Huang, Xusheng Cao, Haori Lu, Yifan Meng, Fei Yang, and Xialei Liu.
\newblock Mind the gap: Preserving and compensating for the modality gap in
  clip-based continual learning.
\newblock In \emph{Proceedings of the IEEE/CVF International Conference on
  Computer Vision}, pp.\  3777--3786, 2025.

\bibitem[Ilharco et~al.(2021)Ilharco, Wortsman, Carlini, Taori, Dave, Shankar,
  Namkoong, Miller, Hajishirzi, Farhadi, et~al.]{ilharco2021openclip}
Gabriel Ilharco, Mitchell Wortsman, Nicholas Carlini, Rohan Taori, Achal Dave,
  Vaishaal Shankar, Hongseok Namkoong, John Miller, Hannaneh Hajishirzi, Ali
  Farhadi, et~al.
\newblock Openclip.
\newblock \emph{Zenodo}, 2021.

\bibitem[Kirkpatrick et~al.(2017)Kirkpatrick, Pascanu, Rabinowitz, Veness,
  Desjardins, Rusu, Milan, Quan, Ramalho, Grabska-Barwinska,
  et~al.]{kirkpatrick2017overcoming}
James Kirkpatrick, Razvan Pascanu, Neil Rabinowitz, Joel Veness, Guillaume
  Desjardins, Andrei~A Rusu, Kieran Milan, John Quan, Tiago Ramalho, Agnieszka
  Grabska-Barwinska, et~al.
\newblock Overcoming catastrophic forgetting in neural networks.
\newblock \emph{Proceedings of the national academy of sciences}, 114\penalty0
  (13):\penalty0 3521--3526, 2017.

\bibitem[Krause et~al.(2013)Krause, Stark, Deng, and Fei-Fei]{krause20133d}
Jonathan Krause, Michael Stark, Jia Deng, and Li~Fei-Fei.
\newblock 3d object representations for fine-grained categorization.
\newblock In \emph{Proceedings of the IEEE international conference on computer
  vision workshops}, pp.\  554--561, 2013.

\bibitem[Krizhevsky et~al.(2009)Krizhevsky, Hinton,
  et~al.]{krizhevsky2009learning}
Alex Krizhevsky, Geoffrey Hinton, et~al.
\newblock Learning multiple layers of features from tiny images.
\newblock 2009.

\bibitem[Lewandowski et~al.(2025)Lewandowski, Bortkiewicz, Kumar, Gy{\"o}rgy,
  Schuurmans, Ostaszewski, and Machado]{lewandowski2025learning}
Alex Lewandowski, Micha{\l} Bortkiewicz, Saurabh Kumar, Andr{\'a}s Gy{\"o}rgy,
  Dale Schuurmans, Mateusz Ostaszewski, and Marlos~C. Machado.
\newblock Learning continually by spectral regularization.
\newblock In \emph{The International Conference on Learning Representations},
  2025.

\bibitem[Li et~al.(2025)Li, Zhou, Ye, and Zhan]{li2025addressing}
Lan Li, Da-Wei Zhou, Han-Jia Ye, and De-Chuan Zhan.
\newblock Addressing imbalanced domain-incremental learning through
  dual-balance collaborative experts.
\newblock \emph{arXiv preprint arXiv:2507.07100}, 2025.

\bibitem[Li et~al.(2026)Li, Hu, Zhou, Yang, Ye, and Zhan]{li2026bofa}
Lan Li, Tao Hu, Da-Wei Zhou, Jia-Qi Yang, Han-Jia Ye, and De-Chuan Zhan.
\newblock Bofa: Bridge-layer orthogonal low-rank fusion for clip-based
  class-incremental learning.
\newblock In \emph{Proceedings of the AAAI Conference on Artificial
  Intelligence}, volume~40, pp.\  22967--22975, 2026.

\bibitem[Li \& Hoiem(2017)Li and Hoiem]{li2017learning}
Zhizhong Li and Derek Hoiem.
\newblock Learning without forgetting.
\newblock \emph{IEEE transactions on pattern analysis and machine
  intelligence}, 40\penalty0 (12):\penalty0 2935--2947, 2017.

\bibitem[Liang et~al.(2022)Liang, Zhang, Kwon, Yeung, and Zou]{liang2022mind}
Victor~Weixin Liang, Yuhui Zhang, Yongchan Kwon, Serena Yeung, and James~Y Zou.
\newblock Mind the gap: Understanding the modality gap in multi-modal
  contrastive representation learning.
\newblock \emph{Advances in Neural Information Processing Systems},
  35:\penalty0 17612--17625, 2022.

\bibitem[Liu et~al.(2023)Liu, Cao, Lu, Xiao, Bagdanov, and Cheng]{liu2023class}
Xialei Liu, Xusheng Cao, Haori Lu, Jia-wen Xiao, Andrew~D Bagdanov, and
  Ming-Ming Cheng.
\newblock Class incremental learning with pre-trained vision-language models.
\newblock \emph{arXiv preprint arXiv:2310.20348}, 2023.

\bibitem[Liu et~al.(2025)Liu, Wang, Zhang, Liu, and Huang]{liu2025unveiling}
Yajie Liu, Guodong Wang, Jinjin Zhang, Qingjie Liu, and Di~Huang.
\newblock Unveiling the knowledge of clip for training-free open-vocabulary
  semantic segmentation.
\newblock In \emph{Proceedings of the AAAI Conference on Artificial
  Intelligence}, volume~39, pp.\  5649--5657, 2025.

\bibitem[Liu et~al.(2021)Liu, Schiele, and Sun]{liu2021adaptive}
Yaoyao Liu, Bernt Schiele, and Qianru Sun.
\newblock Adaptive aggregation networks for class-incremental learning.
\newblock In \emph{Proceedings of the IEEE/CVF conference on Computer Vision
  and Pattern Recognition}, pp.\  2544--2553, 2021.

\bibitem[Lu et~al.(2025)Lu, Zhang, Moore, Xue, Yao, Hengel, and
  Gong]{lu2025continual}
Haodong Lu, Xinyu Zhang, Kristen Moore, Jason Xue, Lina Yao, Anton van~den
  Hengel, and Dong Gong.
\newblock Continual learning on clip via incremental prompt tuning with
  intrinsic textual anchors.
\newblock \emph{arXiv preprint arXiv:2505.20680}, 2025.

\bibitem[Maji et~al.(2013)Maji, Rahtu, Kannala, Blaschko, and
  Vedaldi]{maji2013fine}
Subhransu Maji, Esa Rahtu, Juho Kannala, Matthew Blaschko, and Andrea Vedaldi.
\newblock Fine-grained visual classification of aircraft.
\newblock \emph{arXiv preprint arXiv:1306.5151}, 2013.

\bibitem[Masana et~al.(2022)Masana, Liu, Twardowski, Menta, Bagdanov, and Van
  De~Weijer]{masana2022class}
Marc Masana, Xialei Liu, Bart{\l}omiej Twardowski, Mikel Menta, Andrew~D
  Bagdanov, and Joost Van De~Weijer.
\newblock Class-incremental learning: survey and performance evaluation on
  image classification.
\newblock \emph{IEEE Transactions on Pattern Analysis and Machine
  Intelligence}, 45\penalty0 (5):\penalty0 5513--5533, 2022.

\bibitem[McDonnell et~al.(2023)McDonnell, Gong, Parvaneh, Abbasnejad, and
  Van~den Hengel]{mcdonnell2023ranpac}
Mark~D McDonnell, Dong Gong, Amin Parvaneh, Ehsan Abbasnejad, and Anton Van~den
  Hengel.
\newblock Ranpac: Random projections and pre-trained models for continual
  learning.
\newblock \emph{Advances in Neural Information Processing Systems},
  36:\penalty0 12022--12053, 2023.

\bibitem[Mensink et~al.(2013)Mensink, Verbeek, Perronnin, and
  Csurka]{mensink2013distance}
Thomas Mensink, Jakob Verbeek, Florent Perronnin, and Gabriela Csurka.
\newblock Distance-based image classification: Generalizing to new classes at
  near-zero cost.
\newblock \emph{IEEE transactions on pattern analysis and machine
  intelligence}, 35\penalty0 (11):\penalty0 2624--2637, 2013.

\bibitem[Momeni et~al.(2025)Momeni, Mazumder, and
  Liu]{DBLP:conf/aaai/MomeniM025}
Saleh Momeni, Sahisnu Mazumder, and Bing Liu.
\newblock Continual learning using a kernel-based method over foundation
  models.
\newblock In \emph{Proceedings of the AAAI Conference on Artificial
  Intelligence}, pp.\  19528--19536, 2025.

\bibitem[Oquab et~al.(2023)Oquab, Darcet, Moutakanni, Vo, Szafraniec, Khalidov,
  Fernandez, Haziza, Massa, El-Nouby, et~al.]{oquab2023dinov2}
Maxime Oquab, Timoth{\'e}e Darcet, Th{\'e}o Moutakanni, Huy Vo, Marc
  Szafraniec, Vasil Khalidov, Pierre Fernandez, Daniel Haziza, Francisco Massa,
  Alaaeldin El-Nouby, et~al.
\newblock Dinov2: Learning robust visual features without supervision.
\newblock \emph{arXiv preprint arXiv:2304.07193}, 2023.

\bibitem[Ostapenko et~al.(2019)Ostapenko, Puscas, Klein, Jahnichen, and
  Nabi]{ostapenko2019learning}
Oleksiy Ostapenko, Mihai Puscas, Tassilo Klein, Patrick Jahnichen, and Moin
  Nabi.
\newblock Learning to remember: A synaptic plasticity driven framework for
  continual learning.
\newblock In \emph{Proceedings of the IEEE/CVF conference on computer vision
  and pattern recognition}, pp.\  11321--11329, 2019.

\bibitem[Paszke et~al.(2019)Paszke, Gross, Massa, Lerer, Bradbury, Chanan,
  Killeen, Lin, Gimelshein, Antiga, et~al.]{paszke2019pytorch}
Adam Paszke, Sam Gross, Francisco Massa, Adam Lerer, James Bradbury, Gregory
  Chanan, Trevor Killeen, Zeming Lin, Natalia Gimelshein, Luca Antiga, et~al.
\newblock Pytorch: An imperative style, high-performance deep learning library.
\newblock \emph{Advances in neural information processing systems}, 32, 2019.

\bibitem[Peng et~al.(2025)Peng, Elenter, Agterberg, Ribeiro, and
  Vidal]{Peng-ICLR2025}
Liangzu Peng, Juan Elenter, Joshua Agterberg, Alejandro Ribeiro, and Rene
  Vidal.
\newblock {LoRanPAC}: Low-rank random features and pre-trained models for
  bridging theory and practice in continual learning.
\newblock In \emph{The International Conference on Learning Representations},
  2025.

\bibitem[Qi et~al.(2025)Qi, Zhou, Yao, Ye, and Zhan]{qi2025adaptive}
Zhi-Hong Qi, Da-Wei Zhou, Yiran Yao, Han-Jia Ye, and De-Chuan Zhan.
\newblock Adaptive adapter routing for long-tailed class-incremental learning.
\newblock \emph{Machine Learning}, 114\penalty0 (3):\penalty0 68, 2025.

\bibitem[Radford et~al.(2021)Radford, Kim, Hallacy, Ramesh, Goh, Agarwal,
  Sastry, Askell, Mishkin, Clark, et~al.]{radford2021learning}
Alec Radford, Jong~Wook Kim, Chris Hallacy, Aditya Ramesh, Gabriel Goh,
  Sandhini Agarwal, Girish Sastry, Amanda Askell, Pamela Mishkin, Jack Clark,
  et~al.
\newblock Learning transferable visual models from natural language
  supervision.
\newblock In \emph{International conference on machine learning}, pp.\
  8748--8763. PmLR, 2021.

\bibitem[Raghu et~al.(2021)Raghu, Unterthiner, Kornblith, Zhang, and
  Dosovitskiy]{raghu2021vision}
Maithra Raghu, Thomas Unterthiner, Simon Kornblith, Chiyuan Zhang, and Alexey
  Dosovitskiy.
\newblock Do vision transformers see like convolutional neural networks?
\newblock \emph{Advances in neural information processing systems},
  34:\penalty0 12116--12128, 2021.

\bibitem[Rahimi \& Recht(2007)Rahimi and Recht]{rahimi2007random}
Ali Rahimi and Benjamin Recht.
\newblock Random features for large-scale kernel machines.
\newblock \emph{Advances in neural information processing systems}, 20, 2007.

\bibitem[Rebuffi et~al.(2017)Rebuffi, Kolesnikov, Sperl, and
  Lampert]{rebuffi2017icarl}
Sylvestre-Alvise Rebuffi, Alexander Kolesnikov, Georg Sperl, and Christoph~H
  Lampert.
\newblock icarl: Incremental classifier and representation learning.
\newblock In \emph{Proceedings of the IEEE conference on Computer Vision and
  Pattern Recognition}, pp.\  2001--2010, 2017.

\bibitem[Serra et~al.(2018)Serra, Suris, Miron, and
  Karatzoglou]{serra2018overcoming}
Joan Serra, Didac Suris, Marius Miron, and Alexandros Karatzoglou.
\newblock Overcoming catastrophic forgetting with hard attention to the task.
\newblock In \emph{International conference on machine learning}, pp.\
  4548--4557. PMLR, 2018.

\bibitem[Shi et~al.(2021)Shi, Chen, Zhang, Zhan, and Wu]{shi2021overcoming}
Guangyuan Shi, Jiaxin Chen, Wenlong Zhang, Li-Ming Zhan, and Xiao-Ming Wu.
\newblock Overcoming catastrophic forgetting in incremental few-shot learning
  by finding flat minima.
\newblock \emph{Advances in neural information processing systems},
  34:\penalty0 6747--6761, 2021.

\bibitem[Smith et~al.(2023)Smith, Karlinsky, Gutta, Cascante-Bonilla, Kim,
  Arbelle, Panda, Feris, and Kira]{smith2023coda}
James~Seale Smith, Leonid Karlinsky, Vyshnavi Gutta, Paola Cascante-Bonilla,
  Donghyun Kim, Assaf Arbelle, Rameswar Panda, Rogerio Feris, and Zsolt Kira.
\newblock Coda-prompt: Continual decomposed attention-based prompting for
  rehearsal-free continual learning.
\newblock In \emph{Proceedings of the IEEE/CVF conference on computer vision
  and pattern recognition}, pp.\  11909--11919, 2023.

\bibitem[Soomro et~al.(2012)Soomro, Zamir, and Shah]{soomro2012ucf101}
Khurram Soomro, Amir~Roshan Zamir, and Mubarak Shah.
\newblock Ucf101: A dataset of 101 human actions classes from videos in the
  wild.
\newblock \emph{arXiv preprint arXiv:1212.0402}, 2012.

\bibitem[Sun \& Zhou(2026)Sun and Zhou]{sun2026c3box}
Hao Sun and Da-Wei Zhou.
\newblock C3box: A clip-based class-incremental learning toolbox.
\newblock \emph{arXiv preprint arXiv:2601.20852}, 2026.

\bibitem[Suykens \& Vandewalle(1999)Suykens and Vandewalle]{suykens1999least}
Johan~AK Suykens and Joos Vandewalle.
\newblock Least squares support vector machine classifiers.
\newblock \emph{Neural processing letters}, 9\penalty0 (3):\penalty0 293--300,
  1999.

\bibitem[Wah et~al.(2011)Wah, Branson, Welinder, Perona, and
  Belongie]{WahCUB2002011}
C.~Wah, S.~Branson, P.~Welinder, P.~Perona, and S.~Belongie.
\newblock {The Caltech-UCSD Birds-200-2011 Dataset}.
\newblock Technical Report CNS-TR-2011-001, California Institute of Technology,
  2011.

\bibitem[Wang et~al.(2022{\natexlab{a}})Wang, Zhou, Liu, Ye, Bian, Zhan, and
  Zhao]{wang2022beef}
Fu-Yun Wang, Da-Wei Zhou, Liu Liu, Han-Jia Ye, Yatao Bian, De-Chuan Zhan, and
  Peilin Zhao.
\newblock Beef: Bi-compatible class-incremental learning via energy-based
  expansion and fusion.
\newblock In \emph{The eleventh international conference on learning
  representations}, 2022{\natexlab{a}}.

\bibitem[Wang et~al.(2023)Wang, Duan, Kang, Liu, Lin, Xu, L{\"u}, and
  Zhang]{wang2023attriclip}
Runqi Wang, Xiaoyue Duan, Guoliang Kang, Jianzhuang Liu, Shaohui Lin, Songcen
  Xu, Jinhu L{\"u}, and Baochang Zhang.
\newblock Attriclip: A non-incremental learner for incremental knowledge
  learning.
\newblock In \emph{Proceedings of the IEEE/CVF Conference on Computer Vision
  and Pattern Recognition}, pp.\  3654--3663, 2023.

\bibitem[Wang et~al.(2022{\natexlab{b}})Wang, Zhang, Ebrahimi, Sun, Zhang, Lee,
  Ren, Su, Perot, Dy, et~al.]{wang2022dualprompt}
Zifeng Wang, Zizhao Zhang, Sayna Ebrahimi, Ruoxi Sun, Han Zhang, Chen-Yu Lee,
  Xiaoqi Ren, Guolong Su, Vincent Perot, Jennifer Dy, et~al.
\newblock Dualprompt: Complementary prompting for rehearsal-free continual
  learning.
\newblock In \emph{European conference on computer vision}, pp.\  631--648.
  Springer, 2022{\natexlab{b}}.

\bibitem[Wang et~al.(2022{\natexlab{c}})Wang, Zhang, Lee, Zhang, Sun, Ren, Su,
  Perot, Dy, and Pfister]{wang2022learning}
Zifeng Wang, Zizhao Zhang, Chen-Yu Lee, Han Zhang, Ruoxi Sun, Xiaoqi Ren,
  Guolong Su, Vincent Perot, Jennifer Dy, and Tomas Pfister.
\newblock Learning to prompt for continual learning.
\newblock In \emph{Proceedings of the IEEE/CVF conference on computer vision
  and pattern recognition}, pp.\  139--149, 2022{\natexlab{c}}.

\bibitem[Xiang et~al.(2019)Xiang, Fu, Ji, and Huang]{xiang2019incremental}
Ye~Xiang, Ying Fu, Pan Ji, and Hua Huang.
\newblock Incremental learning using conditional adversarial networks.
\newblock In \emph{Proceedings of the IEEE/CVF international conference on
  computer vision}, pp.\  6619--6628, 2019.

\bibitem[Xiao et~al.(2010)Xiao, Hays, Ehinger, Oliva, and
  Torralba]{xiao2010sun}
Jianxiong Xiao, James Hays, Krista~A Ehinger, Aude Oliva, and Antonio Torralba.
\newblock Sun database: Large-scale scene recognition from abbey to zoo.
\newblock In \emph{2010 IEEE computer society conference on computer vision and
  pattern recognition}, pp.\  3485--3492. IEEE, 2010.

\bibitem[Xie et~al.(2026)Xie, Tang, Shi, Ye, Zhan, and Zhou]{Xie2026SAME}
Zhen-Hao Xie, Jun-Tao Tang, Yu-Cheng Shi, Han-Jia Ye, De-Chuan Zhan, and Da-Wei
  Zhou.
\newblock Same: Stabilized mixture-of-experts for multimodal continual
  instruction tuning.
\newblock \emph{arXiv preprint arXiv:2602.01990}, 2026.

\bibitem[Xu \& Zhu(2018)Xu and Zhu]{xu2018reinforced}
Ju~Xu and Zhanxing Zhu.
\newblock Reinforced continual learning.
\newblock \emph{Advances in neural information processing systems}, 31, 2018.

\bibitem[Yoon et~al.(2017)Yoon, Yang, Lee, and Hwang]{yoon2017lifelong}
Jaehong Yoon, Eunho Yang, Jeongtae Lee, and Sung~Ju Hwang.
\newblock Lifelong learning with dynamically expandable networks.
\newblock \emph{arXiv preprint arXiv:1708.01547}, 2017.

\bibitem[Yu et~al.(2024)Yu, Zhuge, Zhang, Hu, Wang, Lu, and He]{yu2024boosting}
Jiazuo Yu, Yunzhi Zhuge, Lu~Zhang, Ping Hu, Dong Wang, Huchuan Lu, and You He.
\newblock Boosting continual learning of vision-language models via
  mixture-of-experts adapters.
\newblock In \emph{Proceedings of the IEEE/CVF Conference on Computer Vision
  and Pattern Recognition}, pp.\  23219--23230, 2024.

\bibitem[Yu et~al.(2025)Yu, Han, Tao, Yao, and Xu]{yu2025language}
Lu~Yu, Haoyu Han, Zhe Tao, Hantao Yao, and Changsheng Xu.
\newblock Language guided concept bottleneck models for interpretable continual
  learning.
\newblock In \emph{Proceedings of the Computer Vision and Pattern Recognition
  Conference}, pp.\  14976--14986, 2025.

\bibitem[Zenke et~al.(2017)Zenke, Poole, and Ganguli]{zenke2017continual}
Friedemann Zenke, Ben Poole, and Surya Ganguli.
\newblock Continual learning through synaptic intelligence.
\newblock In \emph{International conference on machine learning}, pp.\
  3987--3995. Pmlr, 2017.

\bibitem[Zhai et~al.(2023)Zhai, Mustafa, Kolesnikov, and
  Beyer]{zhai2023sigmoid}
Xiaohua Zhai, Basil Mustafa, Alexander Kolesnikov, and Lucas Beyer.
\newblock Sigmoid loss for language image pre-training.
\newblock In \emph{2023 IEEE/CVF International Conference on Computer Vision
  (ICCV)}, pp.\  11941--11952. IEEE, 2023.

\bibitem[Zhang et~al.(2025)Zhang, Yu, Wang, Xie, Trucco, Zheng, and
  Yang]{zhang2025visual}
Wentao Zhang, Tong Yu, Ruixuan Wang, Jianhui Xie, Emanuele Trucco, Wei-Shi
  Zheng, and Xiaobo Yang.
\newblock Visual class incremental learning with textual priors guidance based
  on an adapted vision-language model.
\newblock \emph{IEEE Transactions on Multimedia}, 2025.

\bibitem[Zhao et~al.(2020)Zhao, Xiao, Gan, Zhang, and Xia]{zhao2020maintaining}
Bowen Zhao, Xi~Xiao, Guojun Gan, Bin Zhang, and Shu-Tao Xia.
\newblock Maintaining discrimination and fairness in class incremental
  learning.
\newblock In \emph{Proceedings of the IEEE/CVF conference on computer vision
  and pattern recognition}, pp.\  13208--13217, 2020.

\bibitem[Zhou et~al.(2024)Zhou, Wang, Qi, Ye, Zhan, and Liu]{zhou2024class}
Da-Wei Zhou, Qi-Wei Wang, Zhi-Hong Qi, Han-Jia Ye, De-Chuan Zhan, and Ziwei
  Liu.
\newblock Class-incremental learning: A survey.
\newblock \emph{IEEE Transactions on Pattern Analysis and Machine
  Intelligence}, 46\penalty0 (12):\penalty0 9851--9873, 2024.

\bibitem[Zhou et~al.(2025{\natexlab{a}})Zhou, Cai, Ye, Zhan, and
  Liu]{zhou2025revisiting}
Da-Wei Zhou, Zi-Wen Cai, Han-Jia Ye, De-Chuan Zhan, and Ziwei Liu.
\newblock Revisiting class-incremental learning with pre-trained models:
  Generalizability and adaptivity are all you need.
\newblock \emph{International Journal of Computer Vision}, 133\penalty0
  (3):\penalty0 1012--1032, 2025{\natexlab{a}}.

\bibitem[Zhou et~al.(2025{\natexlab{b}})Zhou, Li, Ning, Ye, Zhang, and
  Zhan]{zhou2025external}
Da-Wei Zhou, Kai-Wen Li, Jingyi Ning, Han-Jia Ye, Lijun Zhang, and De-Chuan
  Zhan.
\newblock External knowledge injection for clip-based class-incremental
  learning.
\newblock In \emph{Proceedings of the IEEE/CVF International Conference on
  Computer Vision}, pp.\  3314--3325, 2025{\natexlab{b}}.

\bibitem[Zhou et~al.(2025{\natexlab{c}})Zhou, Zhang, Wang, Ning, Ye, Zhan, and
  Liu]{zhou2025learning}
Da-Wei Zhou, Yuanhan Zhang, Yan Wang, Jingyi Ning, Han-Jia Ye, De-Chuan Zhan,
  and Ziwei Liu.
\newblock Learning without forgetting for vision-language models.
\newblock \emph{IEEE Transactions on Pattern Analysis and Machine
  Intelligence}, 47\penalty0 (6):\penalty0 4489--4504, 2025{\natexlab{c}}.

\bibitem[Zhou et~al.(2022)Zhou, Yang, Loy, and Liu]{zhou2022learning}
Kaiyang Zhou, Jingkang Yang, Chen~Change Loy, and Ziwei Liu.
\newblock Learning to prompt for vision-language models.
\newblock \emph{International journal of computer vision}, 130\penalty0
  (9):\penalty0 2337--2348, 2022.

\bibitem[Zhu et~al.(2021)Zhu, Zhang, Wang, Yin, and Liu]{zhu2021prototype}
Fei Zhu, Xu-Yao Zhang, Chuang Wang, Fei Yin, and Cheng-Lin Liu.
\newblock Prototype augmentation and self-supervision for incremental learning.
\newblock In \emph{Proceedings of the IEEE/CVF conference on computer vision
  and pattern recognition}, pp.\  5871--5880, 2021.

\bibitem[Zhuang et~al.(2022)Zhuang, Weng, Wei, Xie, Toh, and
  Lin]{zhuang2022acil}
Huiping Zhuang, Zhenyu Weng, Hongxin Wei, Renchunzi Xie, Kar-Ann Toh, and
  Zhiping Lin.
\newblock Acil: Analytic class-incremental learning with absolute memorization
  and privacy protection.
\newblock \emph{Advances in Neural Information Processing Systems},
  35:\penalty0 11602--11614, 2022.

\bibitem[Zhuang et~al.(2023)Zhuang, Weng, He, Lin, and Zeng]{zhuang2023gkeal}
Huiping Zhuang, Zhenyu Weng, Run He, Zhiping Lin, and Ziqian Zeng.
\newblock Gkeal: Gaussian kernel embedded analytic learning for few-shot class
  incremental task.
\newblock In \emph{Proceedings of the IEEE/CVF conference on computer vision
  and pattern recognition}, pp.\  7746--7755, 2023.

\bibitem[Zhuang et~al.(2024)Zhuang, He, Tong, Zeng, Chen, and
  Lin]{zhuang2024ds}
Huiping Zhuang, Run He, Kai Tong, Ziqian Zeng, Cen Chen, and Zhiping Lin.
\newblock Ds-al: A dual-stream analytic learning for exemplar-free
  class-incremental learning.
\newblock In \emph{Proceedings of the AAAI Conference on Artificial
  Intelligence}, volume~38, pp.\  17237--17244, 2024.

\end{thebibliography}
\bibliographystyle{iclr2027_conference}

\clearpage
\appendix
\section*{\centering Appendix}
\renewcommand{\thesection}{\Alph{section}}
In this appendix, we provide additional details and results for \name, including the algorithmic procedure, implementation details, supplementary analyses, complete benchmark curves, and broader-impact discussion.

\noindent\textbf{Section~\ref{sec:alg-summary}} summarizes the task-wise update of \name, including residual fusion, fixed kernel feature mapping, sufficient-statistics updates, and the closed-form LS-SVM classifier recomputation.

\noindent\textbf{Section~\ref{sec:suppl-impl}} provides additional implementation details and further discussion on the LS-SVM classifier and kernel feature map.

\noindent\textbf{Section~\ref{sec:app_more_results}} presents supplementary analyses of the robustness, generalization, efficiency, and forgetting behavior of \name.

\noindent\textbf{Section~\ref{sec:suppl-full-results}} provides the complete incremental learning curves under the zero-base and half-base settings.

\noindent\textbf{Section~\ref{sec:suppl-compared-methods}} describes the compared methods used in our benchmark.

\noindent\textbf{Section~\ref{sec:suppl-classifier-choice}} studies the effect of the final classifier under fixed kernel-induced visual features, comparing NCM, Ridge, and LS-SVM.

\noindent\textbf{Section~\ref{sec:suppl-linear-comparison}} compares \name with related classifier-based CIL methods, particularly ACIL and RanPAC, from both methodological and empirical perspectives.

\noindent\textbf{Section~\ref{sec:suppl-pre}} provides additional preliminary experiments that further support our diagnosis of textual classifier weights, including modality-gap analysis, optimization dynamics, cross-class gradient statistics, and t-SNE visualizations.

\section{Algorithmic Summary}
\label{sec:alg-summary}

Algorithm~\ref{alg:vocal} summarizes the task-wise update of \name. 
For compactness, we denote by 
$\mathrm{Lift}(\D_t;\Phi_v,\mathcal{B},\mathcal{M},R)$ 
the frozen visual pipeline that extracts selected visual-layer CLS features, applies residual fusion, computes the fixed kernel feature map, and returns the augmented feature matrix $\tilde{\Phi}_t$. 
The layer subset $\mathcal{B}$ and LS-SVM regularizer are selected on the base-session split as described in \autoref{sec:method-summary}.

\begin{algorithm}[t]
\small
\caption{\name task-wise update.}
\label{alg:vocal}
\raggedright
\textbf{Input:} Task data $\D_t$; frozen CLIP visual encoder $\Phi_v$; selected layers $\mathcal{B}$; fusion module $\mathcal{M}$; fixed projection $R$; statistics $(G,Q,\mathbf{s})$; LS-SVM parameters $C_{\mathrm{svm}},\Gamma$.\\
\textbf{Output:} Updated statistics $(G,Q,\mathbf{s})$ and classifier $\tilde W_t^\star$.
\begin{algorithmic}[1]
\IF{$t=1$}
    \STATE Train $\mathcal{M}$ with $\mathcal{L}_{\mathrm{adapt}}$ in Eq.~\ref{eq:method-adaptation-loss}, then freeze it;
    \STATE Initialize $G\leftarrow0,\ Q\leftarrow[\,],\ \mathbf{s}\leftarrow0$;
\ENDIF
\STATE Extract augmented lifted features:
\[
    \tilde{\Phi}_t
    \leftarrow
    \mathrm{Lift}(\D_t;\Phi_v,\mathcal{B},\mathcal{M},R).
\]
\STATE Expand $Q$ for the new classes:
\[
    Q
    \leftarrow
    \big[Q,\ -\mathbf{s}\mathbf{1}_{|\Y_t|}^{\top}\big].
\]
\STATE Build $Y_t^{(t)}\in\{-1,+1\}^{N_t\times C_t}$ over $\Y_{1:t}$, with $+1$ for the ground-truth class and $-1$ otherwise;
\STATE Update sufficient statistics:
\[
    G\leftarrow G+\tilde{\Phi}_t^\top\tilde{\Phi}_t,\quad
    Q\leftarrow Q+\tilde{\Phi}_t^\top Y_t^{(t)},\quad
    \mathbf{s}\leftarrow\mathbf{s}+\tilde{\Phi}_t^\top\mathbf{1}.
\]
\STATE Recompute the classifier:
\[
    \tilde W_t^\star
    \leftarrow
    (\Gamma+C_{\mathrm{svm}}G)^{-1}(C_{\mathrm{svm}}Q).
\]
\STATE Predict by
\[
    \hat y(\x)=
    \operatorname*{arg\,max}_{c\in\Y_{1:t}}
    \tilde W_t^\star[:,c]^\top\tilde{\boldsymbol{\phi}}(\x).
\]
\RETURN $(G,Q,\mathbf{s})$ and $\tilde W_t^\star$.
\end{algorithmic}
\end{algorithm}

\section{More Implementation and Reproducibility Details}
\label{sec:suppl-impl}
We provide implementation details that are omitted from the main text for clarity. 
For the base-session search of the LS-SVM regularizer $\lambda$, we use the following $16$ candidate values:
$\{10^{-6},10^{-5},10^{-4},3\!\cdot\!10^{-4},10^{-3},3\!\cdot\!10^{-3},10^{-2},3\!\cdot\!10^{-2},10^{-1},3\!\cdot\!10^{-1},1,3,10,100,10^3,10^4\}$.
We set $C_{\mathrm{svm}}=10$ in all experiments; since it only changes the relative scale between the regularization and data-fitting terms, its effect can be absorbed into the selected regularization coefficient $\lambda$.
For the fixed nonlinear feature map in Eq.~\ref{eq:method-random-feature}, we use ReLU as $\sigma(\cdot)$, and the projection matrix $R$ is randomly initialized once and kept fixed throughout all sessions.
For the residual mixer, the hidden dimension is set to $d_h=256$.
For efficient layer subset selection, we evaluate candidate subsets using the same base-session train/validation split described in the main text.
For the CLIP ViT-B/16 backbone used in our experiments, the selected subset is $\mathcal{B}=\{6,8,10,12\}$, which is fixed for all incremental sessions.
Both $\lambda$ and $\mathcal{B}$ are selected only from the base-session training data and are never tuned on incremental-session data or test data.
All datasets and pre-trained models used in this paper are publicly available research assets, and we use them only for academic evaluation while following their official terms of use and licenses.

\subsection{Remark on the LS-SVM Classifier}
\label{sec:suppl-lssvm}

We adopt the least-squares SVM formulation~\citep{suykens1999least} for the visual-space classifier in \name. 
Although the final objective has a squared-error form, it should not be interpreted as ordinary regression on class indices or one-hot labels. 
Instead, each class is learned in a one-vs-all manner with $\pm1$ coding, where the ground-truth class is assigned $+1$ and all other seen classes are assigned $-1$. 
This coding preserves the SVM-style positive-versus-negative supervision, while the least-squares relaxation replaces hinge constraints with equality residuals and yields a closed-form solution. 
Another advantage of the SVM formulation is its \emph{natural compatibility with kernel methods}. 
When an SVM is written on a feature map $\psi(\x)$, its decision function can be formulated through inner products in the corresponding feature space, which induces a kernelized classifier. 
In our case, the explicit map $\boldsymbol{\phi}(\x)$ induces the finite-dimensional kernel $k_D(\x,\x')=\boldsymbol{\phi}(\x)^\top\boldsymbol{\phi}(\x')$, so applying LS-SVM on $\boldsymbol{\phi}(\x)$ corresponds to applying LS-SVM in the induced kernel feature space. 
This motivates our use of the fixed nonlinear feature map before the LS-SVM classifier.

This property is particularly useful for CIL: the closed-form system depends on the data only through feature correlations $\tilde{\Phi}^{\top}\tilde{\Phi}$ and feature-label correlations $\tilde{\Phi}^{\top}Y$, which can be accumulated across sessions as additive sufficient statistics rather than optimized by repeatedly fine-tuning on the latest task. 
When new classes arrive, \name expands the one-vs-all target matrix to include the new classes and recomputes the classifier weights for all seen classes from the accumulated statistics. 
Therefore, the classifier corresponds to fitting an LS-SVM over all seen data represented by sufficient statistics, without storing or revisiting previous samples.

\subsection{Kernel Interpretation of the Fixed Nonlinear Feature Map}
\label{sec:suppl-kernel-map}

We provide a detailed explanation of why the fixed nonlinear feature map in Eq.~\ref{eq:method-random-feature} can be interpreted from a kernel perspective. 
Recall that \name maps the enhanced visual representation $\mathbf{u}(\x)$ to
\begin{equation}
    \boldsymbol{\phi}(\x)
    =
    \sigma(R^\top \mathbf{u}(\x))
    \in \mathbb{R}^{D},
\end{equation}
where $R$ is sampled once and then fixed, and $\sigma(\cdot)$ is a nonlinear activation.

\paragraph{Finite-dimensional kernel induced by the explicit map.}
For any fixed $R$, the explicit feature map $\boldsymbol{\phi}$ defines the following finite-dimensional kernel:
\begin{equation}
    k_D(\x,\x')
    =
    \boldsymbol{\phi}(\x)^\top
    \boldsymbol{\phi}(\x').
    \label{eq:suppl-finite-kernel}
\end{equation}
This is a valid positive semi-definite kernel. 
To see this, consider any finite set of samples $\{\x_i\}_{i=1}^{n}$ and form the feature matrix
\begin{equation}
    \Phi_D
    =
    \begin{bmatrix}
    \boldsymbol{\phi}(\x_1)^\top\\
    \boldsymbol{\phi}(\x_2)^\top\\
    \vdots\\
    \boldsymbol{\phi}(\x_n)^\top
    \end{bmatrix}
    \in\mathbb{R}^{n\times D}.
\end{equation}
The Gram matrix induced by $k_D$ is
\begin{equation}
    K_D
    =
    \big[k_D(\x_i,\x_j)\big]_{i,j=1}^{n}
    =
    \Phi_D\Phi_D^\top .
\end{equation}
Therefore, for any coefficient vector $\boldsymbol{\alpha}\in\mathbb{R}^{n}$,
\begin{align}
    \boldsymbol{\alpha}^{\top}K_D\boldsymbol{\alpha}
    &=
    \boldsymbol{\alpha}^{\top}
    \Phi_D\Phi_D^\top
    \boldsymbol{\alpha}
    \nonumber\\
    &=
    \left\|
    \Phi_D^\top\boldsymbol{\alpha}
    \right\|_2^2
    =
    \left\|
    \sum_{i=1}^{n}
    \alpha_i\boldsymbol{\phi}(\x_i)
    \right\|_2^2
    \geq 0 .
\end{align}
Thus, $K_D$ is positive semi-definite for any finite input set, and $k_D$ is a valid kernel. 
Consequently, training a linear LS-SVM on $\boldsymbol{\phi}(\x)$ is equivalent to training an LS-SVM in the finite-dimensional feature space induced by $k_D$. 
This argument is exact for any fixed $D$ and does not rely on asymptotic approximation.

\paragraph{Connection to random-feature kernel approximation.}
The randomness of $R$ further provides a connection to random-feature kernel approximation~\citep{rahimi2007random}. 
Let the columns of $R$ be independently sampled as $\mathbf{r}_1,\dots,\mathbf{r}_D$. 
For analysis, consider the normalized kernel
\begin{align}
    \bar{k}_D(\x,\x')
    &=
    \frac{1}{D}
    \boldsymbol{\phi}(\x)^\top
    \boldsymbol{\phi}(\x')
    \nonumber\\
    &=
    \frac{1}{D}
    \sum_{j=1}^{D}
    \sigma(\mathbf{r}_j^\top\mathbf{u}(\x))
    \sigma(\mathbf{r}_j^\top\mathbf{u}(\x')) .
    \label{eq:suppl-normalized-kernel}
\end{align}
For fixed $\x$ and $\x'$, by the law of large numbers,
\begin{equation}
    \bar{k}_D(\x,\x')
    \xrightarrow[D\rightarrow\infty]{}
    k_{\infty}(\x,\x')
    =
    \mathbb{E}_{\mathbf{r}}
    \left[
    \sigma(\mathbf{r}^{\top}\mathbf{u}(\x))
    \sigma(\mathbf{r}^{\top}\mathbf{u}(\x'))
    \right].
    \label{eq:suppl-limit-kernel}
\end{equation}
Therefore, the fixed finite-dimensional feature map used by \name can also be viewed as a Monte Carlo approximation to the nonlinear kernel $k_{\infty}$. 
The unnormalized kernel $k_D$ in Eq.~\ref{eq:suppl-finite-kernel} differs from $\bar{k}_D$ only by the constant factor $D$, which can be absorbed into the LS-SVM regularization coefficient.

\paragraph{ReLU activation and the arc-cosine kernel.}
In our implementation, $\sigma$ is the ReLU activation, and $R$ is randomly initialized once and then fixed. 
The finite-dimensional kernel interpretation above holds for any fixed realization of $R$. 
To connect this construction to a standard closed-form kernel, we consider the common Gaussian case, where each random direction satisfies $\mathbf{r}\sim\mathcal{N}(0,I)$. 
In this case, the limiting kernel in Eq.~\ref{eq:suppl-limit-kernel} corresponds to the first-order arc-cosine kernel~\citep{cho2009kernel}.

Let $\theta$ be the angle between $\mathbf{u}(\x)$ and $\mathbf{u}(\x')$, \ie,
\begin{equation}
    \cos\theta
    =
    \frac{\mathbf{u}(\x)^\top\mathbf{u}(\x')}
    {\|\mathbf{u}(\x)\|_2\|\mathbf{u}(\x')\|_2}.
\end{equation}
For ReLU activation and Gaussian random directions, we have
\begin{align}
    &\mathbb{E}_{\mathbf{r}\sim\mathcal{N}(0,I)}
    \left[
    \mathrm{ReLU}(\mathbf{r}^{\top}\mathbf{u}(\x))
    \mathrm{ReLU}(\mathbf{r}^{\top}\mathbf{u}(\x'))
    \right]
    \nonumber\\
    &\quad =
    \frac{\|\mathbf{u}(\x)\|_2\|\mathbf{u}(\x')\|_2}{2\pi}
    \left[
    \sin\theta
    +
    (\pi-\theta)\cos\theta
    \right].
    \label{eq:suppl-relu-arc-cosine}
\end{align}
This expression is the ReLU form of the first-order arc-cosine kernel. 
Thus, with Gaussian random directions and ReLU activation, the fixed nonlinear feature map provides a finite-dimensional approximation to a ReLU-induced nonlinear kernel space. 
For other random initializations, the exact limiting kernel may differ, but any fixed realization of $R$ still defines a valid finite-dimensional kernel through the inner product $\boldsymbol{\phi}(\x)^\top\boldsymbol{\phi}(\x')$. 
Since $R$ is fixed and never optimized, this kernelized feature space introduces no additional trainable parameters during incremental learning.

\noindent{\bf Why introduce a kernel-induced feature space.}
A linear classifier on $\mathbf{u}(\x)$ can only form hyperplane decision boundaries in the enhanced visual space. 
By contrast, applying a nonlinear feature map $\boldsymbol{\phi}(\x)=\sigma(R^\top\mathbf{u}(\x))$ and then learning a linear classifier on $\boldsymbol{\phi}(\x)$ yields decision boundaries that are nonlinear with respect to $\mathbf{u}(\x)$. 
Specifically, the class boundary between $c$ and $c'$ is
\begin{equation}
    (\mathbf{w}_c-\mathbf{w}_{c'})^\top \boldsymbol{\phi}(\x)
    +
    (b_c-b_{c'})
    =
    0,
\end{equation}
whose preimage in the original visual space is generally nonlinear because $\boldsymbol{\phi}$ is nonlinear. 
Therefore, the kernel-induced feature space increases classifier capacity while preserving the closed-form LS-SVM solution in the explicit feature space.

\noindent{\bf Why use the explicit feature map.}
Although one could use the implicit kernel $k_{\infty}$, doing so would require maintaining or recomputing sample-level kernel matrices over the data observed so far, which is inconvenient for CIL. 
Instead, \name uses the explicit finite-dimensional feature map $\boldsymbol{\phi}(\x)$. 
This keeps the LS-SVM in a primal closed-form formulation, where the required feature correlations and feature-label correlations can be accumulated as additive sufficient statistics. 
Therefore, the fixed nonlinear feature map provides nonlinear classifier capacity while preserving the efficient incremental update rule of \name without storing previous samples.

\begin{figure*}[ht]
    \centering
    \begin{subfigure}[t]{0.32\textwidth}
        \centering
        \includegraphics[width=1\linewidth]{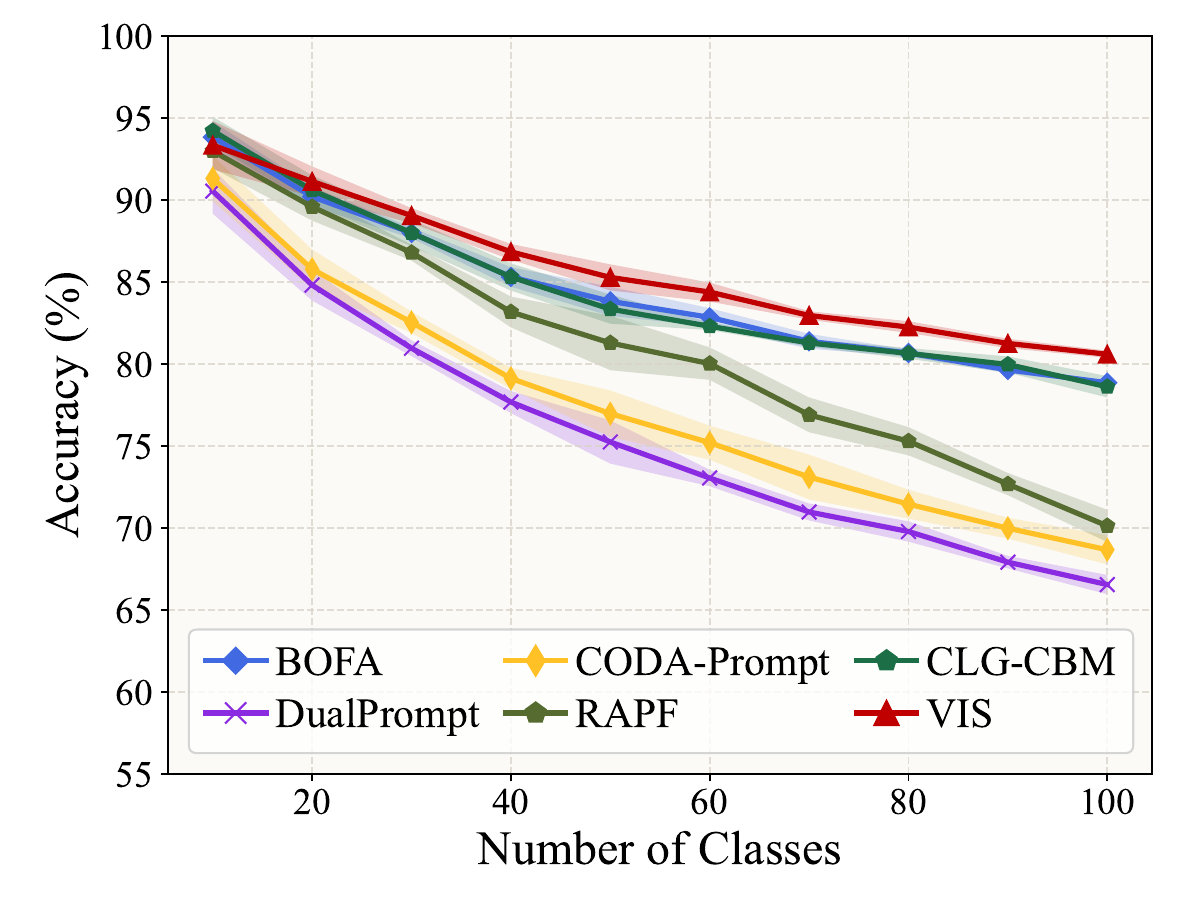}
        \caption{Results on ImageNet-R B0 Inc20 averaged over five class-order seeds. \name consistently outperforms the compared methods across incremental stages.}
        \label{fig:app_seed}
    \end{subfigure}%
    \hfill
    \begin{subfigure}[t]{0.32\textwidth}
        \centering
        \includegraphics[width=1\linewidth]{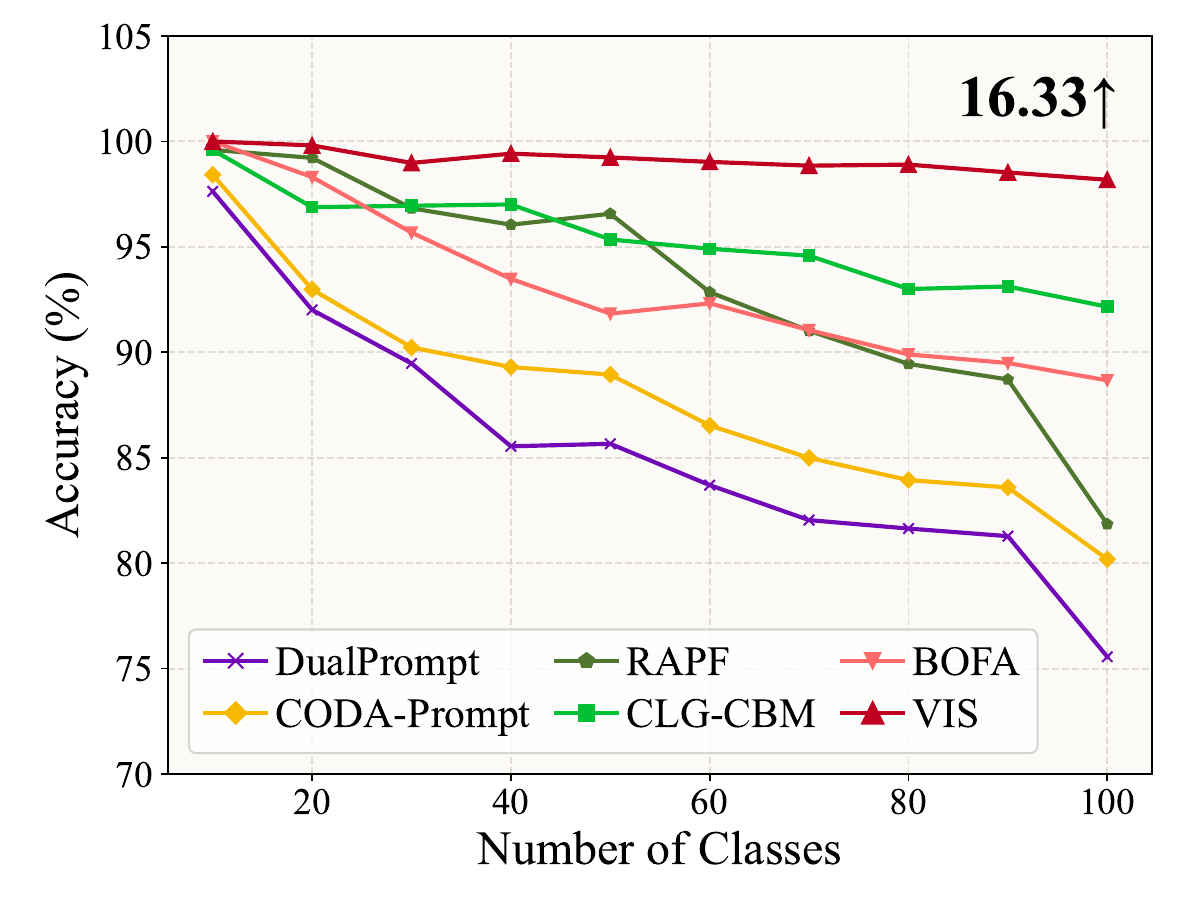}
        \caption{Experiments when using OpenAI weights on UCF B0 Inc10. \name consistently outperforms other methods across different backbone weights.}
        \label{fig:app_backbone}
    \end{subfigure}%
    \hfill
    \begin{subfigure}[t]{0.32\textwidth}
        \centering
        \includegraphics[width=1\linewidth]{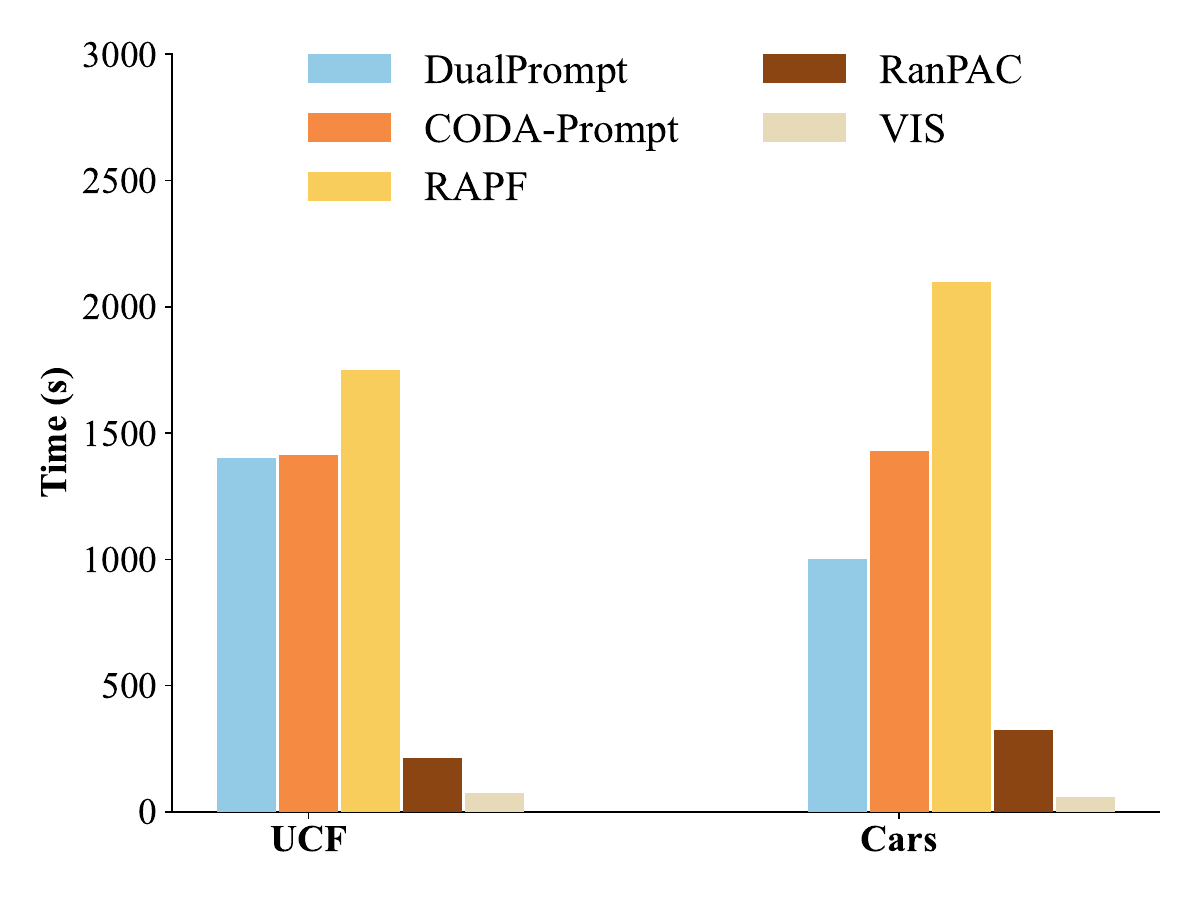}
        \caption{Training time comparison on UCF B0 Inc10 and Cars B0 Inc10.}
        \label{fig:training_time}
    \end{subfigure}
\caption{
Additional analysis of \name.
Left: robustness to different class-order seeds.
Middle: performance under different CLIP pre-trained weights.
Right: training time comparison with representative baselines.
}
    \label{fig:overall_analysis}
\end{figure*}

\section{Supplementary Results and Analyses}
\label{sec:app_more_results}

\subsection{Evaluation Across Multiple Random Seeds}
The main experiments follow the standard CIL protocol~\citep{rebuffi2017icarl} and use the class-order seed 1993. To assess robustness to class ordering, we further run five independent splits with seeds \{1993, 1994, 1995, 1996, 1997\} and report the mean accuracy with standard deviation. As shown in Figure~\ref{fig:app_seed}, \name maintains the best performance on ImageNet-R B0 Inc20 across repeated runs, indicating that its advantage is stable under different class orders.

\subsection{Random-Feature Robustness}
\label{sec:suppl-random-feature-seed}

The projection matrix $R$ in Eq.~\ref{eq:method-random-feature} is randomly initialized once and then fixed throughout all incremental sessions. 
To examine whether \name depends on a particular realization of the random feature map, we vary only the initialization seed of $R$, while keeping the class order and all other experimental settings unchanged. 
For reference, we also report representative analytic and CLIP-based baselines under the same incremental protocols. 
The results of \name are reported as mean $\pm$ standard deviation across different random-feature seeds, whereas the baseline results correspond to their standard evaluations.

\begin{table*}[t]
    \centering
    \caption{
    Random-feature robustness under different initialization seeds of $R$.
    Results of \name are reported as mean $\pm$ standard deviation across random-feature seeds, while the other methods are included as references under the same incremental protocols.
    }
    \label{tab:random_feature_seed}
    \footnotesize
    \setlength{\tabcolsep}{10pt}
    \renewcommand{\arraystretch}{0.95}
    \resizebox{\textwidth}{!}{
    \begin{NiceTabular}{@{}lcccc@{}}
        \toprule
        \multirow{2}{*}{Method}
        & \multicolumn{2}{c}{Aircraft B0 Inc10}
        & \multicolumn{2}{c}{CIFAR100 B0 Inc10} \\
        \cmidrule(lr){2-3}
        \cmidrule(lr){4-5}
        & $\bar{\mathcal{A}}$
        & $\mathcal{A}_B$
        & $\bar{\mathcal{A}}$
        & $\mathcal{A}_B$ \\
        \midrule
        ACIL~\citep{zhuang2022acil}
        & 64.99 & 56.68
        & 89.41 & 83.73 \\

        RanPAC~\citep{mcdonnell2023ranpac}
        & 69.77 & 60.28
        & 89.30 & 83.18 \\

        BOFA~\citep{li2026bofa}
        & 70.96 & 60.43
        & 86.07 & 79.19 \\

        ENGINE~\citep{zhou2025external}
        & 69.74 & 58.51
        & 86.89 & 79.29 \\

        \midrule
        \rowcolor{LightCyan}
        \name (Ours)
        & $\mathbf{75.236 \pm 0.106}$
        & $\mathbf{66.520 \pm 0.277}$
        & $\mathbf{90.598 \pm 0.073}$
        & $\mathbf{85.428 \pm 0.167}$ \\
        \bottomrule
    \end{NiceTabular}
    }
\end{table*}
As shown in Table~\ref{tab:random_feature_seed}, \name exhibits consistently small variations across different random feature mappings on both datasets. 
The standard deviations are at most $0.106$ for average accuracy and $0.277$ for final-session accuracy on Aircraft, and further decrease to $0.073$ and $0.167$ on CIFAR100. 
Moreover, these variations are substantially smaller than the performance margins over the compared baselines. 
These results indicate that the performance of \name is stable across different samplings of $R$ and that its gains do not depend on a favorable random feature initialization.
\subsection{Parameter Robustness}
\label{sec:suppl-param-robustness}

For the parameter sensitivity analysis reported in the main paper, we use a validation split from the original training data. 
Specifically, the training data are split into training and validation subsets with a ratio of $4{:}1$, and all sensitivity results are reported as final-stage accuracies on the validation subset. 
The results show stable performance trends across different hyperparameter settings, further supporting the robustness of \name.

\subsection{Results with Different Backbones}
Our main experiments use CLIP ViT-B/16 with LAION-400M pre-trained weights~\citep{ilharco2021openclip}. To evaluate whether the proposed visual-only analytic framework generalizes across CLIP initializations, we further test \name with OpenAI CLIP weights~\citep{radford2021learning} on UCF B0 Inc10, as shown in Figure~\ref{fig:app_backbone}. \name consistently outperforms the compared methods under this alternative backbone, indicating that its advantage is not tied to a specific CLIP pre-training source.

\begin{table*}[t]
    \centering
    \caption{
    Generalization across different pre-trained backbones under the B0 Inc10 protocol.
    All methods use the same backbone within each group.
    We report average accuracy $\bar{\mathcal{A}}$ and final-session accuracy $\mathcal{A}_B$.
    }
    \label{tab:backbone_generalization}
    \footnotesize
    \setlength{\tabcolsep}{15pt}
    \renewcommand{\arraystretch}{1.08}
    \resizebox{\textwidth}{!}{
    \begin{NiceTabular}{@{}llcccc}
        \toprule
        \multirow{2}{*}{Backbone}
        & \multirow{2}{*}{Method}
        & \multicolumn{2}{c}{Aircraft}
        & \multicolumn{2}{c}{CIFAR100} \\
        \cmidrule(lr){3-4}
        \cmidrule(lr){5-6}
        &
        & $\bar{\mathcal{A}}$ & $\mathcal{A}_B$
        & $\bar{\mathcal{A}}$ & $\mathcal{A}_B$ \\
        \midrule

        \multirow{3}{*}{DINOv2 ViT-B/14}
        & SimpleCIL~\citep{zhou2025revisiting}
        & 46.00 & 37.23
        & 92.36 & 88.10 \\
        & RanPAC~\citep{mcdonnell2023ranpac}
        & 87.39 & 79.81
        & 90.09 & 83.58 \\
        \rowcolor{LightCyan}
        & \name (Ours)
        & \textbf{87.68} & \textbf{80.53}
        & \textbf{95.17} & \textbf{91.98} \\

        \midrule

        \multirow{3}{*}{SigLIP ViT-B/16}
        & SimpleCIL~\citep{zhou2025revisiting}
        & 78.18 & 69.76
        & 81.16 & 74.34 \\
        & RanPAC~\citep{mcdonnell2023ranpac}
        & 76.35 & 66.31
        & 86.87 & 79.85 \\
        \rowcolor{LightCyan}
        & \name (Ours)
        & \textbf{82.68} & \textbf{75.40}
        & \textbf{89.36} & \textbf{83.83} \\

        \bottomrule
    \end{NiceTabular}
    }
\end{table*}

\subsection{Backbone Generalization}
\label{sec:suppl-backbone-generalization}

To examine whether \name is tied to the CLIP backbone used in our main experiments, we further evaluate it with two representative pre-trained backbones: the visual-only self-supervised DINOv2 ViT-B/14~\citep{oquab2023dinov2} and the more recent vision-language model SigLIP ViT-B/16~\citep{zhai2023sigmoid}. We compare \name with SimpleCIL and RanPAC using the same backbone and B0 Inc10 protocol on Aircraft and CIFAR100.

As shown in Table~\ref{tab:backbone_generalization}, \name consistently achieves the best performance under both DINOv2 and SigLIP, despite noticeable changes in the relative strength of SimpleCIL and RanPAC across different pre-trained representations. Compared with RanPAC, \name improves average/final-session accuracy by $0.30/0.72$ and $5.08/8.40$ points with DINOv2 on Aircraft and CIFAR100, respectively, and by $6.33/9.09$ and $2.49/3.98$ points with the more recent vision-language model SigLIP. These matched-backbone comparisons show that the effectiveness of \name generalizes across different pre-training paradigms. In particular, its consistent gains with SigLIP indicate that the visual-space formulation remains effective with a modern vision-language-pretrained representation, further supporting that strong continual classification does not require continued dependence on textual classifier weights.

\subsection{Forgetting Analysis}
\label{sec:suppl-forgetting}

In addition to average and final-stage accuracy, we further evaluate the forgetting behavior of different methods. 
Following standard CIL evaluation protocols~\citep{rebuffi2017icarl,zhou2025learning}, let $\mathcal{A}_{l,b}$ denote the accuracy on task $b$ after learning stage $l$, and let $B$ be the total number of stages. 
The forgetting measure after the final stage is defined as:
\begin{equation}
      F_{B}
      =
      \frac{1}{B-1}
      \sum_{b=1}^{B-1}
      \left(
      \max_{l \in \{b,\dots,B-1\}}
      \mathcal{A}_{l,b}
      -
      \mathcal{A}_{B,b}
      \right),
      \label{eq:suppl-forgetting}
\end{equation}
where a lower value indicates less forgetting on previously learned tasks. 
\begin{table*}[ht]
    \centering
    \caption{
    Forgetting measure $F_B$ of different methods under different incremental settings. 
    Lower $F_B$ indicates less forgetting, and the best result in each column is highlighted in bold.
    }
    \label{tab:forgetting}
    \footnotesize
    \setlength{\tabcolsep}{8pt}
    \renewcommand{\arraystretch}{0.95}
    \resizebox{\textwidth}{!}{
    \begin{NiceTabular}{@{}lcccccccc}
        \toprule
        \multirow{2}{*}{Method}
        & \multicolumn{2}{c}{Food}
        & \multicolumn{2}{c}{Cars}
        & \multicolumn{2}{c}{UCF}
        & \multicolumn{2}{c}{SUN} \\
        \cmidrule(lr){2-3}
        \cmidrule(lr){4-5}
        \cmidrule(lr){6-7}
        \cmidrule(lr){8-9}
        & B0 Inc10 & B50 Inc10
        & B0 Inc10 & B50 Inc10
        & B0 Inc10 & B50 Inc10
        & B0 Inc30 & B150 Inc30 \\
        \midrule
        SimpleCIL~\citep{zhou2025revisiting}
        & 6.22 & 3.76 & 5.11 & 2.51 & 4.86 & 2.31 & 8.12 & 3.51 \\
        ACIL~\citep{zhuang2022acil}
        & 4.97 & 2.94 & 5.48 & 2.27 & 1.17 & 0.74 & 8.41 & 4.52 \\
        DualPrompt~\citep{wang2022dualprompt}
        & 24.28 & 20.77 & 7.35 & 4.71 & 9.26 & 7.95 & 23.54 & 17.63 \\
        CODA-Prompt~\citep{smith2023coda}
        & 25.82 & 22.41 & 6.14 & 4.59 & 9.80 & 9.21 & 22.71 & 17.69 \\
        RanPAC~\citep{mcdonnell2023ranpac}
        & 5.39 & \textbf{2.56} & 3.96 & \textbf{1.71} & 1.32 & 0.98 & 7.09 & \textbf{3.34} \\
        RAPF~\citep{huang2024class}
        & 8.16 & 7.83 & 28.27 & 29.30 & 15.89 & 20.50 & 14.58 & 9.36 \\
        CLG-CBM~\citep{yu2025language}
        & 11.79 & 9.38 & 6.86 & 4.86 & 8.25 & 6.87 & 14.62 & 12.26 \\
        PROOF~\citep{zhou2025learning}
        & 16.58 & 16.94 & 4.48 & 2.30 & 4.22 & 4.85 & 18.33 & 15.71 \\
        BOFA~\citep{li2026bofa}
        & 5.87 & 4.08 & 3.94 & 2.17 & 4.84 & 4.13 & 8.47 & 4.16 \\
        \midrule
        \rowcolor{LightCyan}
        \name (Ours)
        & \textbf{4.23} & 2.68
        & \textbf{3.57} & 1.85
        & \textbf{0.72} & \textbf{0.38}
        & \textbf{6.97} & {3.79} \\
        \bottomrule
    \end{NiceTabular}
    }
\end{table*}

Table~\ref{tab:forgetting} shows that \name consistently achieves low forgetting across datasets and split protocols. 
It obtains the lowest forgetting in five out of eight settings, including Food B0 Inc10, Cars B0 Inc10, both UCF settings, and SUN B0 Inc30. 
On the remaining settings, \name remains close to the best method, with small gaps to RanPAC on Food B50 Inc10, Cars B50 Inc10, and SUN B150 Inc30. 
These results suggest that recomputing the classifier over all seen classes from additive sufficient statistics effectively preserves old-class information. 
Compared with optimization-based prompt or adaptation methods, which often suffer larger forgetting under several settings, closed-form statistic-based classifiers generally show more stable old-class retention. 
Among them, \name further combines low forgetting with the strong average and final-stage accuracy reported in the main experiments, indicating that its performance gain is not achieved by sacrificing old-class knowledge.

\subsection{Controlled Evaluation of Textual Information}
\label{sec:suppl-text-ablation}

The comparison between ZS-CLIP~\citep{radford2021learning} and \name changes both the use of textual information and the classifier design. 
To isolate the effect of textual information itself, we therefore construct a controlled variant, denoted as \textbf{``\name + Text''}. 
It keeps the visual representation, nonlinear feature map, LS-SVM classifier, hyperparameters, and incremental update procedure identical to \name, while additionally incorporating CLIP textual class prototypes into the same classifier feature space and sufficient statistics. 
Thus, \name + Text and \name differ only in whether textual class information is introduced. 
For reference, we also report ZS-CLIP results from ENGINE~\citep{zhou2025external}, which uses the same CLIP backbone and incremental protocols.

\begin{table}[t]
    \centering
    \caption{
    Controlled evaluation of textual information.
    ZS-CLIP serves as a textual-classifier reference, while \name + Text and \name form a controlled pair differing only in the use of textual class prototypes.
    }
    \label{tab:text_ablation}
    \vspace{-2mm}
    \small
    \setlength{\tabcolsep}{7pt}
    \begin{tabular}{lcccc}
        \toprule
        \multirow{2}{*}{Method}
        & \multicolumn{2}{c}{Aircraft B0 Inc10}
        & \multicolumn{2}{c}{CIFAR100 B0 Inc10} \\
        \cmidrule(lr){2-3}
        \cmidrule(lr){4-5}
        & $\bar{\mathcal{A}}$ & $\mathcal{A}_B$
        & $\bar{\mathcal{A}}$ & $\mathcal{A}_B$ \\
        \midrule
        ZS-CLIP~\citep{radford2021learning}
        & 26.66 & 17.22
        & 81.81 & 71.38 \\

        \name + Text
        & 73.18 & 64.69
        & 90.56 & \textbf{85.43} \\

        \rowcolor{LightCyan}
        \name
        & \textbf{75.21} & \textbf{66.46}
        & \textbf{90.64} & 85.26 \\
        \bottomrule
    \end{tabular}
    \vspace{-2mm}
\end{table}

As shown in Table~\ref{tab:text_ablation}, ZS-CLIP provides a reference for directly using textual embeddings as classifier weights, while the comparison between \name + Text and \name more directly isolates the effect of textual information within our framework. 
Adding textual prototypes provides no consistent benefit: on Aircraft, it decreases average and final-session accuracy by $2.03$ and $1.77$ points, respectively, while on CIFAR100 the differences are marginal and mixed. 
These results show that the strong performance of \name primarily comes from its visual-space formulation and does not rely on additional textual class information.
\subsection{Training Time}
\label{sec:training_time}

We compare the training time of \name with representative baselines on UCF and Cars in Figure~\ref{fig:training_time}. 
All training-time results are measured on a single NVIDIA RTX 4090 GPU. 
\name is consistently the most efficient method, since only the lightweight visual modules are trained in the base task and later tasks require only sufficient-statistic updates and a closed-form LS-SVM solve. 
Compared with prompt-based and CLIP-adaptation methods, \name avoids repeated task-wise optimization, and it is also faster than RanPAC under the same epoch budget.

\begin{table*}[ht]
    \caption{\small Overhead comparison on \textbf{Cars B0 Inc10}. 
    All methods start from the same pre-trained CLIP backbone and are evaluated under the same hardware environment, using a \textbf{single RTX 3090 GPU} per method.
    ``Total Time'' denotes the end-to-end wall-clock time from the start of continual training to the completion of the final evaluation.
    ``Peak GPU'' denotes the peak training GPU memory of the current process, and ``Infer.'' reports the average evaluation latency per test image aggregated over all sessions. 
    Lower is better for all overhead metrics, while higher is better for $\bar{\mathcal{A}}$.}
    \label{tab:overhead_cars_b0_inc10}
    \vspace{-2mm}
    \centering
    \resizebox{\textwidth}{!}{%
    \begin{NiceTabular}{@{}lcccccc@{}}
        \toprule
        Method 
        & $\bar{\mathcal{A}}$ 
        & Total Time (min)$\downarrow$ 
        & Train Time (min)$\downarrow$ 
        & Eval Time (min)$\downarrow$ 
        & Peak GPU (GB)$\downarrow$ 
        & Infer. (ms/img)$\downarrow$ \\
        \midrule
        DualPrompt~\citep{wang2022dualprompt} & 63.43 & 48.90 & 45.68 & 3.22 & 6.95 & 8.58 \\
        CODA-Prompt~\citep{smith2023coda} & 66.81 & 53.23 & 49.64 & 3.59 & 15.53 & 9.58 \\
        SimpleCIL~\citep{zhou2025revisiting} & 92.11 & \textbf{1.50} & \textbf{0.37} & \textbf{1.13} & \textbf{0.92} & 3.02 \\
        RAPF~\citep{huang2024class} & 82.12 & 19.96 & 18.62 & 1.34 & 6.43 & 3.59 \\
        CLG-CBM~\citep{yu2025language} & 93.15 & 32.90 & 31.77 & \textbf{1.13} & 1.05 & \textbf{3.01} \\
        PROOF~\citep{zhou2025learning} & 90.44 & 33.26 & 30.85 & 2.41 & 1.16 & 6.43 \\
        BOFA~\citep{li2026bofa} & 94.26 & 11.83 & 9.70 & 2.13 & 1.73 & 5.68 \\
        ACIL~\citep{zhuang2022acil} & 93.08 & 1.60 & 0.41 & 1.20 & 1.11 & 3.19 \\
        RanPAC~\citep{mcdonnell2023ranpac} & 93.92 & 6.47 & 5.16 & 1.31 & 5.30 & 3.49 \\
        \midrule
        \rowcolor{LightCyan}
        \name\ (Ours) & \textbf{94.84} & 1.72 & 0.55 & 1.17 & 4.53 & 3.12 \\
        \bottomrule
    \end{NiceTabular}}
    \vspace{-4mm}
\end{table*}
\subsection{Overhead Analysis}
\label{sec:overhead_analysis}

Table~\ref{tab:overhead_cars_b0_inc10} reports the computational overhead on Cars B0 Inc10. 
All methods use the same pre-trained CLIP backbone, and each method is evaluated on a single NVIDIA RTX 3090 GPU. 
Among the compared methods, \name achieves the best average accuracy ($94.84\%$) with competitive overall efficiency.

\noindent\textbf{Training cost.}
\name completes the whole continual learning process in $1.72$ minutes, which is close to the most efficient baselines SimpleCIL ($1.50$ min) and ACIL ($1.60$ min). 
Its training time is also low ($0.55$ min), only slightly higher than SimpleCIL ($0.37$ min) and ACIL ($0.41$ min). 
Compared with methods that require heavier task-wise optimization or adaptation, \name is substantially faster than RanPAC ($6.47$ min), BOFA ($11.83$ min), RAPF ($19.96$ min), CLG-CBM ($32.90$ min), PROOF ($33.26$ min), DualPrompt ($48.90$ min), and CODA-Prompt ($53.23$ min) in total time. 
This indicates that the closed-form update in \name keeps the training overhead low.

\noindent\textbf{Inference cost.}
\name requires $3.12$ ms per image during evaluation, which is close to efficient baselines such as CLG-CBM ($3.01$ ms), SimpleCIL ($3.02$ ms), ACIL ($3.19$ ms), and RanPAC ($3.49$ ms). 
It is also clearly lower than BOFA ($5.68$ ms), PROOF ($6.43$ ms), DualPrompt ($8.58$ ms), and CODA-Prompt ($9.58$ ms). 
Thus, the improved accuracy of \name does not introduce a large inference-time burden.

\noindent\textbf{Memory usage.}
For peak training GPU memory, \name uses $4.53$ GB. 
This is higher than lightweight methods such as SimpleCIL ($0.92$ GB), CLG-CBM ($1.05$ GB), ACIL ($1.11$ GB), PROOF ($1.16$ GB), and BOFA ($1.73$ GB), but lower than RanPAC ($5.30$ GB), RAPF ($6.43$ GB), DualPrompt ($6.95$ GB), and CODA-Prompt ($15.53$ GB). 
Therefore, \name maintains a moderate memory footprint among the compared methods.

Overall, these results indicate that \name improves accuracy without sacrificing practical efficiency, achieving the best performance among the compared methods while maintaining competitive training time, inference latency, and memory usage.

\begin{figure*}[t]
	\centering
	\begin{subfigure}{0.32\linewidth}
		\includegraphics[width=1\columnwidth]{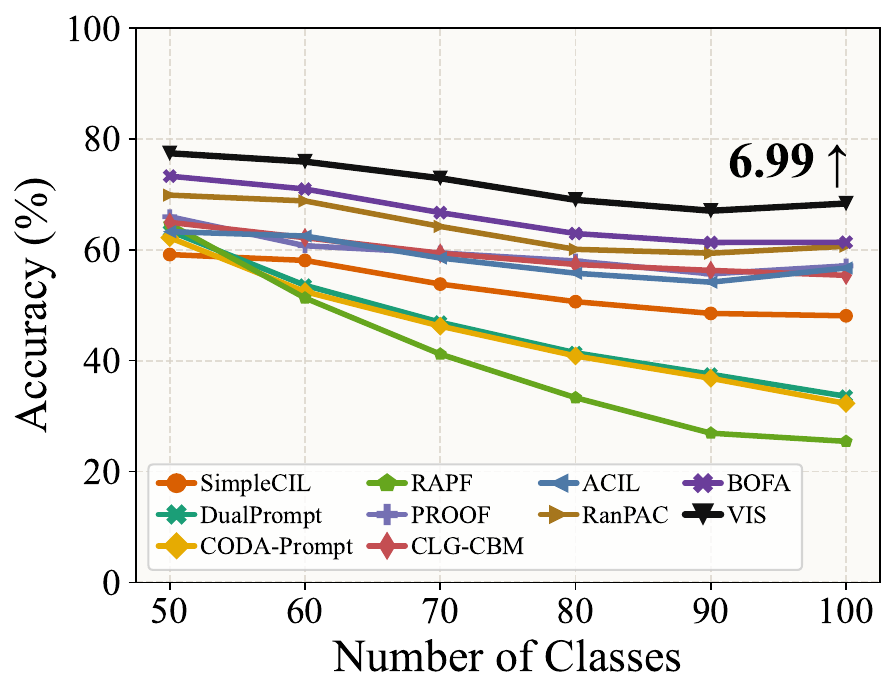}
		\caption{Aircraft Base50 Inc10}
		\label{fig:benchmark-aircraft50}
	\end{subfigure}
	\hfill
	\begin{subfigure}{0.32\linewidth}
		\includegraphics[width=1\linewidth]{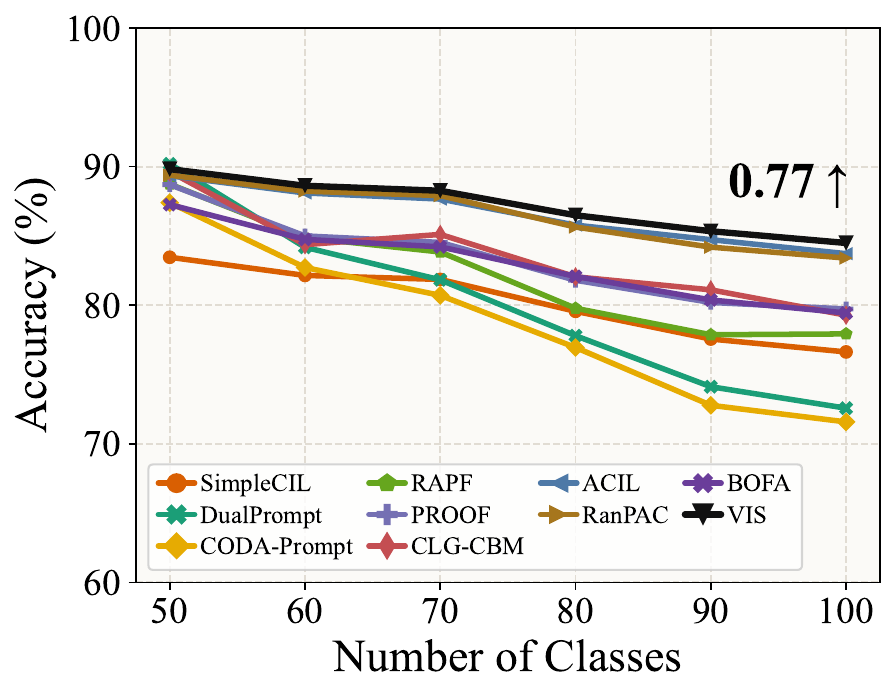}
		\caption{CIFAR100 Base50 Inc10}
		\label{fig:benchmark-cifar50}
	\end{subfigure}
	\hfill
	\begin{subfigure}{0.32\linewidth}
		\includegraphics[width=1\linewidth]{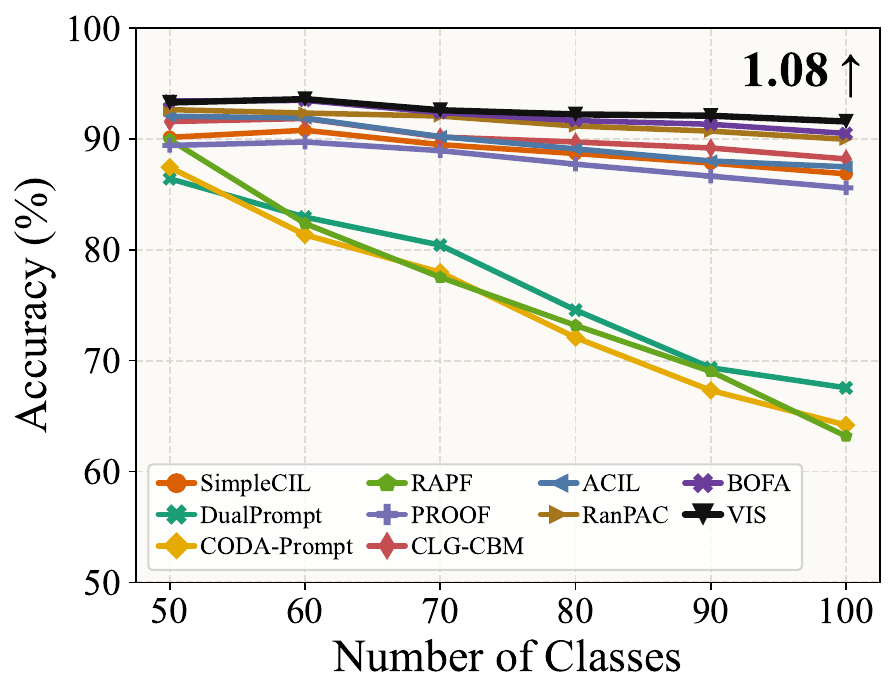}
		\caption{Cars Base50 Inc10}
		\label{fig:benchmark-cars50}
	\end{subfigure}
	\\
	\begin{subfigure}{0.32\linewidth}
		\includegraphics[width=1\linewidth]{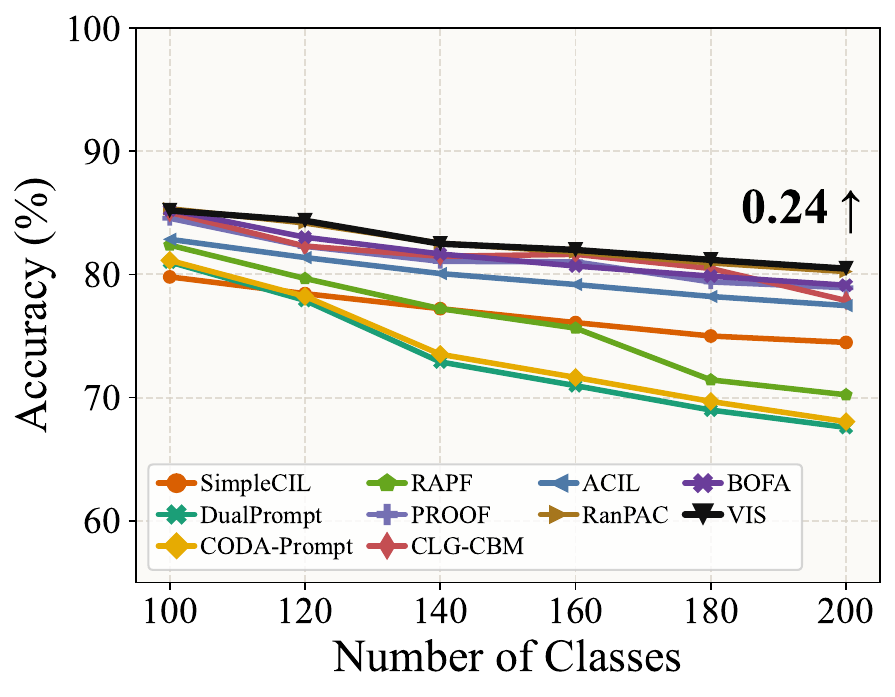}
		\caption{ImageNet-R Base100 Inc20}
		\label{fig:benchmark-imagenetr100}
	\end{subfigure}
	\hfill
	\begin{subfigure}{0.32\linewidth}
		\includegraphics[width=1\linewidth]{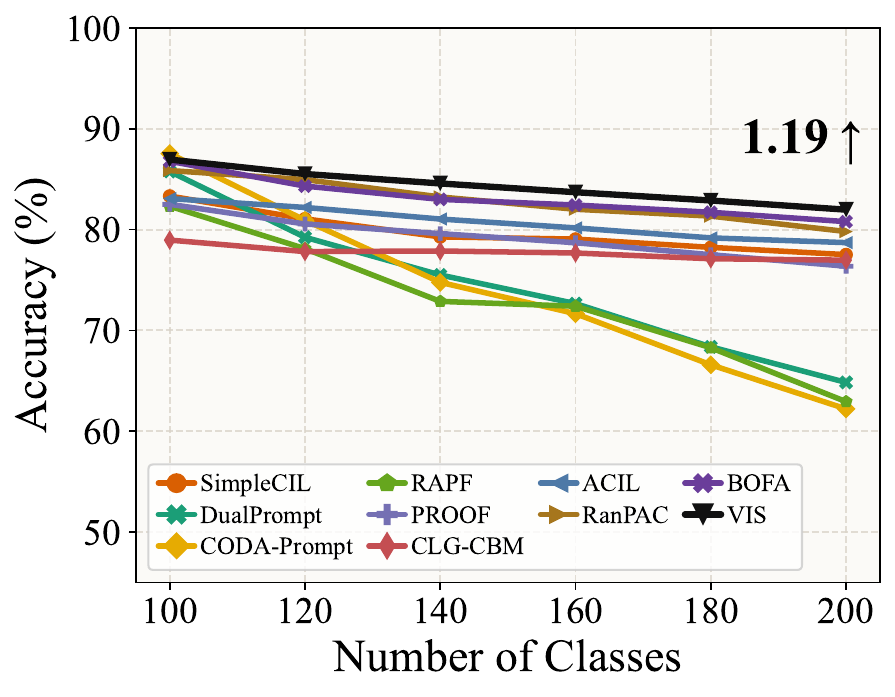}
		\caption{CUB Base100 Inc20}
		\label{fig:benchmark-cub100}
	\end{subfigure}
	\hfill
	\begin{subfigure}{0.32\linewidth}
		\includegraphics[width=1\columnwidth]{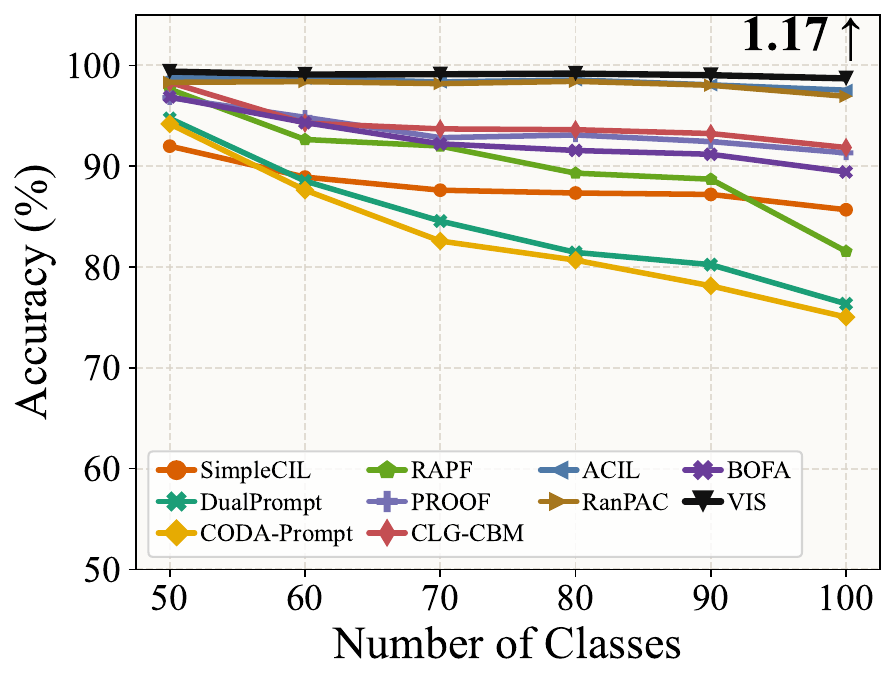}
		\caption{UCF Base50 Inc10}
		\label{fig:benchmark-ucf50}
	\end{subfigure}
	\\
	\begin{subfigure}{0.32\linewidth}
		\includegraphics[width=1\linewidth]{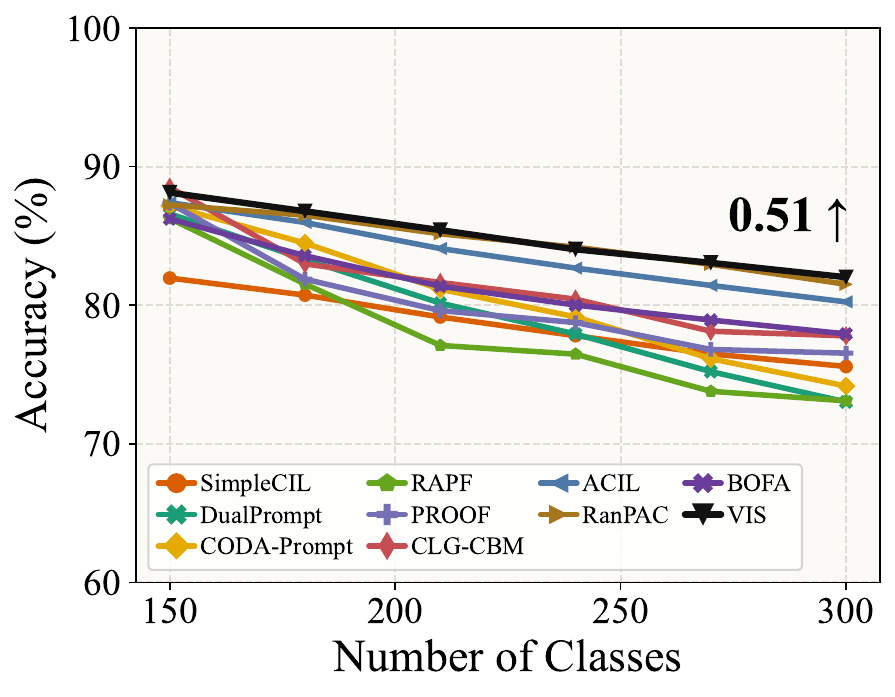}
		\caption{SUN Base150 Inc30}
		\label{fig:benchmark-sun150}
	\end{subfigure}
	\hfill
	\begin{subfigure}{0.32\linewidth}
		\includegraphics[width=1\linewidth]{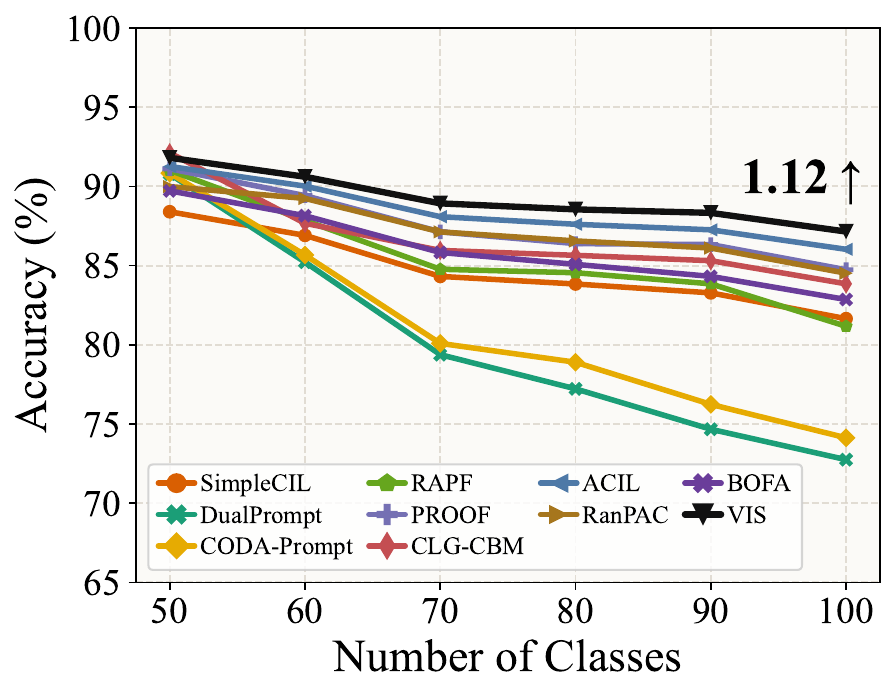}
		\caption{Food Base50 Inc10}
		\label{fig:benchmark-food50}
	\end{subfigure}
	\hfill
	\begin{subfigure}{0.32\linewidth}
		\includegraphics[width=1\columnwidth]{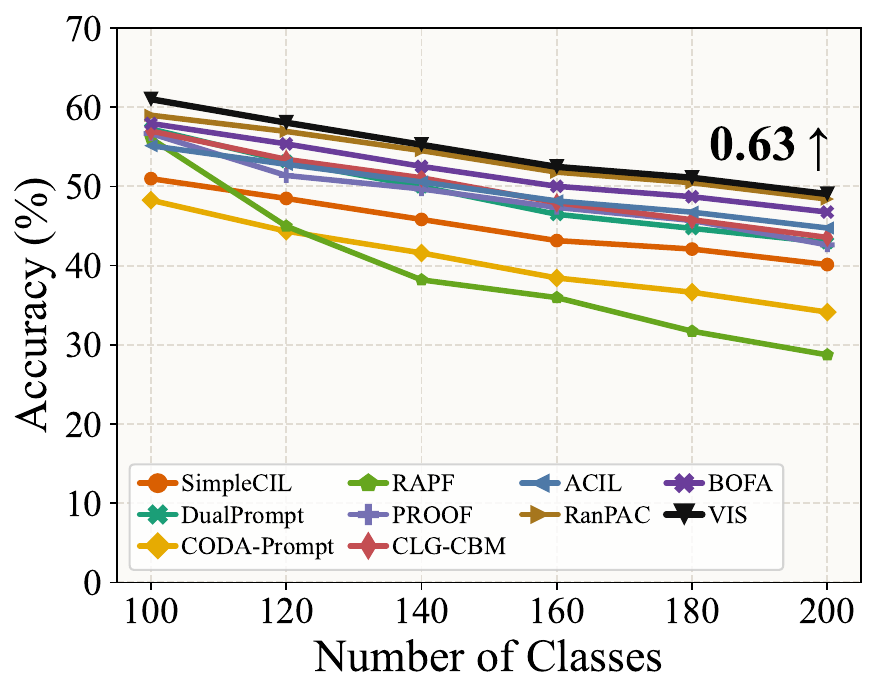}
		\caption{ObjectNet Base100 Inc20}
		\label{fig:benchmark-objectnet100}
	\end{subfigure}
	\caption{ 	Incremental performance of different methods on half-base setting. We report the performance gap after the last incremental stage of \name and the runner-up method at the end of the line.    All methods utilize the same CLIP pre-trained weight. }
	\label{fig:supp-benchmark-b50}
\end{figure*}

\begin{figure*}[t]
	\begin{subfigure}{0.32\linewidth}
		\includegraphics[width=1\columnwidth]{figure/aircraft}
		\caption{Aircraft Base0 Inc10}
	\end{subfigure}
	\hfill
	\begin{subfigure}{0.32\linewidth}
		\includegraphics[width=1\linewidth]{figure/cifar}
		\caption{CIFAR100 Base0 Inc10}
	\end{subfigure}
	\hfill
	\begin{subfigure}{0.32\linewidth}
		\includegraphics[width=1\linewidth]{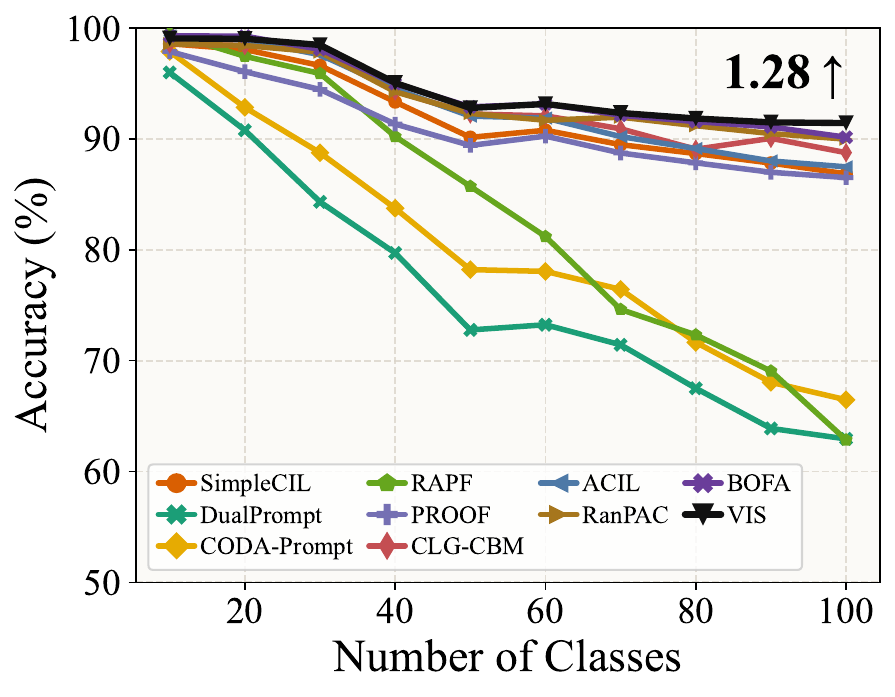}
		\caption{Cars Base0 Inc10}
	\end{subfigure}
	\\
	\begin{subfigure}{0.32\linewidth}
		\includegraphics[width=1\linewidth]{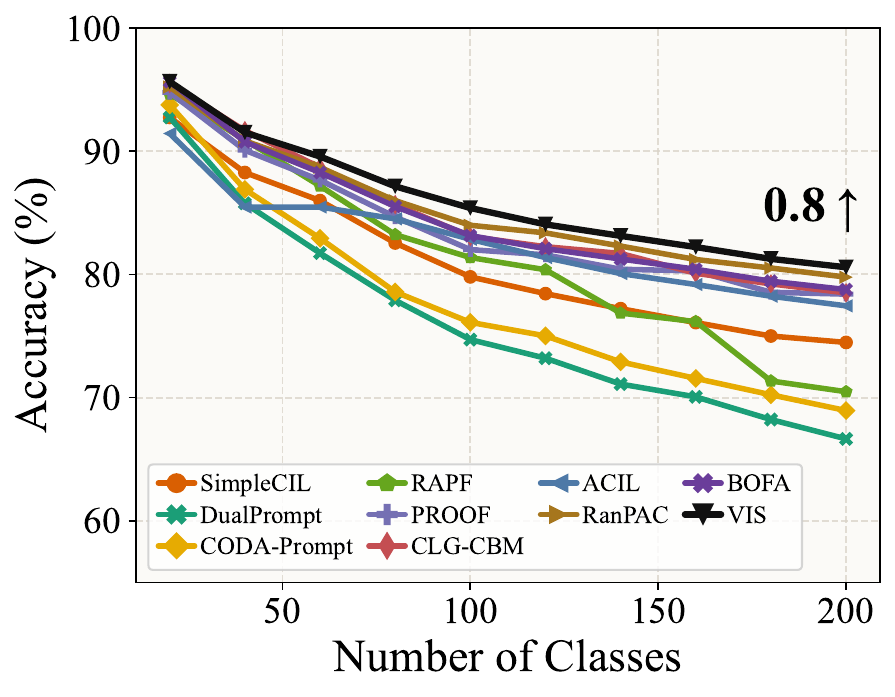}
		\caption{ImageNet-R Base0 Inc20}
	\end{subfigure}
	\hfill
	\begin{subfigure}{0.32\linewidth}
		\includegraphics[width=1\linewidth]{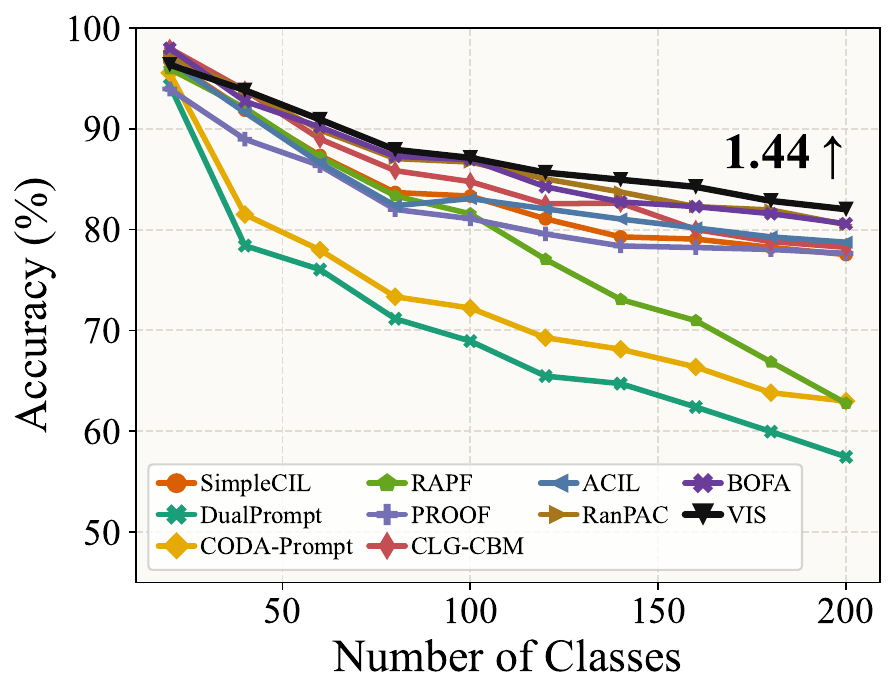}
		\caption{CUB Base0 Inc20}
	\end{subfigure}
	\hfill
	\begin{subfigure}{0.32\linewidth}
		\includegraphics[width=1\columnwidth]{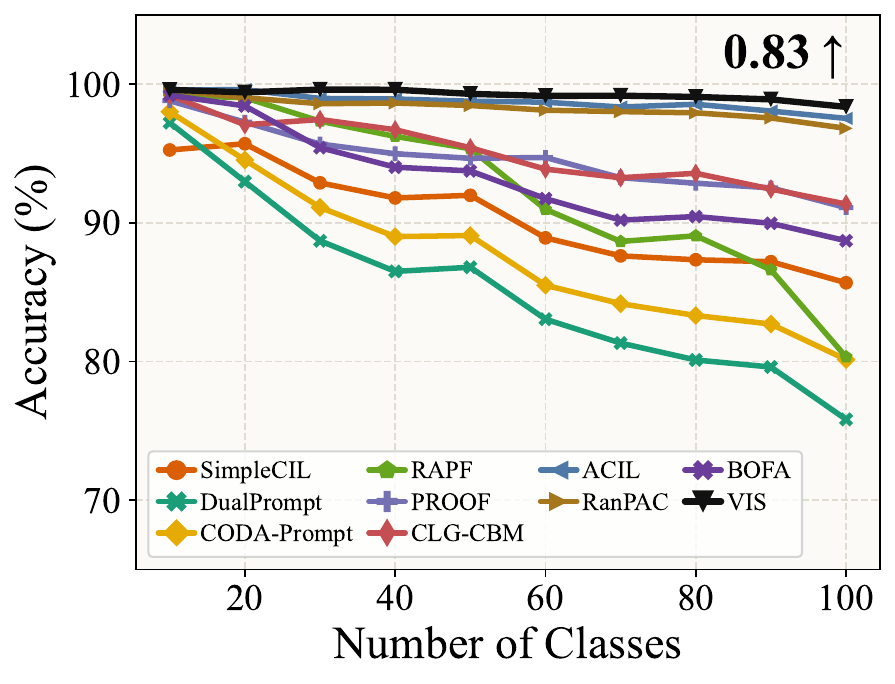}
		\caption{UCF Base0 Inc10}
	\end{subfigure}
	\\
	\begin{subfigure}{0.32\linewidth}
		\includegraphics[width=1\linewidth]{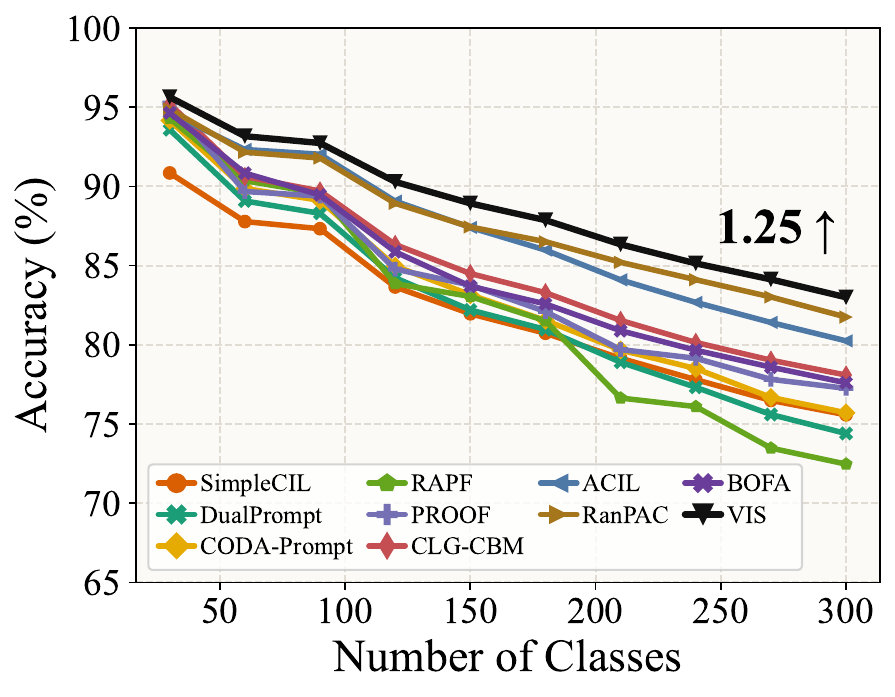}
		\caption{SUN Base0 Inc30}
	\end{subfigure}
	\hfill
	\begin{subfigure}{0.32\linewidth}
		\includegraphics[width=1\linewidth]{figure/food}
		\caption{Food Base0 Inc10}
	\end{subfigure}
	\hfill
	\begin{subfigure}{0.32\linewidth}
		\includegraphics[width=1\columnwidth]{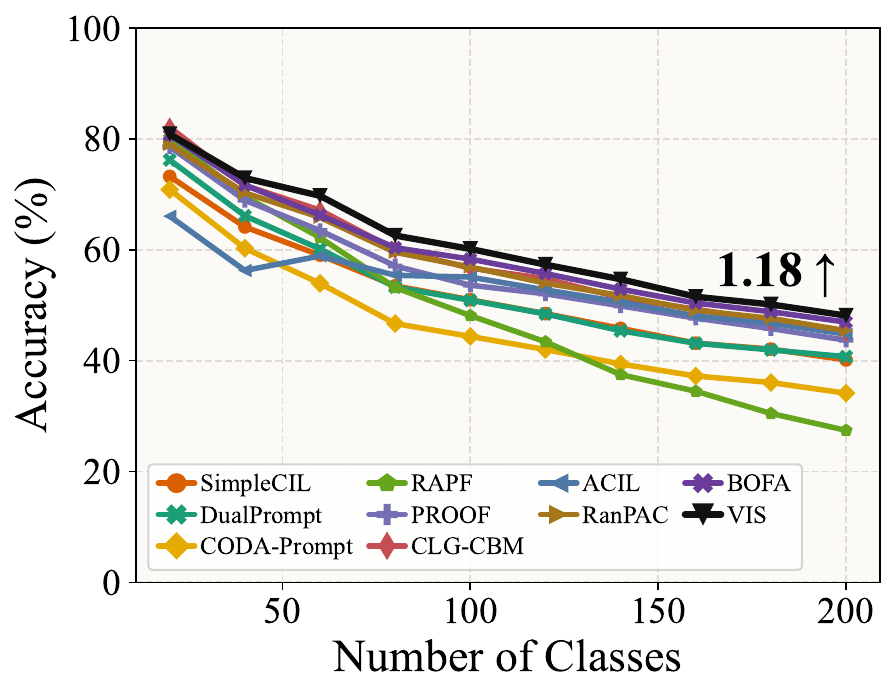}
		\caption{ObjectNet Base0 Inc20}
	\end{subfigure}
	\caption{ 	Incremental performance of different methods on B0 setting. We report the performance gap after the last incremental stage of \name and the runner-up method at the end of the line.    All methods utilize the same CLIP pre-trained weight. }
	\label{fig:supp-benchmark-b0}
\end{figure*}
\section{Full Results}
\label{sec:suppl-full-results}
In this section, we provide the complete incremental performance curves for all compared methods. 
While the main paper reports three representative trends, here we include the full set of curves corresponding to Table~1. 
Specifically, Figure~\ref{fig:supp-benchmark-b0} shows the results under the zero-base setting, and Figure~\ref{fig:supp-benchmark-b50} reports the half-base setting. 
Across datasets and split protocols, \name consistently maintains strong performance and outperforms competing methods in most cases.

\section{Details of Compared Methods}
\label{sec:suppl-compared-methods}
We provide details of the compared methods in the main paper. 
For fair comparison, all methods are evaluated with the same pre-trained CLIP backbone, and the reproduction is conducted based on the C3Box toolbox~\citep{sun2026c3box}. 
The methods listed in Table~1 are described as follows:

\begin{itemize}
\item \textbf{SimpleCIL~\citep{zhou2025revisiting}:} 
uses the frozen CLIP visual encoder as a general-purpose feature extractor and removes the language branch during evaluation. For each newly observed class, it computes visual prototypes from the extracted features and performs prediction with a cosine classifier. This baseline reflects the strength of frozen CLIP visual representations without any task-wise parameter update.

\item \textbf{ACIL~\citep{zhuang2022acil}:} 
ACIL learns an analytic classifier for class-incremental learning by updating sufficient statistics and solving the classifier in closed form. 
The original ACIL uses a separately trained visual backbone, whereas we adapt it to our CLIP-based setting by using the same frozen CLIP visual backbone as the feature extractor. 
This removes backbone differences and makes the comparison focus on the analytic classification mechanism.

\item \textbf{RanPAC~\citep{mcdonnell2023ranpac}:}
RanPAC combines pre-trained representations with random projections and parameter-efficient adaptation for continual learning. 
For fair comparison, we evaluate it under the same CLIP backbone and the unified training budget used in our benchmark.
This allows us to compare its PETL-based random-projection pipeline with our frozen-backbone visual-only statistic-based classifier under a shared pre-trained representation.

\item \textbf{DualPrompt~\citep{wang2022dualprompt}:} 
is a prompt-based continual learning method that introduces both general prompts and expert prompts on top of a frozen pre-trained backbone. It selects task-adaptive prompts from a prompt pool to guide the visual representation, and in our comparison it operates on the visual branch of CLIP.

\item \textbf{CODA-Prompt~\citep{smith2023coda}:} 
extends prompt-based adaptation by replacing hard prompt selection with attention-based prompt recombination. Instead of choosing a fixed prompt for each instance, it learns to compose prompts dynamically, while still adapting the frozen CLIP visual branch.

\item \textbf{RAPF~\citep{huang2024class}:} 
is a CLIP-based CIL method that updates the model with adaptive representation adjustment and parameter fusion. It introduces class-separation constraints and decomposed fusion to incorporate new-task information while mitigating interference with previously learned knowledge.

\item \textbf{PROOF~\citep{zhou2025learning}:} 
improves continual learning for vision-language models by introducing expandable projection layers and a cross-modal fusion mechanism. It leverages both visual and textual prototypes and refines their interaction to enhance incremental recognition.

\item \textbf{CLG-CBM~\citep{yu2025language}:} 
builds a language-guided concept bottleneck model for interpretable continual learning. By aligning CLIP representations with semantic concepts, it aims to learn concepts that are understandable and transferable across tasks.

\item \textbf{BOFA~\citep{li2026bofa}:} 
proposes bridge-layer orthogonal low-rank fusion for CLIP-based CIL. It uses lightweight low-rank updates at intermediate layers and imposes orthogonality constraints to reduce interference between old and new tasks during incremental adaptation.
\end{itemize}

\begin{figure*}[ht]
    \centering
    \begin{subfigure}[t]{0.49\textwidth}
        \centering
        \includegraphics[width=1\linewidth]{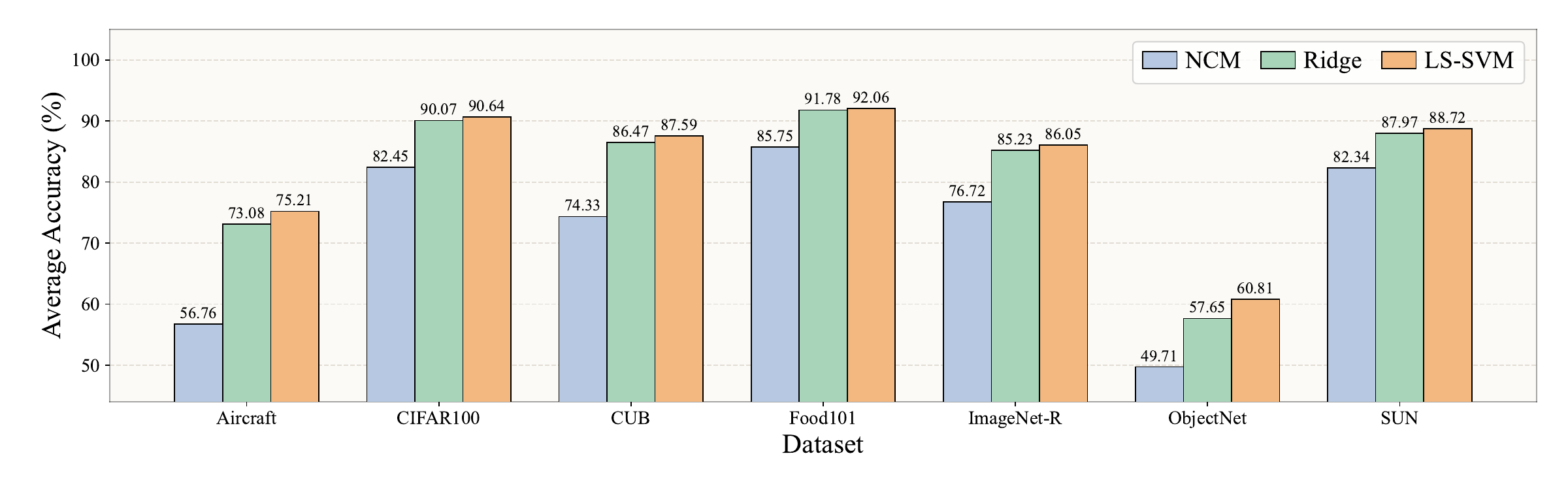}
        \caption{B0 protocol}
        \label{fig:classifier_choice_b0}
    \end{subfigure}%
    \hfill
    \begin{subfigure}[t]{0.49\textwidth}
        \centering
        \includegraphics[width=1\linewidth]{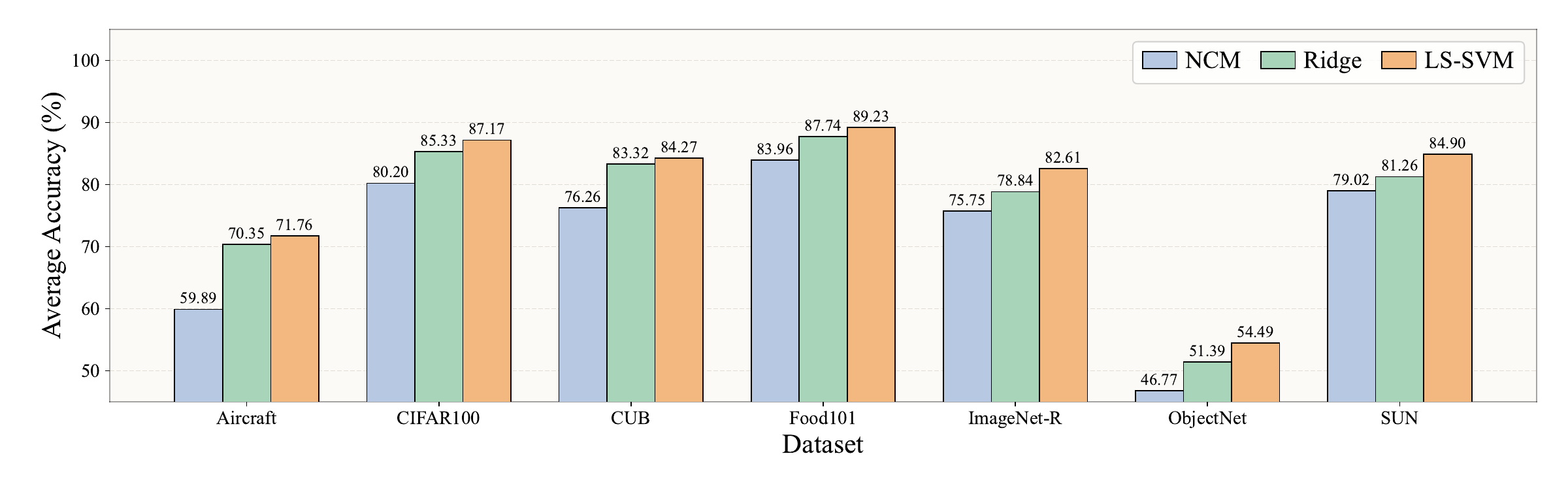}
        \caption{B-half protocol}
        \label{fig:classifier_choice_bhalf}
    \end{subfigure}%

\caption{
Classifier-choice ablation under fixed kernel-induced visual features.
All methods use the same enhanced representation and kernel feature map, and differ only in the final classifier.
NCM uses class means, Ridge uses one-hot least-squares targets, and LS-SVM uses one-vs-all $\pm1$ targets.
Ridge and LS-SVM both outperform NCM, while LS-SVM achieves comparable or better performance than Ridge in most settings, supporting our SVM-style positive-versus-negative classifier formulation.
}
    \label{fig:classifier_choice}
\end{figure*}

\section{Classifier Choice under Fixed Features}
\label{sec:suppl-classifier-choice}

We further study the effect of the final classifier while keeping the feature pipeline fixed. 
All variants use the same enhanced visual representation and the same kernel-induced feature map, and differ only in the final classification rule. 
We compare three classifiers: NCM, which assigns samples to the nearest class mean; Ridge, which solves an ordinary least-squares classifier with one-hot targets; and LS-SVM, which uses one-vs-all $\pm1$ targets.

Figure~\ref{fig:classifier_choice} reports the results under both B0 and B-half protocols. 
NCM is consistently worse than Ridge and LS-SVM across datasets, showing that class centroids alone are insufficient even in the kernel-induced visual feature space. 
Both Ridge and LS-SVM benefit from closed-form least-squares classification on the same features, but LS-SVM achieves comparable or better performance than Ridge in most settings. 
This suggests that the SVM-style one-vs-all formulation is at least as effective as one-hot least-squares classification under the same feature pipeline, while providing a more suitable formulation for our incremental kernelized classifier. 
Unlike one-hot Ridge, where non-target classes are assigned zero targets, LS-SVM uses one-vs-all $\pm1$ coding and explicitly treats every non-ground-truth class as a negative class. 
This positive-versus-negative supervision matches the SVM-style decision formulation and is consistent with our sufficient-statistics update: when new classes arrive, previous samples can be incorporated as negative evidence for the new-class classifiers through the stored feature-sum statistic, without revisiting the original data. 
Moreover, SVMs are classical kernel machines: their decision functions can be formulated in a feature space induced by a kernel, and LS-SVM inherits this kernelized formulation while yielding a closed-form least-squares solution. 
This naturally aligns with our use of a kernel-induced visual feature space. 
Therefore, we adopt LS-SVM as a principled closed-form classifier that retains the analytic-update advantage of Ridge while providing an SVM-style discriminative interpretation for incremental learning.

\section{Comparison with Related Closed-Form CIL Methods}
\label{sec:suppl-linear-comparison}

ACIL~\citep{zhuang2022acil} and RanPAC~\citep{mcdonnell2023ranpac} are closely related to \name in that they construct incremental classifiers from fixed or expanded feature representations without repeatedly optimizing the classifier through gradient descent. 
Nevertheless, \name differs from them in its motivation, representation construction, and classifier update.

\noindent\textbf{Motivation:}
ACIL is primarily motivated by preserving historical knowledge without replay, and derives a recursive least-squares update that reproduces its joint-learning solution. 
RanPAC instead focuses on exploiting pre-trained representations for continual learning while avoiding forgetting from repeated parameter updates. 
In contrast, \name starts from a CLIP-specific observation: textual classifier weights can be misaligned with visual class distributions and lead to less favorable classifier optimization. 
Our goal is therefore to examine whether the textual branch is necessary for CLIP-based CIL and to construct the incremental classifier entirely in the visual space.

\noindent\textbf{Representation Construction:}
Both ACIL and RanPAC increase classifier capacity through randomized feature expansion. 
ACIL applies a randomly initialized feature-expansion layer with nonlinear activation to the extracted representation, whereas RanPAC combines pre-trained features with an optional first-session PETL adaptation followed by a fixed nonlinear random projection. 
\name differs primarily in how the representation before this expansion is constructed: it learns a residual correction from multi-level CLIP visual features using only base-session data, thereby adapting the final visual representation to the downstream distribution before applying the fixed nonlinear feature map.\looseness -1

\noindent\textbf{Classifier Update:}
The main difference lies in the target formulation and its consequence for class expansion. 
ACIL uses one-hot labels and recursively updates a regularized least-squares classifier, while RanPAC accumulates a Gram matrix and class prototypes, which likewise correspond to a regularized least-squares solution with one-hot targets. 
For these formulations, historical samples have zero targets for newly introduced output dimensions. 
In contrast, \name adopts a one-vs-all LS-SVM with $\pm1$ targets. 
When a new class arrives, every historical sample should contribute a target of $-1$ to its classifier; \name therefore maintains the additional feature-sum statistic $\mathbf{s}_t$ to recover this contribution without revisiting previous data, and recomputes the classifiers for all seen classes from $(G_t,Q_t,\mathbf{s}_t)$. 
The effect of this $\pm1$ formulation relative to one-hot Ridge regression is further isolated in Section~\ref{sec:suppl-classifier-choice}.

\noindent\textbf{Empirical Comparison:}
Under the same CLIP backbone and continual-learning protocols, \name consistently outperforms both ACIL and RanPAC in average and final-session accuracy across the benchmark settings in Table~\ref{tab:benchmark}. 
Together with the controlled Ridge-versus-LS-SVM comparison in Section~\ref{sec:suppl-classifier-choice}, these results indicate that the gains of \name do not simply come from randomized feature expansion or least-squares-style classifier updates, but from the combination of task-adaptive visual representation construction and the proposed one-vs-all LS-SVM formulation.

\section{Supplementary Preliminary Experiments}
\label{sec:suppl-pre}

We provide additional preliminary experiments to support the diagnosis in the main paper. 
These results further examine the relationship between image-text modality gap, text-head degradation, and the optimization behavior of textual versus image-prototype initialization.

\begin{figure*}[ht]
    \centering
    \includegraphics[width=\textwidth]{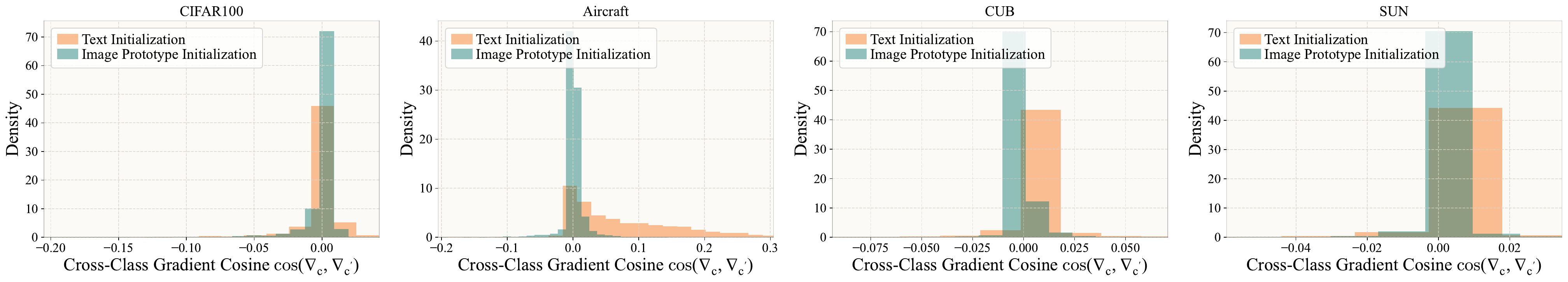}
    \caption{
    Cross-class gradient cosine distributions under textual initialization and image-prototype initialization on representative datasets.
    We compute pairwise cosine similarities $\cos(\nabla_c,\nabla_{c'})$ between class-wise gradient directions.
    Compared with textual initialization, image-prototype initialization generally produces distributions that are more concentrated around zero, indicating more balanced class-wise optimization directions.
    }
    \label{fig:suppl-pre-gradhist}
\end{figure*}
\begin{figure*}[ht]
    \centering
    \includegraphics[width=\textwidth]{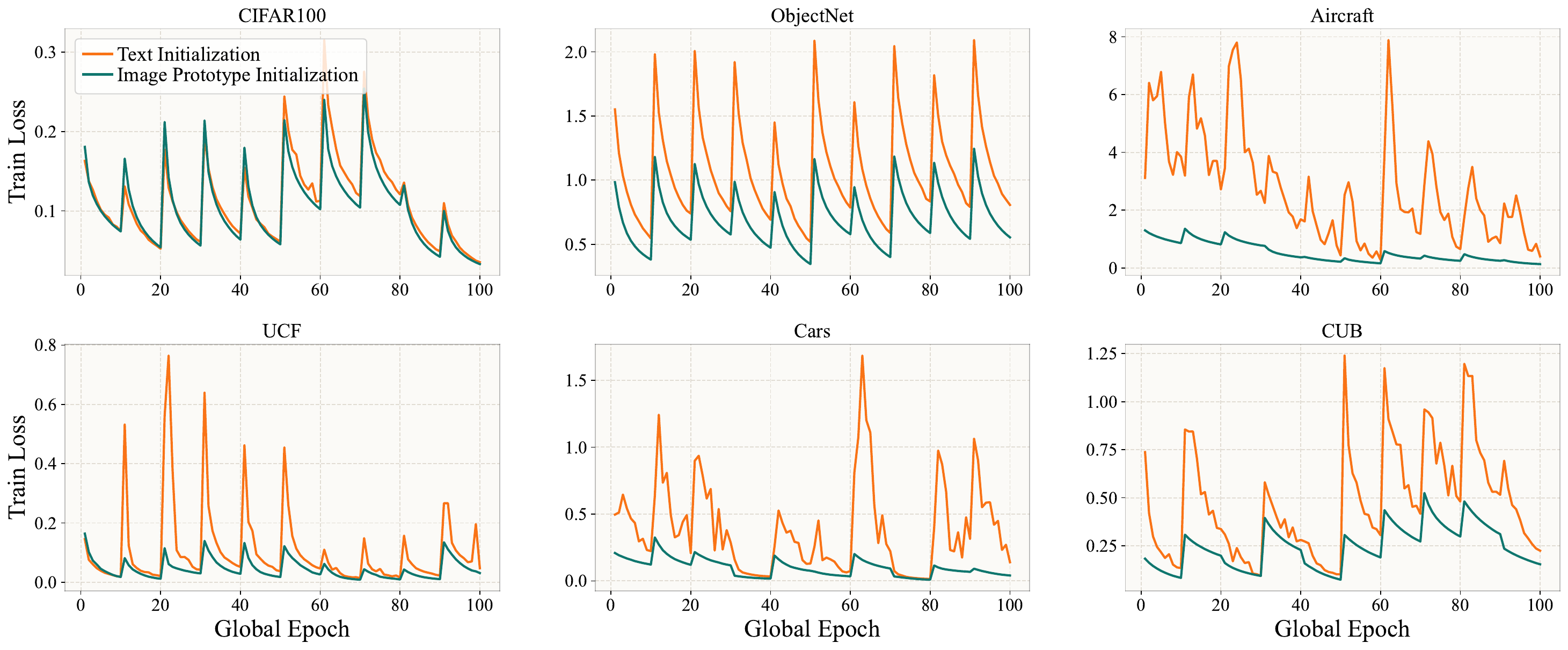}
    \caption{
    Training-loss curves under textual initialization and image-prototype initialization on the same six representative datasets as Figure~\ref{fig:suppl-pre-optim-acc}.
    Textual initialization often leads to larger loss values and stronger loss fluctuations, while image-prototype initialization produces lower or smoother loss curves.
    This suggests that visual prototypes provide a more stable initialization for classifier optimization.
    }
    \label{fig:suppl-pre-optim-loss}
\end{figure*}
\begin{figure*}[t]
    \centering
    \includegraphics[width=\textwidth]{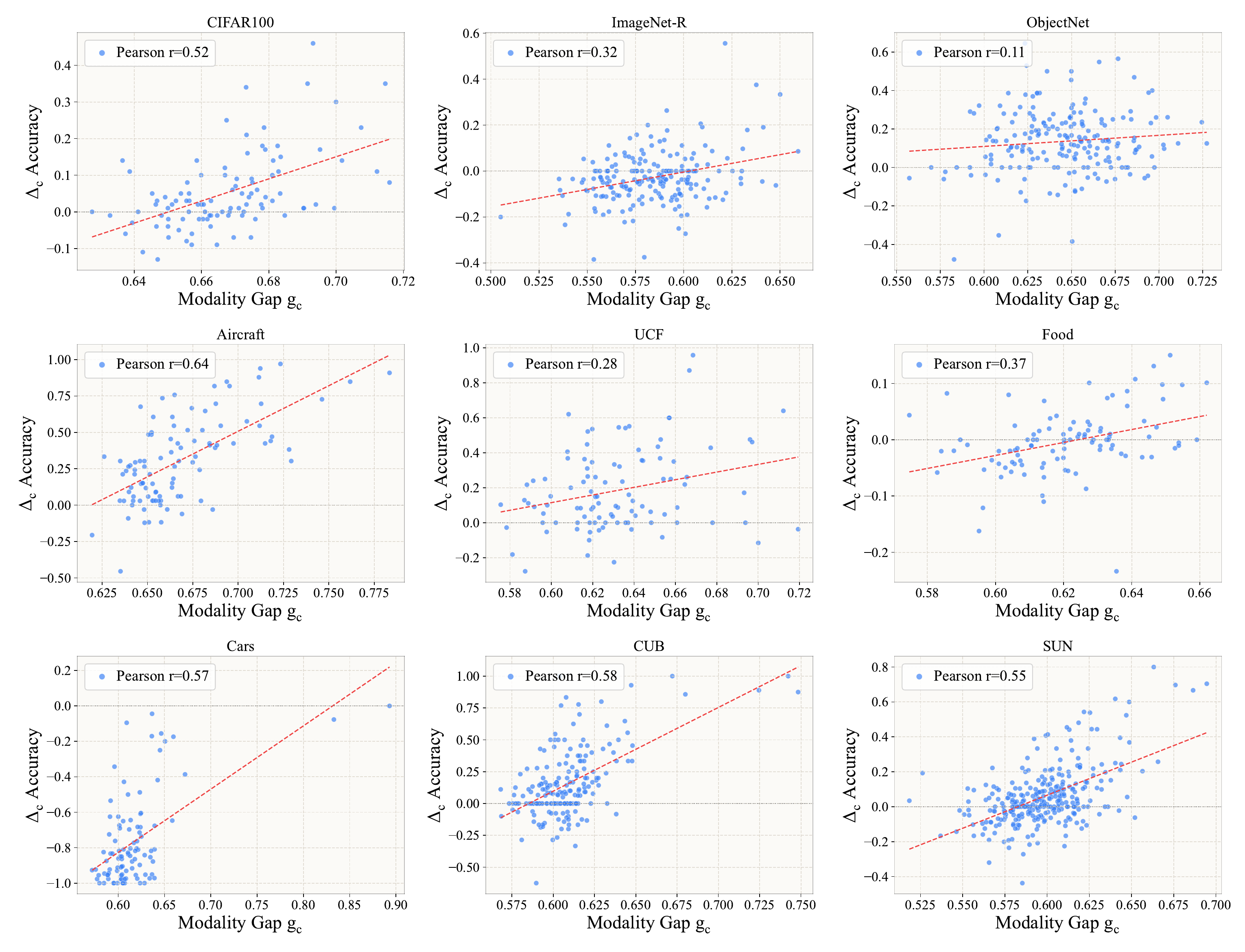}
    \caption{
    Additional class-wise modality-gap analysis on nine datasets.
    Each point denotes one class.
    The x-axis is the class-level modality gap $g_c$ between the image prototype and textual embedding, and the y-axis is the class-wise accuracy improvement $\Delta_c$ of image-prototype initialization over textual initialization.
    The dashed red line shows the linear fit, and the Pearson correlation coefficient is reported in each subplot.
    Most datasets show a positive correlation, suggesting that classes with larger image-text gaps tend to benefit more from image-prototype initialization.
    }
    \label{fig:suppl-pre-gapacc}
\end{figure*}

\noindent{\bf Additional modality-gap analysis.}
Figure~\ref{fig:suppl-pre-gapacc} extends the class-wise gap analysis to nine datasets. 
For each class $c$, we compute the modality gap $g_c$ between its image prototype and textual embedding, and measure the class-wise accuracy improvement $\Delta_c$ of image-prototype initialization over textual initialization. 
Each point in the figure corresponds to one class. 
Across most datasets, $g_c$ and $\Delta_c$ show a positive Pearson correlation, although the correlation strength varies across benchmarks. 
This indicates that classes with larger image-text mismatch tend to suffer more from text-based classifier weights, and therefore benefit more from visual prototypes.

\begin{figure*}[ht]
    \vspace{-2mm}
    \centering
    \includegraphics[width=\textwidth]{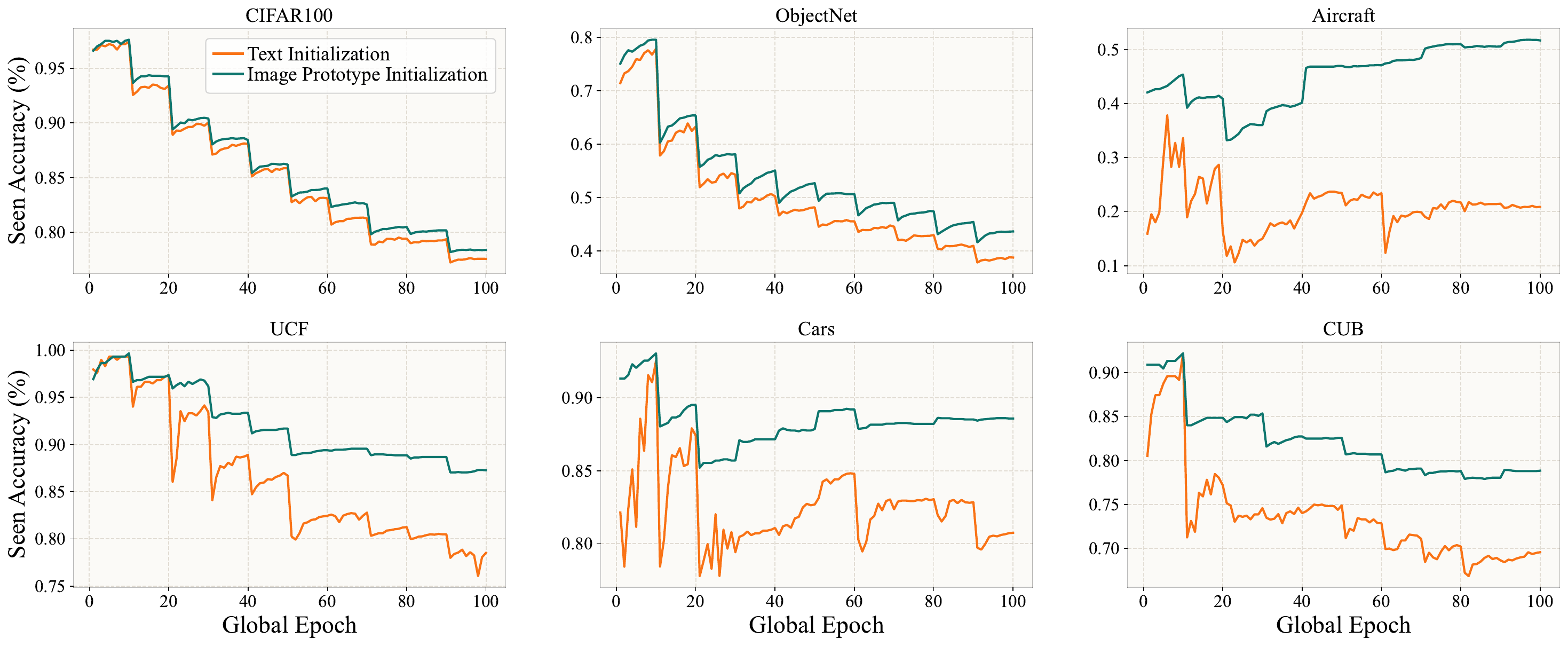}
    \vspace{-5mm}
    \caption{
    Seen-class accuracy curves under textual initialization and image-prototype initialization on six representative datasets.
    The x-axis denotes global training epochs, and the y-axis denotes the accuracy over all seen classes.
    Image-prototype initialization generally yields higher and more stable seen-class accuracy, indicating a more favorable optimization trajectory.
    }
    \label{fig:suppl-pre-optim-acc}
    \vspace{-2mm}
\end{figure*}

\noindent{\bf Additional optimization dynamics.}
We further compare the training dynamics of textual initialization and image-prototype initialization on six representative datasets. 
Figure~\ref{fig:suppl-pre-optim-acc} reports the seen-class accuracy over global epochs, and Figure~\ref{fig:suppl-pre-optim-loss} reports the corresponding training loss. 
Compared with textual initialization, image-prototype initialization generally leads to higher seen-class accuracy and lower or smoother training loss. 
The difference is especially clear on datasets such as ObjectNet, Aircraft, UCF, Cars, and CUB, where textual initialization results in larger loss spikes or worse seen-class accuracy. 
These results provide additional evidence that textual embeddings may provide a less favorable starting point for optimizing the classifier.

\noindent{\bf Cross-class gradient statistics.}
To further inspect the optimization behavior, we analyze cross-class gradient cosine similarity on representative datasets. 
For each class $c$, we compute the gradient direction $\nabla_c$ induced by samples from that class, and then measure pairwise cosine similarities $\cos(\nabla_c,\nabla_{c'})$ between different classes. 
This statistic reflects how aligned or conflicting the class-wise optimization directions are. 
A more concentrated distribution around zero indicates less biased cross-class coupling and more balanced optimization directions. 
As shown in Figure~\ref{fig:suppl-pre-gradhist}, image-prototype initialization generally produces more concentrated gradient-cosine distributions, whereas textual initialization often shows broader distributions or heavier tails. 
This provides another view of why textual initialization can be less favorable for classifier optimization.

\begin{figure*}[ht]
    \centering
    \includegraphics[width=\textwidth]{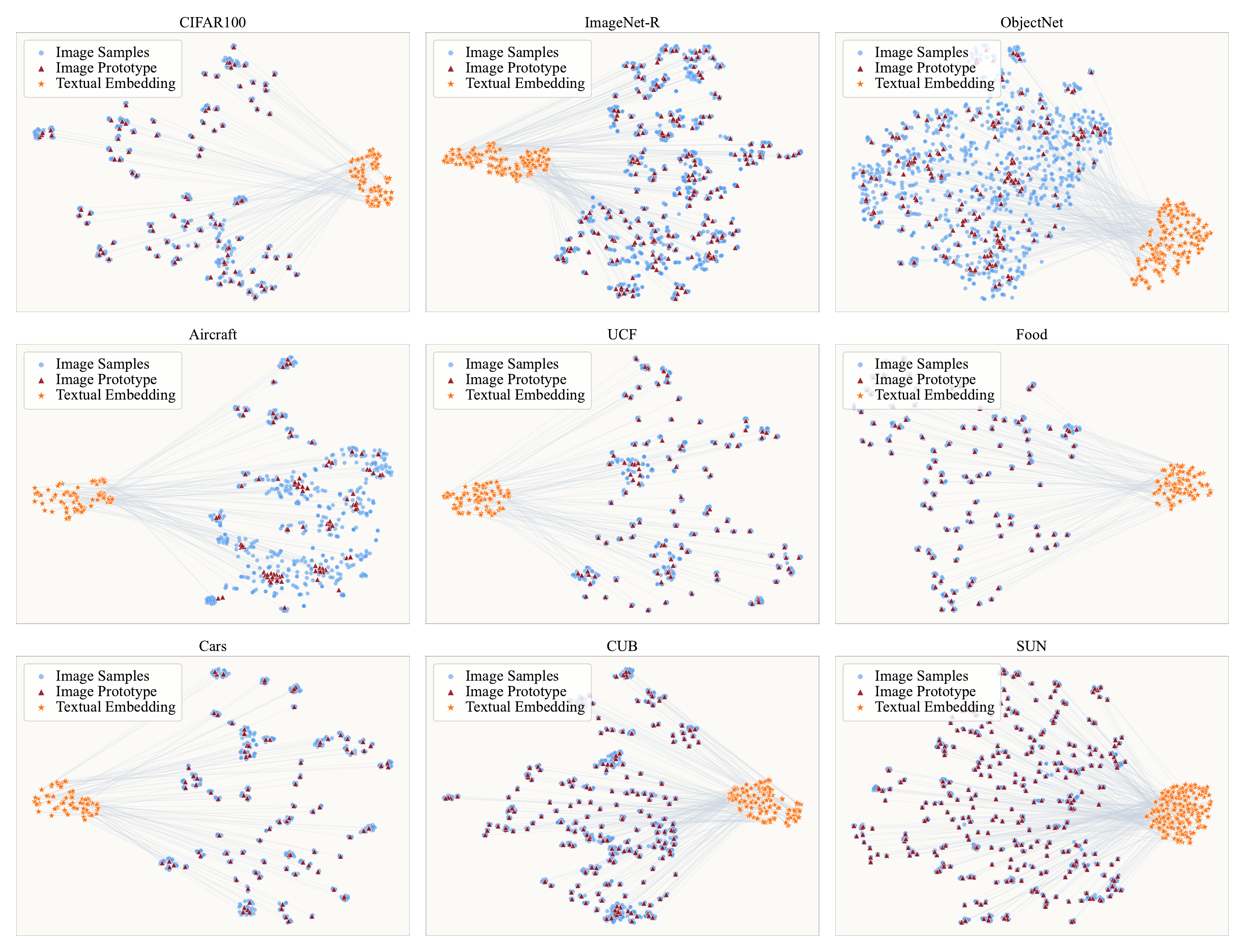}
    \caption{
    Additional t-SNE visualizations of image samples, image prototypes, and textual embeddings on nine datasets.
    Blue dots denote image samples, red triangles denote image prototypes, and orange stars denote textual embeddings.
    Textual embeddings often lie in regions separated from the corresponding image distributions, while image prototypes remain close to image samples.
    This illustrates the image-text modality gap in the shared CLIP embedding space.
    }
    \label{fig:suppl-pre-tsne}
\end{figure*}

\noindent{\bf Additional t-SNE visualization.}
Figure~\ref{fig:suppl-pre-tsne} provides additional t-SNE visualizations of image samples, image prototypes, and textual embeddings in the shared CLIP embedding space. 
Across datasets, textual embeddings often occupy regions that are clearly separated from the corresponding image distributions, while image prototypes lie much closer to image samples. 
This qualitative observation is consistent with the modality-gap analysis above and supports the motivation of constructing the classifier in the visual space.

\end{document}